\documentclass[acmsmall]{acmart}

\setcopyright{acmcopyright}
\acmJournal{PACMHCI}
\acmYear{2021} \acmVolume{5} \acmNumber{CSCW2} 
\acmArticle{408} \acmMonth{10} \acmPrice{15.00}
\acmDOI{10.1145/3479552}

\usepackage{graphics, amsmath}
\usepackage{todonotes}
\usepackage{hyperref}
\usepackage{xspace}
\usepackage{siunitx}
\usepackage{subcaption}
\usepackage{enumitem}
\usepackage{multirow}
\usepackage{booktabs}
\usepackage{xcolor}
\usepackage{pifont}

\usepackage{collectbox}

\makeatletter
\newcommand{\mybox}{%
    \collectbox{%
        \setlength{\fboxsep}{1pt}%
        \fbox{\BOXCONTENT}%
    }%
}
\makeatother

\usepackage{amsthm} %
\newtheorem{theorem}{Theorem}
\newtheoremstyle{exampstyle}
  {2pt} %
  {2pt} %
  {\em} %
  {} %
  {} %
  {.} %
  {.5em} %
  {} %
\theoremstyle{exampstyle} \newtheorem{proposition}[theorem]{Proposition}

\definecolor{blue}{HTML}{457b9d}
\definecolor{red}{HTML}{e63946}
\definecolor{gray}{HTML}{808080}

\newcommand{\domain}{{distribution}\xspace}
\newcommand{\domains}{{distributions}\xspace}
\newcommand{\Domain}{{Distribution}\xspace}
\newcommand{\ind}{{in-dis\-tri\-bu\-tion}\xspace}
\newcommand{\Ind}{{In-distribution}\xspace}
\newcommand{\shortind}{{IND}\xspace}
\newcommand{\ood}{{out-of-dis\-tri\-bu\-tion}\xspace}
\newcommand{\Ood}{{Out-of-distribution}\xspace}
\newcommand{\shortood}{{OOD}\xspace}
\newcommand{\humanaiteam}{{human-AI team}\xspace}
\newcommand{\humanaiteams}{{human-AI teams}\xspace}
\newcommand{\Humanaiteam}{{Human-AI team}\xspace}
\newcommand{\Humanaiteams}{{Human-AI teams}\xspace}
\newcommand{\biostask}{{\sf BIOS}\xspace}
\newcommand{\icpsrtask}{{\sf ICPSR}\xspace}
\newcommand{\compastask}{{\sf COMPAS}\xspace}

\newcommand{\para}[1]{\noindent{\bf #1}}
\newcommand{\figref}[1]{Fig.~\ref{#1}}
\newcommand{\secref}[1]{\S\ref{#1}}
\newcommand{\tabref}[1]{Table~\ref{#1}}
\newcommand{\cmark}{\textcolor{blue}{\ding{51}}}%
\newcommand{\hcmark}{$\textcolor{gray}{\mybox{\text{\ding{51}}}}$}
\newcommand{\xmark}{\textcolor{red}{\ding{55}}}%
\newcommand{\hxmark}{$\textcolor{gray}{\mybox{\text{\ding{55}}}}$}
\newcommand{\reversedmark}{{\large \bf\textcolor{red}{!}}}%
\newcommand{\hreversedmark}{$\textcolor{gray}{\mybox{\large \bf !}}$}
\graphicspath{{figs/}}

\AtBeginDocument{%
  \providecommand\BibTeX{{%
    \normalfont B\kern-0.5em{\scshape i\kern-0.25em b}\kern-0.8em\TeX}}}

\begin{document}

\title[The Effect of Out-of-distribution Examples and Interactive Explanations]{Understanding the Effect of \Ood Examples and Interactive Explanations on Human-AI Decision Making}

\author{Han Liu}
\affiliation{%
   \institution{University of Chicago}
   \department{Department of Computer Science}
   \city{Chicago}
   \state{IL}
   \country{USA}}
\email{hanliu@uchicago.edu}

\author{Vivian Lai}
\affiliation{%
   \institution{University of Colorado Boulder}
   \department{Department of Computer Science}
   \city{Boulder}
   \state{CO}
   \country{USA}}
\email{vivian.lai@colorado.edu}

\author{Chenhao Tan}
\affiliation{%
   \institution{University of Chicago}
   \department{Department of Computer Science}
   \city{Chicago}
   \state{IL}
   \country{USA}}
\email{chenhao@uchicago.edu}

\keywords{Human-AI decision making; distribution shift; interactive explanations; complementary performance; appropriate trust}

\ccsdesc[500]{Human-centered computing~Collaborative and social computing}
\ccsdesc[500]{Computing methodologies~Artificial intelligence}
\ccsdesc[500]{Applied computing~Law, social and behavioral sciences}

\begin{abstract}
Although AI holds promise for improving human decision making in societally critical domains, it remains an open question how human-AI teams can reliably outperform AI alone and human alone in {\em challenging} prediction tasks (also known as {\em complementary performance}). We explore two directions to understand the gaps in achieving complementary performance. First, we argue that the typical experimental setup limits the potential of human-AI teams. To account for lower AI performance \ood than \ind because of distribution shift, we design experiments with different \domain types and investigate human performance for both \ind and \ood examples. Second, we develop novel interfaces to support interactive explanations so that humans can actively engage with AI assistance. Using virtual pilot studies and large-scale randomized experiments across three tasks, we demonstrate a clear difference between \ind and \ood, and observe mixed results for interactive explanations: while interactive explanations improve human perception of AI assistance's usefulness, they may reinforce human biases and lead to limited performance improvement.
Overall, our work points out critical challenges and future directions towards enhancing human performance with AI assistance.
\end{abstract}

\maketitle

\section{Introduction}
\label{sec:intro}

\begin{figure}
	\centering
	\includegraphics[trim=0 20 0 0,clip,width=0.82\textwidth]{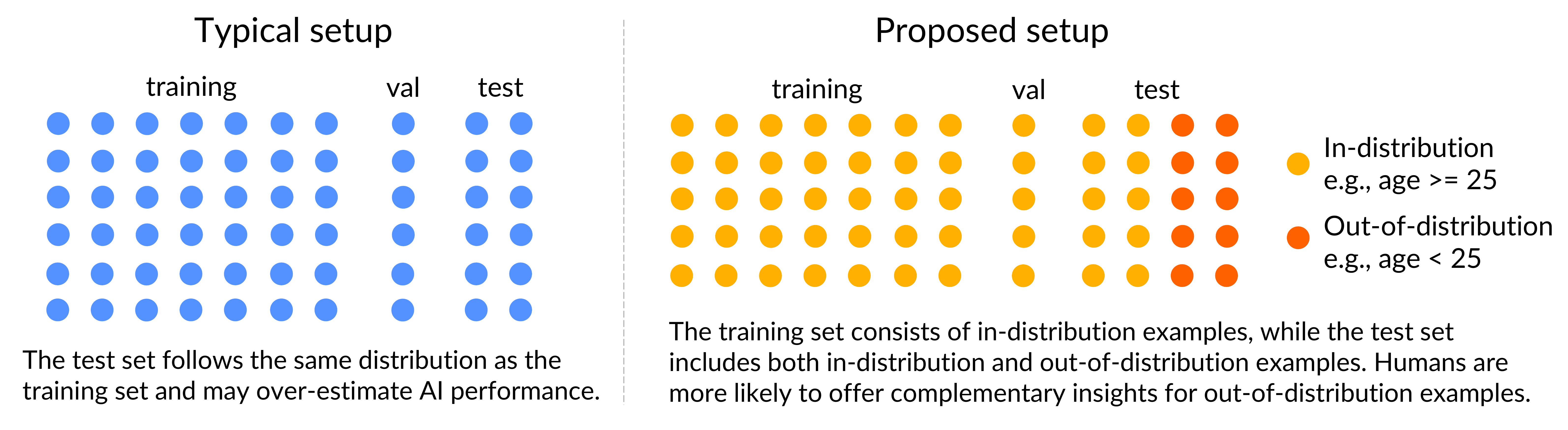}
	\Description{}
  \caption{An illustration of the typical setup and our proposed setup that takes into account \domain types.
  For instance, in the recidivism prediction task we can use defendants of younger ages to simulate \ood examples,
  assuming our training set only contains older defendants referred as \ind examples.
  The fractions of data are only for illustrative purposes.
  See details of \ind vs. \ood setup in \secref{sec:ind_ood}.
  }
	\label{fig:ind_ood}
\end{figure}

As AI performance grows rapidly and often surpasses humans in constrained tasks \citep{kleinberg2018human,he2015delving,mckinney2020international,silver2018general,brown2019superhuman}, 
a critical challenge to enable social good is to understand how AI assistance can be used to enhance {\em human performance}.
AI assistance 
has been shown to
improve people's efficiency in tasks such as transcription
by enhancing their computational capacity~\citep{lasecki2017scribe,gaur2016effects},
support creativity in producing music~\citep{louie2020novice,mccormack2019silent,frid2020music},
and even allow the 
visually impaired
to ``see'' images~\citep{wu2017automatic,gurari2018vizwiz}.
However, it remains difficult
to enhance human decision making in 
{\em challenging} prediction tasks~\cite{kleinberg2015prediction}.
Ideally, with AI assistance, \humanaiteams should outperform AI alone and human alone (e.g., in accuracy; also known as {\em complementary performance} \citep{bansal2021does}).
Instead, researchers have found that while AI assistance improves human performance compared to human alone,
\humanaiteams seldom outperform AI alone in a wide variety of tasks, including recidivism prediction, deceptive review detection, and hypoxemia prediction~\cite{lai+tan:19,lai+liu+tan:20,green2019principles,green2019disparate,zhang2020effect,poursabzi2021manipulating,carton2020feature,jung2020limits,weerts2019human,beede2020human,wang2021explanations,lundberg2018explainable}.
To address the elusiveness of complementary performance,
we study 
two factors: 1) an overlooked factor in the experimental setup that may over-estimate AI performance; 2) the lack of two-way conversations between humans and AI, which may limit
human understanding of AI predictions. 
First, we argue that prior work 
adopts a best-case scenario 
for AI.
Namely, these experiments randomly split a dataset into a training set and a test set (\figref{fig:ind_ood}).
The training set is used to train the AI, and the test set is used to evaluate AI performance and human performance (with AI assistance).
We hypothesize that this evaluation scheme is too optimistic for AI performance and provide limited opportunities for humans to contribute insights because the test set follows the same distribution as the training set (\ind).
In practice, examples during testing may differ substantially from the training set, and AI performance can significantly drop for these \ood examples~\citep{mccoy-etal-2019-right,clark2019don,jia2017adversarial}.
Furthermore, humans are better equipped to detect problematic patterns in AI predictions and offer complementary insights in \ood examples.
Thus, we propose to develop experimental designs with both \ood examples and \ind examples in the test set.

Second, although explaining AI predictions has been hypothesized to help humans understand AI predictions and thus improve human performance \citep{doshi2017towards}, 
static explanations, such as highlighting important features and showing AI confidence, have been mainly explored so far \citep{green2019principles,lai+tan:19,bansal2021does}.
Static explanations represent a one-way conversation from AI to humans and may be insufficient for humans to understand AI predictions.
In fact, psychology literature suggests that interactivity is a crucial component in explanations \citep{lombrozo2006structure,miller2018explanation}.
Therefore, we develop interactive interfaces to enable a two-way conversation between decision makers and AI.
For instance, we allow humans to change the input and observe how AI predictions would have changed in these counterfactual scenarios (\figref{fig:interactive_screenshot}).
We hypothesize that interactive explanations improve the performance of humans and their subjective perception of AI assistance's usefulness.
Although \ood examples and interactive explanations are relatively separate research questions, we study them together in this work as we hypothesize that they are critical missing ingredients towards complementary performance.

To investigate the effect of \ood examples and interactive explanations on human-AI decision making, we choose three datasets spanning two tasks informed by prior work:
1) recidivism prediction (\compastask and \icpsrtask) (a canonical task that has received 
much
attention due to its importance; \compastask became popular because of the ProPublica article on machine bias \citep{angwin2016machine}, and \icpsrtask was recently introduced to the human-AI interaction community by \citet{green2019principles,green2019disparate}, so it would be useful to see whether same results hold in both datasets);
2) profession detection (\biostask) (the task is to predict a person's profession based on a short biography; this task is substantially easier than recidivism prediction and other text-based tasks such as deceptive review detection, so crowdworkers may have more useful insights to offer for this task).
We investigate human-AI decision making in these tasks through both virtual pilot studies and large-scale randomized experiments.
We focus on the following three research questions:

\begin{itemize}[topsep=3pt]
    \item {\bf RQ1}: how do \domain types affect the performance of \humanaiteams, compared to AI alone?
    \item {\bf RQ2}: how do \domain types affect human agreement with AI predictions?
    \item {\bf RQ3}: how do interactive explanations affect human-AI decision making?
\end{itemize}

Our results demonstrate a clear difference between \ind and \ood.
Consistent with prior work, we find that \humanaiteams tend to underperform AI alone in \ind examples in all tasks.
In comparison, \humanaiteams can occasionally outperform AI in \ood examples in recidivism prediction (although the difference is small).
It follows that the performance gap between \humanaiteams and AI is smaller \ood than \ind, confirming that humans are more likely to achieve complementary performance \ood.

\Domain types also affect human agreement with AI predictions.
In recidivism prediction (\compastask and \icpsrtask), humans are more likely to agree with AI predictions \ind than \ood, 
suggesting that humans 
behave differently 
depending on the \domain type.
Moreover, in recidivism prediction, 
human agreement with {\em wrong AI predictions} is lower \ood than \ind, suggesting that humans may be better at providing complementary insights into AI mistakes \ood.
However, in \biostask, where humans may have more intuitions for detecting professions, humans are less likely to agree with AI predictions \ind than \ood.
This observation also 
explains 
the relatively low \ind performance of \humanaiteams in \biostask compared to AI alone.

Finally, although we do not find that interactive explanations lead to improved performance for \humanaiteams, they significantly increase human perception of AI assistance's usefulness.
Participants with interactive explanations are more likely to find real-time assistance useful in \icpsrtask and \compastask, and training more useful in \compastask.
To better understand the limited utility of interactive explanations, we conduct an exploratory study on what features participants find important in recidivism prediction.
We find that participants with interactive explanations are more likely to fixate on demographic features such as age and race, and less likely to identify the computationally important features based on Spearman correlation.
Meanwhile, they make more mistakes when they disagree with AI.
These observations suggest that interactive explanations might reinforce existing human biases and lead to suboptimal decisions.

Overall, we believe that our work 
adds value to the community in the emerging field of human-AI collaborative decision making in {\em challenging} prediction tasks.
Our work points out an important direction in designing future experimental studies on human-AI decision making:
it is critical to think about the concept of \ood examples and evaluate the performance of \humanaiteams both \ind and \ood.
The implications for interactive explanations are mixed.
On the one hand, interactive explanations improve human perception of AI usefulness, despite not reliably improving their performance.
On the other hand,
similar to ethical concerns about static explanations raised in prior work \citep{green2019disparate,green2019principles,bansal2021does}, interactive explanations might reinforce existing human biases.
It is critical to take these factors into account when developing and deploying improved interactive explanations.
Our results also highlight the important role that task properties may play in shaping human-AI collaborative decision making and provide valuable samples for exploring the vast space of tasks.

\section{Related work and Research Questions}
\label{sec:related}

In this section, we review related work and formulate our research questions.

\subsection{Performance of Human-AI Teams in Prediction Tasks}

With a growing interest in understanding human-AI interaction, many recent studies have worked on 
enhancing human performance with AI assistance in decision making.
Typically, these decisions are formulated as prediction tasks where AI can predict the outcome and may offer explanations, e.g., by highlighting important features.
For instance, the bailing decision (whether a defendant should be bailed) can be formulated as a prediction problem of whether a defendant will violate pretrial terms in two years \citep{kleinberg2018human}.
Most studies have reported results aligning with the following proposition:

\begin{proposition}
\label{proposition}
AI assistance improves human performance compared to without any assistance; however, the performance of \humanaiteams seldom surpasses AI alone in challenging prediction tasks~\cite{lai+tan:19,lai+liu+tan:20,green2019principles,green2019disparate,zhang2020effect,poursabzi2021manipulating,carton2020feature,jung2020limits,weerts2019human,beede2020human,lundberg2018explainable,wang2021explanations,buccinca2020proxy}.%
\footnote{Our focus in this work is on understanding the performance of \humanaiteams compared to AI performance
and do not recommend AI to replace humans in any means.
In fact, many studies have argued that humans should be the final decision makers in societally critical domains for ethical and legal reasons such as recidivism prediction and medical diagnosis \citep{green2019principles,lai+tan:19,nytimes,wicourts,daubert}.
}
\end{proposition}
This proposition is supported in a wide variety of tasks, 
including recidivism prediction \citep{green2019principles,green2019disparate,jung2020limits}, deceptive review detection \citep{lai+tan:19,lai+liu+tan:20}, 
income prediction \citep{poursabzi2021manipulating}, 
and hypoxemia prediction \citep{lundberg2018explainable},
despite different forms of AI assistance.
To understand this observation, we point out that Proposition \ref{proposition} entails that AI alone outperforms humans alone in these tasks (human $<$ human + AI $<$ AI).
\citet{lai+liu+tan:20} conjectures that 
the tasks where humans need AI assistance typically fall into the {\em discovering} mode, where the groundtruth is determined by (future) external events (e.g., a defendant's future behavior) rather than human decision makers, instead of the {\em emulating} mode, where humans (e.g., crowdworkers) ultimately define the groundtruth.\footnote{In fact, it is unclear what complementary performance means in the {\em emulating} mode if humans define the groundtruth as human performance is by definition 100\%. A more subtle discussion can be found in footnote \ref{footnote}.}
We refer to prediction tasks in the discovering mode as \textit{challenging prediction tasks}. Example tasks include the aforementioned recidivism prediction, deception detection, hypoxemia prediction, etc. These tasks are non-trivial to humans and two corollaries follow: 1) human performance tend to be far from perfect; 2) the groundtruth labels cannot be crowdsourced.\footnote{Whether a task is challenging (in the discovering mode) also depends on characteristics of humans. For instance, sentiment analysis of English reviews might not be challenging for native speakers, but could remain challenging for non-native speakers.}
In such tasks, AI can identify non-trivial and even counterintuitive patterns to humans. 
These patterns can be hard for humans to digest and leverage when they team up with AI.
As such, it is difficult for \humanaiteams~to achieve complementary performance.

A notable exception is \citet{bansal2021does}, which shows that \humanaiteam performance surpasses AI performance in sentiment classification (beer reviews and Amazon reviews) and LSAT question answering. 
Their key hypothesis 
is that \humanaiteams are likely to excel when human performance and AI performance are comparable, while prior studies tend to look at situations where the performance gap is substantial.
It naturally begs the question of what size of performance gap counts as comparable performance, whether comparable performance alone is sufficient for complementary performance, and whether other factors are associated with the observed complementary performance (we summarize the definitions of complementary performance and comparable performance in \tabref{tb:definitions} to help readers understand these concepts). 
For instance, it is useful to point out that sentiment analysis 
is closer to the emulating mode.\footnote{Although labels in sentiment analysis are determined by the original author, sentiment analysis is generally viewed as a natural language understanding task that humans are capable of.
AI is thus designed to emulate human capability.
In the emulating mode, improving human performance is essentially aligning single-person decisions with the majority of a handful of annotators.
We argue that data annotation is qualitatively different from decision making in challenging tasks such as recidivism prediction.
\label{footnote}
}
We will provide a more in-depth discussion in \secref{sec:discussion}.

\begin{table}
\centering
\begin{tabular}{p{0.95\textwidth}}
\toprule
{\bf Complementary performance}. An ideal outcome of human-AI collaborative decision making: the performance of \humanaiteams is better than AI alone and human alone.\\
{\bf Comparable performance}. The performance of human alone is {\em similar} to AI alone, 
yielding more potential for complementary performance as hypothesized in \citet{bansal2021does}. 
There lacks a quantitative definition of what performance gap counts as comparable. We explore different ranges in this work.\\
\bottomrule
\end{tabular}
\medskip
\caption{Definitions of complementary performance and comparable performance.
}
\label{tb:definitions}
\end{table}

Our core 
hypothesis
is that a standard setup in current experimental studies on human-AI interaction might limit the potential of \humanaiteams.
Namely, researchers typically follow standard machine learning setup in evaluating classifiers by randomly splitting the dataset into a training set and a test set, and using the test set to evaluate the performance of \humanaiteams and AI alone.
It follows that the data distribution in the test set is similar to the training set by design.
Therefore, this setup is designed for AI to best leverage the patterns learned from the training set and provide a strong performance.
In practice, a critical growing concern is distribution shift \citep{goodfellow2016machine,quionero2009dataset,sugiyama2012machine}.
In other words, the test set may differ from the training set, so the patterns that AI identifies can fail during testing, leading to a substantial drop in AI performance~\citep{mccoy-etal-2019-right,clark2019don,jia2017adversarial}. 
Throughout this paper, we refer to testing examples that follow the same distribution as the training set as \ind (\shortind) examples and that follow a different distribution as \ood (\shortood) examples.

Thus, our first research question ({\bf RQ1})
examines how distribution types affect the performance of human-AI teams, compared to AI alone.
We expect our results in \ind examples to replicate previous findings and be consistent with Proposition \ref{proposition}.
In comparison, we hypothesize that humans are more capable of spotting problematic patterns and mistakes in AI predictions when examples are not similar to the training set (\ood), as humans might be robust against distribution shift.
Even if \humanaiteams do not outperform AI alone in \ood examples, we expect the performance gap between \humanaiteams and AI alone to be smaller \ood than \ind.
Inspired by the above insights on comparable performance, we choose three tasks where humans and AI have performance gaps of different sizes so that 
we can investigate the effect of distribution type across tasks.

\subsection{Agreement with AI}
In addition to human performance, human agreement with AI predictions is critical for 
understanding human-AI interaction, especially in tasks where humans are the final decision makers.
When AI predictions are explicitly shown, this agreement can also be interpreted as the trust that humans place in AI.
Prior work has found that in general, the more information about AI predictions is given, the more likely humans are going to agree with AI predictions~\citep{lai+tan:19,feng2019can,bansal2021does,ghai2020explainable}.
For instance, 
explanations, presented along with AI predictions, increase the likelihood that humans agree with AI~\citep{lai+liu+tan:20,bansal2021does,ghai2020explainable}.
Confidence levels have also been shown to help humans calibrate whether to agree with AI \citep{zhang2020effect,bansal2021does}.
In a similar vein, \citet{yin2019understanding} investigate the effect of observed and stated accuracy on humans' trust in AI and find that both stated and observed accuracy can affect human trust in AI.
Finally, expertise may shape humans' trust in AI:
\citet{feng2019can} find that novices in Quiz Bowl trust the AI more than experts when visualizations are enabled.
However, little is known about the effect of \domain types as it has not been examined in prior work.
Our second research question ({\bf RQ2}) inquires into the effect of \domain types on human agreement with AI predictions.
We hypothesize that humans are more likely to agree with AI \ind than \ood because the patterns that AI learns from \ind examples may not apply \ood and AI performance is worse \ood than \ind .
Furthermore, given prior results that humans are more likely to agree with correct AI predictions than wrong AI predictions \citep{lai+tan:19,bansal2021does}, it would be interesting to see whether that trend is different \ood from \ind.%

Additionally, we are interested in having a closer look at the effect of \domain types on human agreement by zooming in on the correctness of AI predictions.
Prior work has introduced three terms to address these different cases of agreement~\citep{wang2021explanations}: appropriate trust~\cite{mcbride2010trust,mcguirl2006supporting,merritt2015well,muir1987trust} (the fraction of instances where humans agree with correct AI predictions and disagree with wrong AI predictions; this is equivalent to \humanaiteam accuracy in binary classification tasks), overtrust~\cite{parasuraman1997humans,de2014design} (the fraction of instances where humans agree with wrong AI predictions), and undertrust~\cite{parasuraman1997humans,de2014design} (the fraction of instances where humans disagree with correct AI predictions). 
To simplify the measurement, we only consider agreement with AI predictions in this work because disagreement and agreement add up to 1. 
We define the fraction of instances where humans agree with {\it correct} AI predictions as {\it appropriate agreement} and the fraction of instances where humans agree with {\it incorrect} AI predictions as {\it overtrust},
and similarly the counterparts in disagreement as {\it undertrust} and {\it appropriate disagreement}.
\tabref{tb:agreement} shows the full combinations of human agreement and AI correctness. 
The term {\it appropriate trust} then is 
the sum of {\it appropriate agreement} and {\it appropriate disagreement}.
We hypothesize that patterns embedded in the AI model may not apply to \ood examples, humans can thus better identify wrong AI predictions in \ood examples (i.e., overtrust is lower \ood).
Similarly, our intuition is that {\it appropriate agreement} is also likely lower \ood as AI may make correct predictions based on non-sensible patterns.
While we focus on how \domain types affect {\it appropriate agreement} and {\it overtrust}, it also entails how \domain types affect {\it undertrust} and {\it appropriate disagreement}.

\begin{table}
  \centering
  \begin{tabular}{ l | c c }
    \toprule
                               & {\bf Correct AI predictions} & {\bf Wrong AI predictions} \\ \hline
    {\bf Humans agree with}    & Appropriate agreement        & Overtrust                  \\
    {\bf Humans disagree with} & Undertrust                   & Appropriate disagreement   \\
    \bottomrule
  \end{tabular}
  \medskip
  \caption{Definition of human agreement based on the correctness of AI predictions.}
  \label{tb:agreement}
\end{table}

\subsection{Interactive Explanations}
A key element in developing AI assistance are explanations of AI predictions,
which have attracted a lot of interest from the research community \citep{lipton2016mythos,doshi2017towards,ribeiro2016should,lundberg2017unified,koh2017understanding,lakkaraju2016interpretable,gilpin2018explaining}.
Experimental studies in human-AI decision making have so far employed 
static explanations 
such as highlighting important features and showing similar examples,
a few studies have also investigated the effect of explanations with an interactive interface.
However, 
literature in social sciences has argued that explanations should be interactive.
For instance, \citet{lombrozo2006structure} suggests that an explanation is a byproduct of an interaction process between an explainer and an explainee,
and \citet{miller2018explanation} says that explanations are social in that they are transferable knowledge that is passed from one person to the other in a conversation.
We hypothesize that 
the one-way conversation in static explanations is insufficient for humans to understand 
AI predictions, contributing to the proposition that \humanaiteams have yet to outperform AI alone.

It is worth pointing out that industry practitioners have worked towards developing interactive interfaces to take advantage of deep learning models' superior predictive power.
For instance, \citet{tenney2020language} develop an interative interpretability tool that provide insightful visualizations for NLP tasks.
Similar interactive tools have been used to support data scientists in debugging machine learning models and improving model performance \cite{kaur2020interpreting,hohman2019gamut,wu2019local}.
While data scientists are familiar with 
machine learning, 
laypeople may not 
have the basic knowledge of machine learning.
We thus focus on developing an interface that enables meaningful interactive explanations for laypeople to support decision making rather than debugging.
Our ultimate goal is to improve human performance instead of model performance.
In addition,
there have been interactive systems that provide AI assistance for complicated tasks beyond constrained prediction tasks \citep{cai2019human,xie2020chexplain,yang2019unremarkable}.
Our scope in this work is limited to explanations of AI predictions where the human task is to make a simple categorical prediction.
Most similar to our work is \citet{cheng2019explaining}, which examines the effect of different explanation interfaces on user understanding of a model and shows improved understandings with interactive explanations,
whereas our work focuses on the effect of interactive explanations on human-AI decision making.
As such, our final research question ({\bf RQ3}) 
investigates the effect of interactive explanations on human-AI decision making.
We hypothesize that interactive explanations lead to better human-AI performance, compared to static explanations.
We further examine the effect of interactive explanations on human agreement with AI predictions.
If interactive explanations enable humans to better critique incorrect AI predictions, then humans may become less reliant on the incorrect predicted labels (i.e., lower overtrust).
Finally, 
we expect interactive explanations to improve subjective perception of usefulness over static explanations because interactive explanations enable users to have two-way conversations with the model. 

\subsection{Differences from Interactive Machine Learning and Transfer Learning}
It is important to 
note that our focus in this work is on how \domain types and interactive explanations affect human performance in decision making and our ultimate goal is to enhance human performance.
While other areas such as transfer learning and interactive machine learning have conducted user studies where people interact with machine learning models, the goal is usually to improve model performance.
Specifically, interactive machine learning tends to involve machine learning practitioners,
while our work considers the population that does not have a machine learning background \cite{hohman2019gamut,krause2016interacting,tenney2020language,wexler2019if}.
Similarly, transfer learning focuses on improving models that would generalize well on other domains (distributions), whereas our work investigates how examples in different \domains affect {\em human performance} \cite{zhuang2020comprehensive,liang2020transfer,torrey2010transfer}.
Although improving AI will likely improve human performance in the long run, we focus on the effect of AI assistance on human decision making where the AI is not updated.

\section{Methods}
\label{sec:methods}

In order to evaluate the performance of \humanaiteams, we consider three important ingredients in this work:
1) Prediction tasks: we consider three prediction tasks that include both tabular and text datasets as well as varying performance gaps between human alone and AI alone (\secref{sec:prediction_tasks});
2) \Ind (\shortind) vs. \ood (\shortood): a key contribution of our work is to highlight the importance of distribution shift and explore ways to design human-AI experimental studies with considerations of \ind and \ood examples (\secref{sec:ind_ood});
3) Explanation type: another contribution of our work is to design novel interactive explanations for both tabular data and text data (\secref{sec:interactive_conditions}).
We further use virtual pilot studies to gather qualitative insights and validate our interface design (\secref{sec:qual}), and then conduct large-scale experiments with crowdworkers on Mechanical Turk (\secref{sec:mturk}).

\subsection{Prediction Tasks}
\label{sec:prediction_tasks}

We use two types of tasks, recidivism prediction, and profession prediction.
Recidivism prediction is based on tabular datasets, while profession prediction is based on text datasets.

\begin{itemize}[topsep=0pt,leftmargin=*]
	\item {\icpsrtask} \cite{icpsr}.
  This dataset was collected by the U.S. Department of Justice.
  It contains defendants who were arrested between 1990 and 2009, and the task is to predict if a defendant will violate the terms of pretrial release.
	Violating terms of pretrial release means that the defendant is rearrested before trial, or fails to appear in court for trial, or both.
	We clean the dataset to remove incomplete rows, restrict the analysis to defendants who were at least 18 years old, and consider only defendants who were released before trial as we only have ground truth for this group.
  We consider seven attributes as features in this dataset: Gender, Age, Race, Prior Arrests, Prior Convictions, Prior Failure to Appear, and Offense Type (e.g., drug, violent).
	To protect defendant privacy, we only selected defendants whose features are
  identical to at least two other defendants in the dataset.
	This yielded a dataset of 40,551 defendants.
	\item {\compastask} \citep{angwin2016machine}.
  The task is to predict if the defendant will recidivate in two years.
	The features in this dataset are Sex, Age, Race, Prior Crimes, Charge Degree, Juvenile Felony Count, and Juvenile Misdemeanor Count.
	Both datasets have overlapping features such as Age and Race.
	There are 7,214 defendants in this dataset.
	\item {\biostask} \citep{de2019bias}.
  This dataset contains hundreds of thousands of online biographies from the Common Crawl corpus.
  The task is to predict a person's profession given a biography.
	The original dataset consists of 29 professions, and we narrow it down to five professions to make the task feasible for humans, namely, psychologist, physician, surgeon, teacher, and professor.\footnote{%
	To choose these five professions, we built
  maximum spanning trees with 4, 5, 6 nodes from a graph
  based on the confusion matrix of a classifier trained with all biographies.
  Thus, the maximum spanning tree identifies the most confusing professions for the AI.%
  }
	This yielded a dataset of 205,360 biographies.
\end{itemize}

As \citet{bansal2021does} hypothesize that comparable performance between humans and AI is critical for complementary performance, our tasks cover varying performance gaps.
The \ind performance gap between AI alone and human alone \ind is relatively small ($\sim$7\%) in recidivism prediction  (68.4\% vs. 60.9\% in \icpsrtask and 65.5\% vs. 60.0\% in \compastask), but large ($\sim$20\%) in profession prediction (see \tabref{tb:performance_summary} and \secref{sec:performance} for a more detailed discussion on performance gap). 
Note that human performance in ICPSR and COMPAS is derived from our experiments with crowdworkers.
Although they are not representative of judges (see more discussion in \secref{sec:discussion}), they outperform random baselines and can potentially be improved with AI assistance.
In fact, human performance in LSAT is also $\sim$60\% in \citet{bansal2021does}, 
and crowdworkers were able to achieve complementary performance.
Finally, we include gender and race for recidivism prediction to understand how humans might use the information, but they should not be included
in AI 
for deployment.

\subsection{\Ind vs. \Ood Setup}
\label{sec:ind_ood}

As argued in \secref{sec:related}, prior work randomly split a dataset to evaluate the performance of \humanaiteams.
This setup constitutes a best-case scenario for AI performance and may have contributed to the elusiveness of complementary performance.
We expect humans to be more capable of providing complementary insights (e.g., recognizing that AI falsely generalizes a pattern) on examples following different distributions from the training data (\ood).
Therefore, it is crucial to evaluate the performance of \humanaiteams on \ood examples.
We thus provide the first attempt to incorporate distribution shift into experimental studies in the context of human-AI decision making.

\subsubsection{Designing \Ind vs. \Ood}

To simulate the differences between \ind and \ood examples, our strategy is to split the dataset into an \ind (\shortind) subset and an \ood (\shortood) subset based on a single attribute (e.g., age $\geq$ 25 as \ind and age $<$ 25 as \ood to simulate a scenario where young adults are not presented in the training set).
We develop the following desiderata for selecting an attribute to split the dataset:
1) splitting by this attribute is sensible and interpretable to human (e.g., it makes little sense to split biographies based on the number of punctuation marks);
2) splitting by this attribute could yield ~a difference in AI performance between \ind and \ood so that we might expect different human behavior in different \domain types;
3) this attribute is ``smoothly'' distributed in the dataset 
to avoid extreme distributions that can limit
plausible ways to simulate \shortind and \shortood examples (see the supplementary materials for details).
Now we discuss the attribute selected for each dataset and present rationales for not using other attributes.

\begin{figure}[t]
  \centering
    \centering
    \includegraphics[width=.32\textwidth]{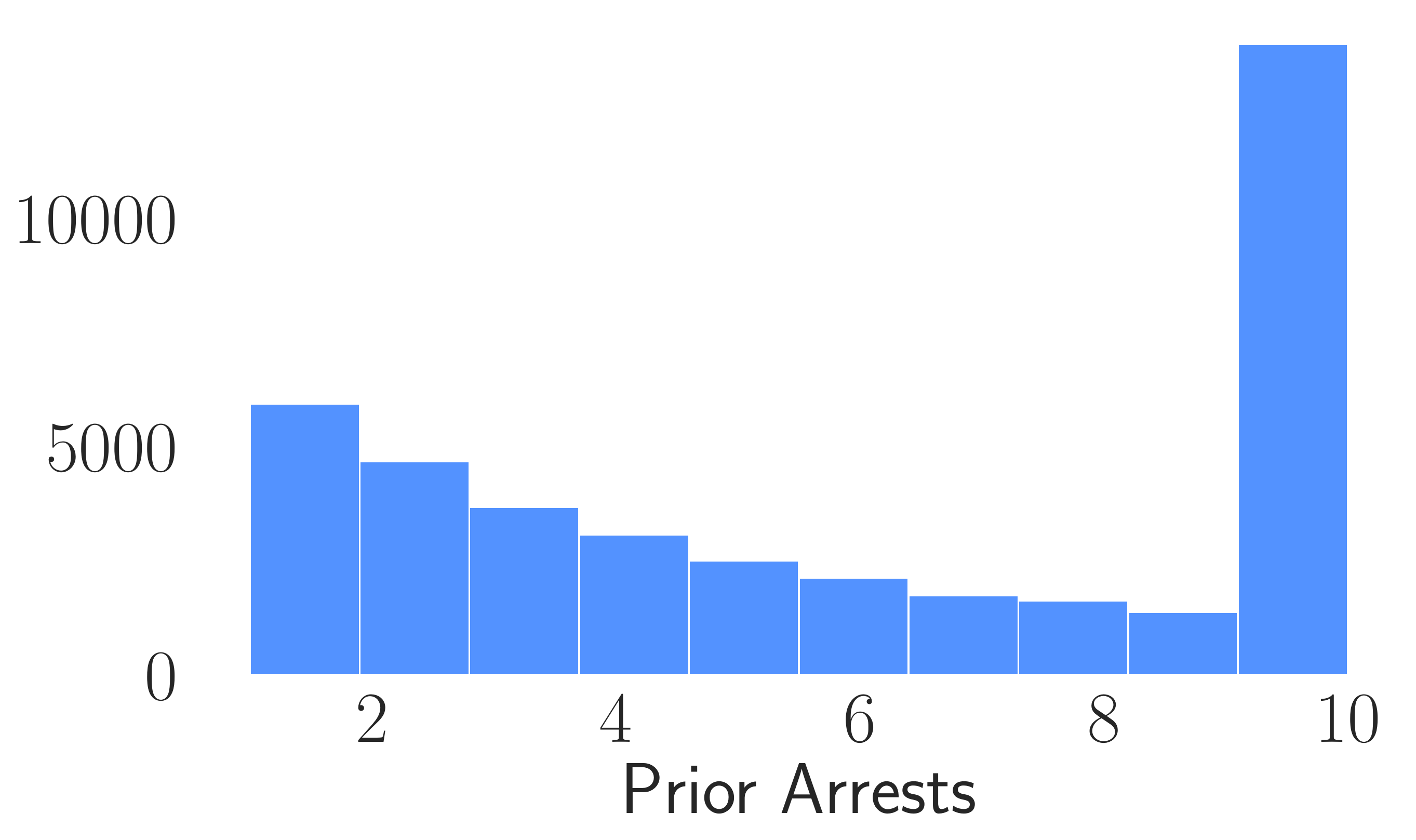}
    \includegraphics[width=.32\textwidth]{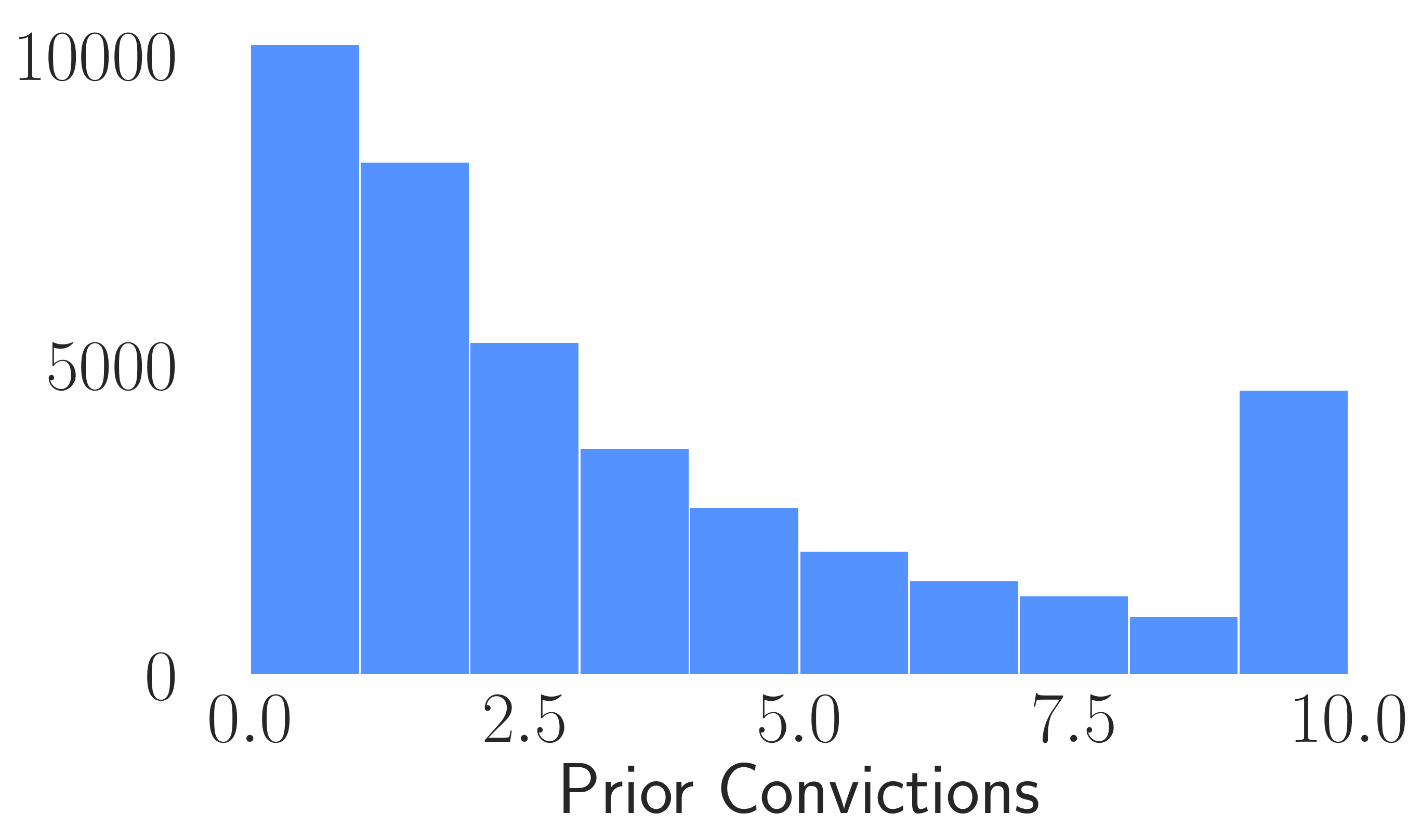}
  \caption{Histograms of numbers of instances for ``Prior Arrests'' and ``Prior Convictions'' in \icpsrtask.}
  \label{fig:icpsr_setup}
  \Description{Numbers of instances for ``Prior Arrests'' 0-9 decrease logarithmicly from around 5000 to around 1000, while the number of instances for ``Prior Arrests'' 10 is more than 12000. Numbers of instances for ``Prior Convictions'' 0-9 decrease logarithmicly from around 10000 to around 1000, while the number of instances for ``Prior Convictions'' 10 is around 5000.}
\end{figure}

\begin{itemize}[topsep=0pt,leftmargin=*]
	\item \icpsrtask.
    We choose the age of the defendant as the attribute. We also tried Gender, but it failed 
    desiderata 2 due to a small AI performance  difference (1\%) between \ind and \ood. Other features such as Prior Arrests and Prior Convictions do not satisfy desiderata 3, because they have a huge spike towards the end (see \figref{fig:icpsr_setup})
    and thus limit possible \shortind/\shortood splits.
  \item \compastask.
    We choose the age of the defendant as the attribute. We also tried Sex and Prior Crimes, but they failed desiderata 2 and 3 respectively as
    Gender and Prior Convictions did in \icpsrtask.
  \item \biostask.
    We choose the length of the biography (i.e., the total number of characters) as the attribute. 
    Note that our dataset contains biographies from the web, a dataset created by \citet{de2019bias}.
    Although one may think that professor, surgeon, psychologist, and physician require more education than teacher and thus resulting in longer biographies, the average biography length of a teacher's biography is not the shortest in our dataset.
Interestingly, physicians have the shortest biographies with 348 characters and teachers have an average biography length of 367 characters.
    We also experimented with gender but it does not satisfy desiderata 2 since we
    observed a small AI performance difference (3\%) between \ind and \ood.
\end{itemize}

Given the selected attribute, for each dataset, we split the data into 10 bins of equal size based on the attribute of choice.
Then, we investigate which bins to use as \ind and 
\ood. Our goal in this step is to maximize the AI performance gap between \ind and \ood so that we can observe whether humans would behave differently with AI assistance depending on \domain types (see supplementary materials).
The chosen splits for each dataset are:
1) age $\geq$ 25 as \shortind and age $<$ 25 as \shortood in \icpsrtask,
2) age $\geq$ 26 as \shortind and age $<$ 26 as \shortood in \compastask, and
3) length $\geq$ 281 characters as \shortind and length $<$ 281 characters as \shortood in \biostask.
For each potential split, we use 70\% of the data in the \shortind bins for training and 10\% of the data in the \shortind bins for validation.
Our test set includes two subsets: 1) the remaining 20\% of the data in the \shortind bins, and 2) the data in the \shortood bins. We also balance the labels in each bin of our test set for performance evaluation.

\begin{figure}
  \centering
  \includegraphics[width=0.5\textwidth]{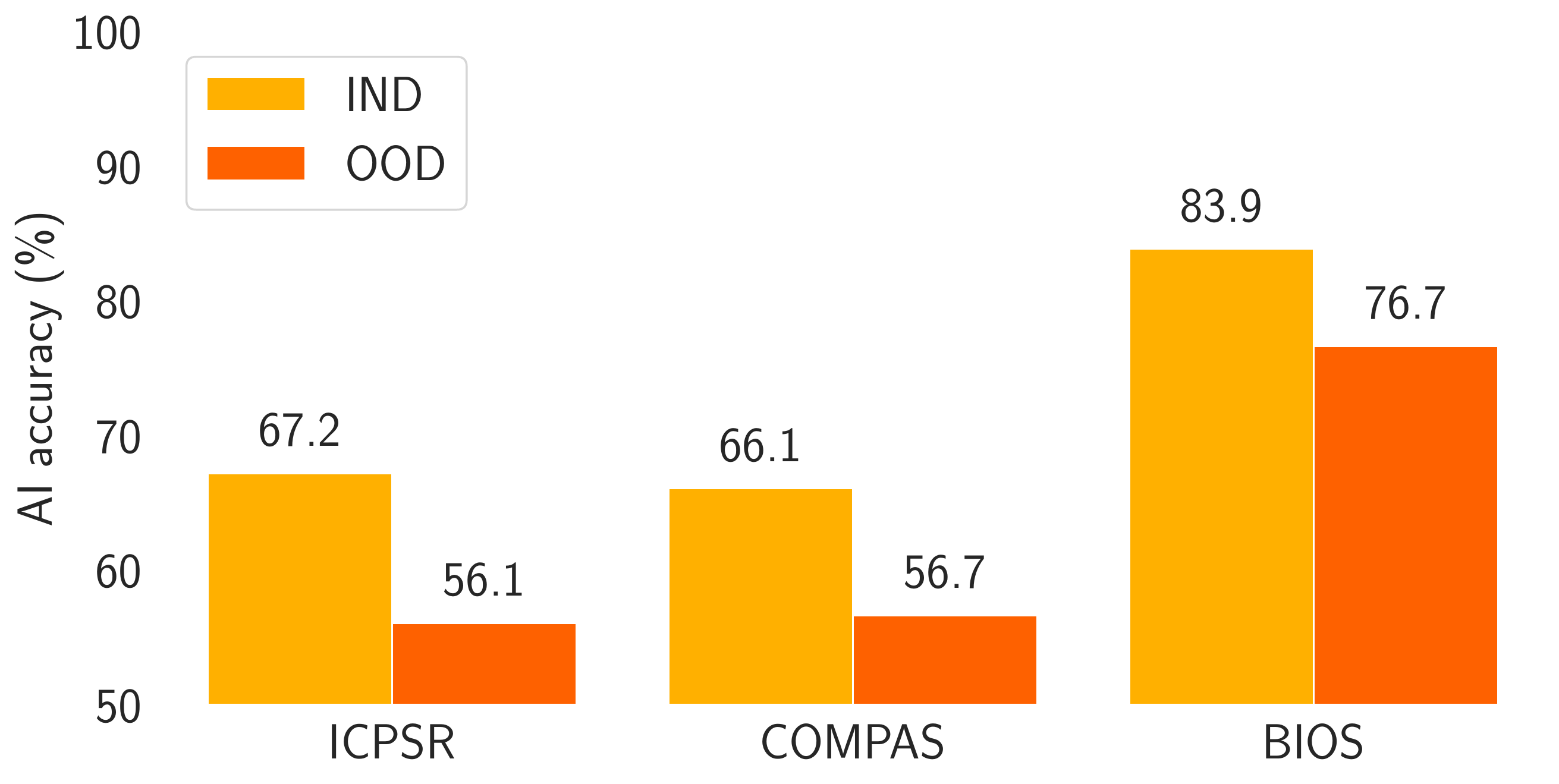}
  \Description{}
  \caption{Accuracy of machine learning models on the \ind and \ood test set 
  for the user study. Since the test set is balanced, the baseline in \icpsrtask and \compastask is 50\%. AI outperforms the random baseline even \ood in \icpsrtask and \compastask despite that its performance is lower \ood than \ind.
  AI performance drops by about 10\% in recidivism prediction and about 7\% in \biostask~for \ood examples compared to \ind examples.
  }
  \label{fig:ai_acc}
\end{figure}

\subsubsection{AI Performance \ind and \ood.}
Following prior work \cite{lai+liu+tan:20,de2019bias}, we use a linear SVM classifier with unigram bag-of-words
for \biostask
and
with one-hot encoded features for recidivism prediction tasks.
The standard procedure of hyperparameter selection (a logarithmic scale between $10^{-4}$ and $10^{4}$ for the inverse of regularization strength) is done with the validation set.
We focus on linear models in this work for three reasons:
1) linear models are easier to explain than deep models and are a good starting point to develop interactive explanations \citep{feng2019can,poursabzi2021manipulating};
2) prior work has shown that human performance is better when explanations from simple models are shown \cite{lai+liu+tan:20};
3) there is a sizable performance gap between humans and AI even with a linear model, although smaller than the case of deception detection \citep{lai+tan:19,lai+liu+tan:20}.
Finally, to reduce the variance of human performance so that each example receives multiple human evaluations,
we
randomly sample 180 IND examples and 180 OOD examples from the test set to create a balanced
pool for our final user study.\footnote{We choose from five random seeds the one that leads to the greatest AI performance difference between \ind samples and \ood samples.}
\figref{fig:ai_acc} shows AI performance on these samples:
the \shortind-\shortood gap is about 10\% in recidivism prediction and 7\% in \biostask.
It entails that the absolute performance necessary to achieve complementary performance is lower \shortood than \shortind.
Because of this AI performance gap \ind and \ood, we will focus on understanding the performance difference between \humanaiteams and AI alone ({\em accuracy gain}).
As discussed in \secref{sec:related}, we hypothesize that the accuracy gain is greater \ood than \ind.

\subsection{Interactive Explanations and Explanation Type}
\label{sec:interactive_conditions}

\begin{figure}
  \centering
  \includegraphics[width=0.5\textwidth]{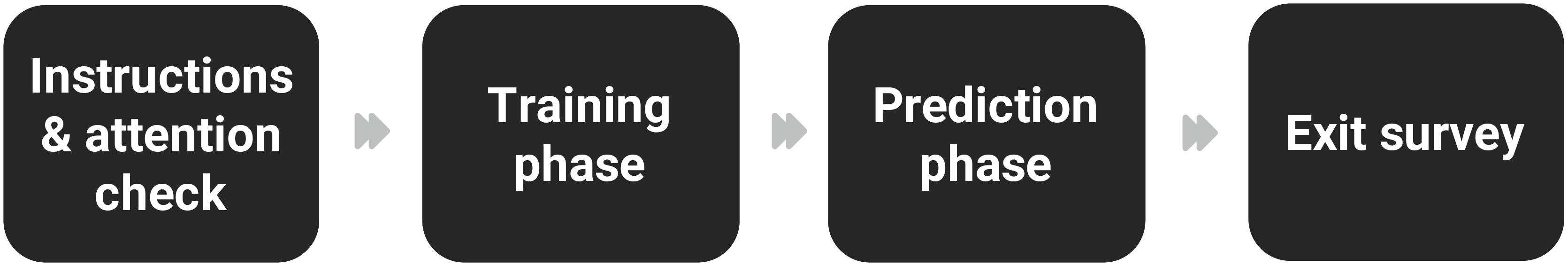}
  \Description{The workflow contains four stages: 1) instructions and attention check; 2) training phase; 3) prediction phase; 4) exit survey.}
  \caption{The workflow of our experiments. In the training phase, we introduce a novel feature quiz where users choose one positive and one negative feature after each example. Human decisions in the prediction phase are used to study human-AI decision making.}
  \label{fig:exp_workflow}
\end{figure}

To help users understand the patterns embedded in machine learning models,
following \citet{lai+liu+tan:20},
our experiments include two phases: 1) a training phase where users are shown no more than six representative examples and the associated explanations; and 2) a prediction phrase that is used to evaluate the performance of \humanaiteams with 10 random \ind examples and 10 random \ood examples.
\figref{fig:exp_workflow} shows the workflow of our experiments.
Our contribution is to develop interactive explanations to enable a two-way conversation between humans and AI and examine the effect of interactive explanations.
We also consider a static version of AI assistance in each phase for comparison.
We refer to AI assistance during the prediction phase as {\em real-time assistance}.

\subsubsection{Static Assistance}
\begin{figure}[t]
  \centering
  \begin{subfigure}{0.48\textwidth}
    \includegraphics[width=\textwidth]{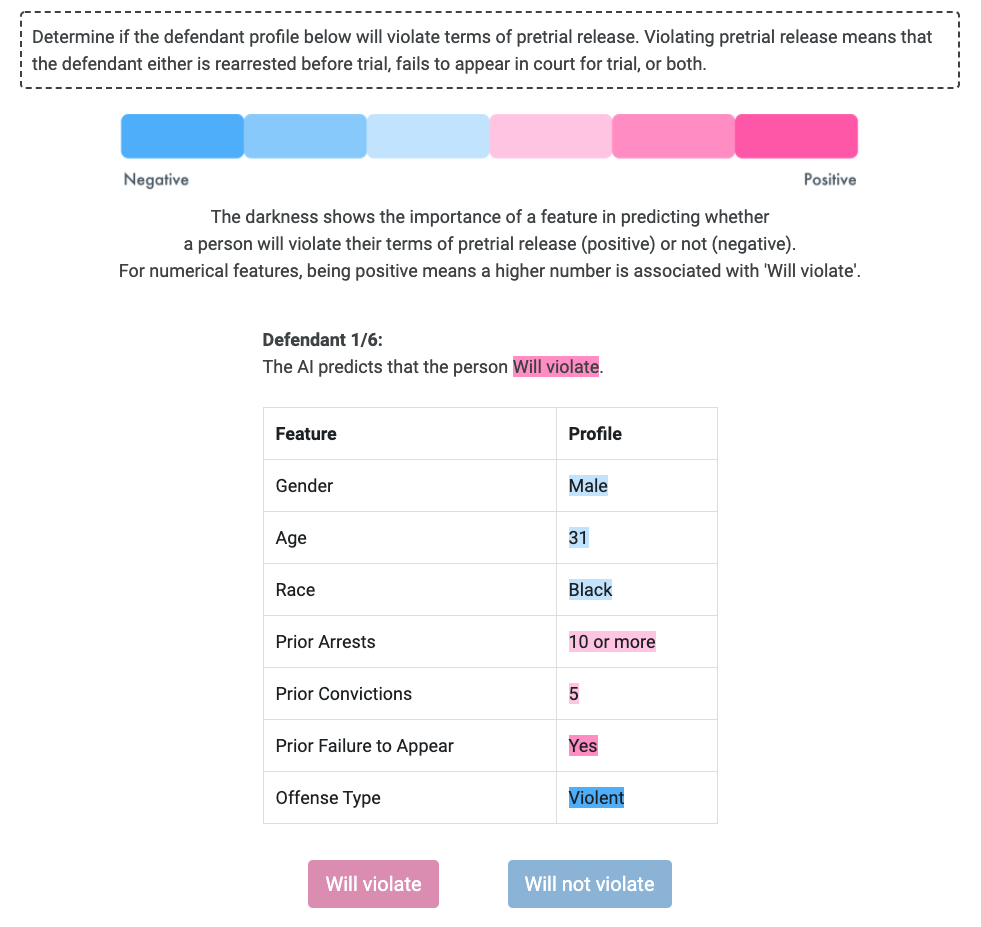}
    \caption{Static assistance for \icpsrtask.}
    \label{fig:icpsr_static}
    \Description{Defendant profile with features highlighted in three shades of blue and three shades of pink.}
  \end{subfigure}
  \hfill
  \begin{subfigure}{0.48\textwidth}
    \includegraphics[width=\textwidth]{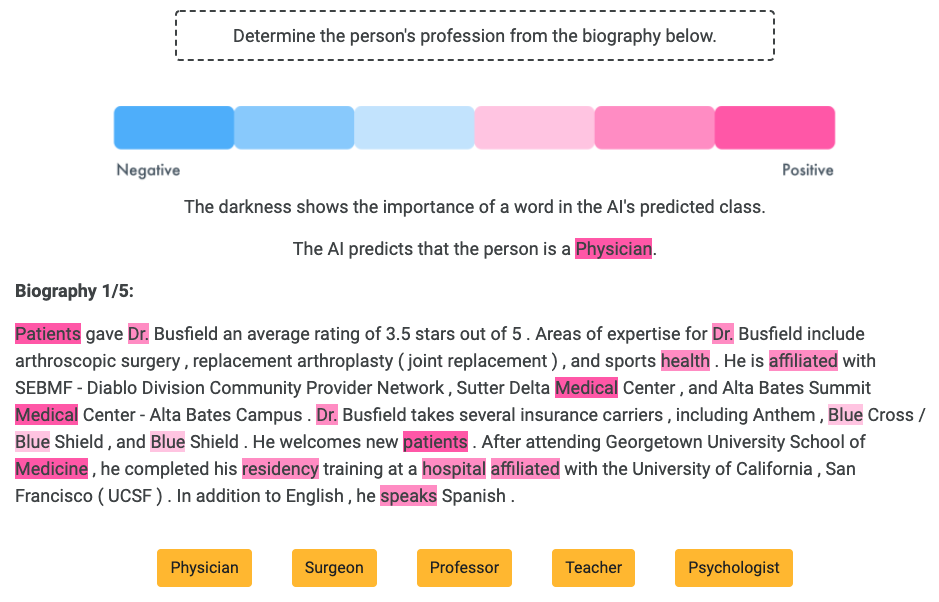}
    \caption{Static assistance for \biostask.}
    \label{fig:bios_static}
    \Description{Biography texts with words highlighted in three shades of blue and three shades of pink.}
  \end{subfigure}
  \caption{Screenshots for static assistance in \icpsrtask and \biostask. The interface for \compastask is similar to \icpsrtask (see \figref{fig:compas_static}.}
  \label{fig:static_assistance}
\end{figure}

Our static assistance for an AI prediction includes two components (see \figref{fig:static_assistance}).
First, we highlight important features based on the absolute value of feature coefficients to help users understand what factors determine the AI prediction.
We color all seven features in \icpsrtask and \compastask to indicate whether a feature contributes positively or negatively to the prediction (\figref{fig:icpsr_static}).
As \biostask has many words as features, we highlight the top 10 most important words.
We only show the colors but hide the feature coefficient numbers because 1) we have not introduced the notion of prediction score; 2) showing numerical values without interaction may increase the cognitive burden without much gain.
Second, we also show the AI predicted label along with the highlights.
In the training phase, following \citet{lai+liu+tan:20}, ~the actual label is revealed after users make their predictions so that they can reflect on their decisions and actively think about the task at hand.

The purpose of the training examples is to allow participants to familiarize themselves with the task, extract useful and insightful patterns, and apply them during the prediction phase.
We use
SP-LIME \cite{ribeiro2016should,lai+liu+tan:20} to identify 5-6 representative training examples that capture important features (6 in \icpsrtask and \compastask and 5 in \biostask).\footnote{We include 10 examples in the pilot studies, but mechanical turkers commented that the experiment took too long.}
We 
make sure the selected examples are balanced across classes.
For the control condition, we simply include the first two examples.
Finally, during training, to ensure that users understand the highlighted important features, we add a feature 
quiz after each example where users are required to choose a positive and a negative feature (see \figref{fig:feature_quiz}).

\begin{figure}[t]
  \centering
  \begin{subfigure}{0.48\textwidth}
    \includegraphics[width=\textwidth,page=2,trim=80 0 100 0,clip]{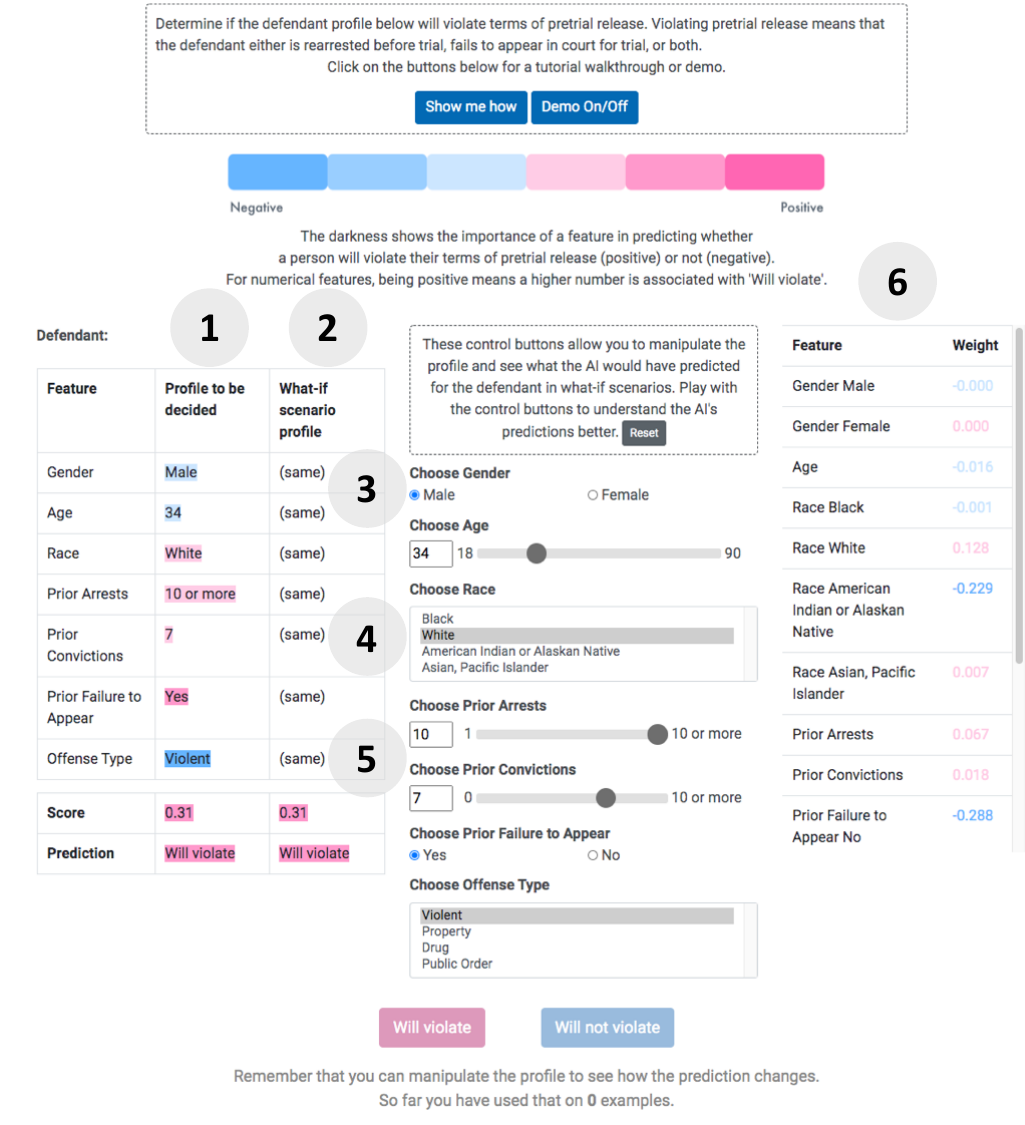}
    \caption{Interactive explanation for \icpsrtask.}
    \label{fig:icpsr_interactive}
    \Description{Picture shows an interactive console that allows user to manipulate feature values with sliders and radio buttons. Changes in feature values may result in a change of the AI prediction. A table on the right of the picture displays feature coefficients.}
  \end{subfigure}
  \hfill
  \begin{subfigure}{0.48\textwidth}
    \includegraphics[width=\textwidth,page=3,trim=30 0 20 0,clip]{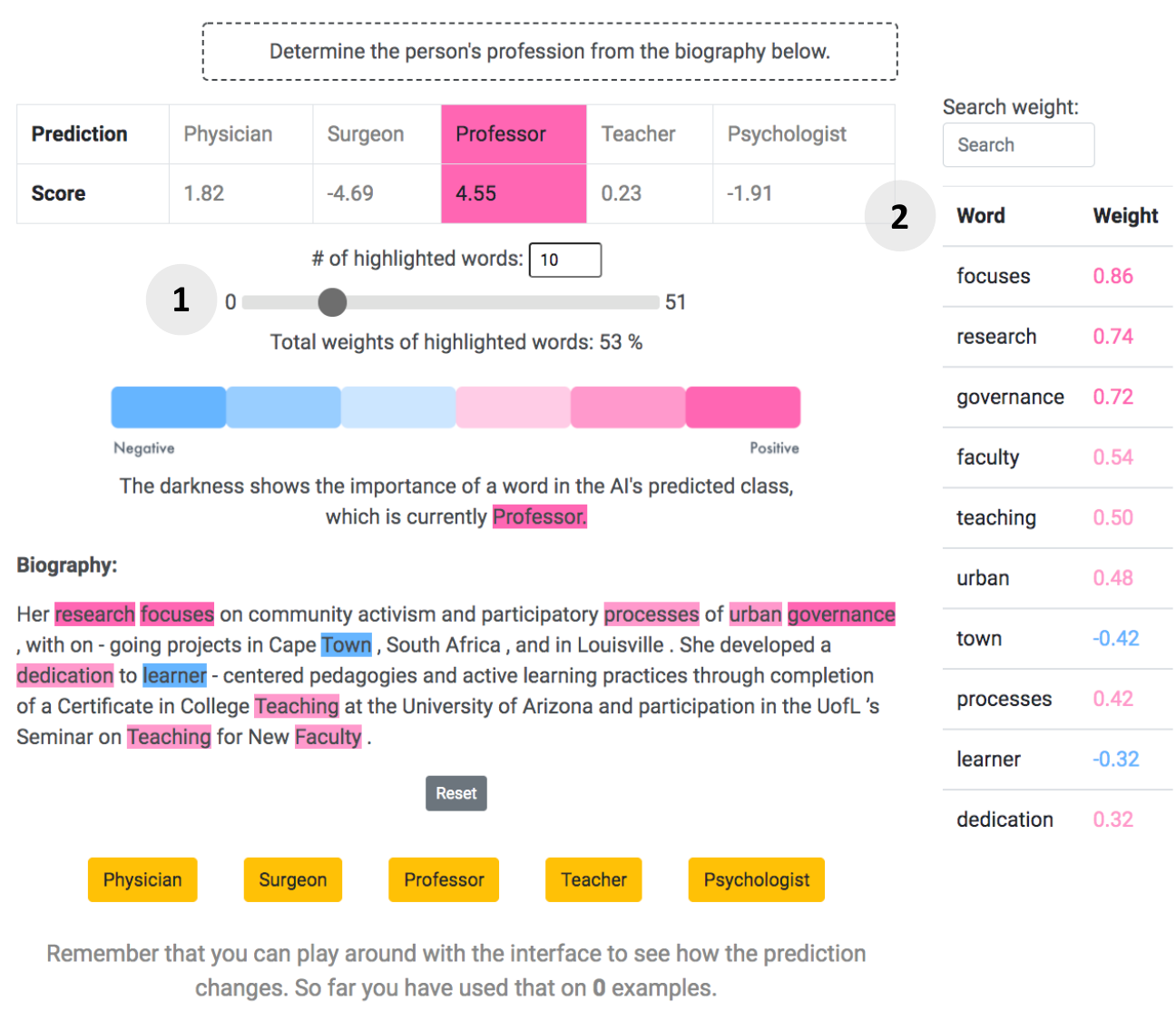}
    \caption{Interactive explanation for \biostask.}
    \label{fig:bios_interactive}
    \Description{Picture shows an interactive console that allows user to manipulate the number of highlighted words with a slider and an input field. A table on the right of the picture displays feature coefficients of words and a search box.}
  \end{subfigure}  
  \caption{Screenshots for interactive explanations in \icpsrtask and \biostask. 
  In addition to static assistance such as feature highlights and showing AI predictions, users are able to manipulate the features of a defendant's profile to see any changes in the AI prediction in \icpsrtask. The interactive console for \icpsrtask includes: 1) the actual defendant's profile; 2) the edited defendant's profile if user manipulates any features; 3) users are able to edit the value of {\em Gender} and {\em Prior Failure to Appear} with radio buttons; 4) users are able to edit the value of {\em Race} and {\em Offense Type} with dropdown; 5) users are able to edit the value {\em Age}, {\em Prior Arrests}, and {\em Prior Convictions} with sliders; 6) a table displaying features and coefficients, the color and darkness of the color shows the feature importance in predicting whether a person will violate their terms of pretrial release or not.
  In \biostask, users are able to remove any words from the biography to see any changes in the AI prediction. The interactive console for \biostask includes: 1) user is able to edit the number of highlighted words with a slider; 2) a table displaying features and respective coefficients, the color and darkness of the color shows the importance of a word in the AI's predicted class. The interface for \compastask is similar to \icpsrtask (see \figref{fig:compas_screenshot}).
  }
  \label{fig:interactive_screenshot}
\end{figure}

\subsubsection{Interactive Explanations}
To help humans better understand how AI makes a prediction and the potential fallacies in AI reasoning, we develop a suite of interactive experiences.
There are two important components.
First, we enable users to experiment with counterfactual examples of a given instance.
This allows participants to interact with each feature and observe changes in AI predictions.
Second, we make the feature highlights dynamic, especially for \biostask where there are many features.
Specifically, our designs are as follows:

\begin{itemize}[topsep=0pt,leftmargin=*]

  \item Interactive explanations for tabular-data classification (\icpsrtask and \compastask; \figref{fig:icpsr_interactive} gives a screenshot for \icpsrtask).
  We present the original profile of the defendant and the counterfactual (``What-if scenario profile'') on the left of the screen (\figref{fig:icpsr_interactive}(1)).
  Users can adjust features to change the counterfactual profile (\figref{fig:icpsr_interactive}(2))
  via sliders, radio buttons, and select lists (\figref{fig:icpsr_interactive}(3-5)).
  For instance, users can investigate how a younger or older age affects the prediction by adjusting a defendant's age using the slider.
  In addition, we show all the features and their associated weight on the  right, sorted in descending order (\figref{fig:icpsr_interactive}(6)).
  \item Interactive explanations for text classification (\biostask; see \figref{fig:bios_interactive}).
  To enable the counterfactuals, users can delete any word in the text and see how the prediction would change (removal can be undone by clicking the same word again).
  For dynamic highlight, 
   a slider is available for users to adjust the number of highlighted words (\figref{fig:bios_interactive}(1)). 
   In addition, we provide a searchable table to display all words presented in the text and their associated feature importance, sorted in descending order (\figref{fig:bios_interactive}(2)).
\end{itemize}

The searchable table allows users to the explore the high-dimensional feature space in \biostask, a text classification task.
While it may seem that showing coefficients in recidivism prediction is not as useful, we highlight that these numerical values make little sense on their own.
The counterfactual profile enables users to examine how these numerical values affect prediction outcomes.

\subsection{Virtual Pilot Studies}
\label{sec:qual}

We conducted virtual pilot studies to obtain a qualitative understanding of human interaction with interactive explanations.
The pilot studies allow us to gather insights on how humans 
use interactive explanations 
in their decision-making process, as well as feedback on the web application before conducting large-scale randomized experiments.

\para{Experimental design.}
We employed a concurrent think-aloud process with participants \citep{nielsen2002getting}.
Participants are told to verbalize the factors they considered 
behind a prediction.
During the user study session, participants first read instructions for the task and proceed to answer a couple of attention-check questions (see \figref{fig:attention}), which ensure that they 
understand the purpose of the user study.
Upon passing the attention-check stage, they undergo a training phase 
before proceeding to the prediction phase.
Finally, they answer an exit survey (see \figref{fig:survey}) that asks for demographic information and semi-structured questions on the web application and interactive explanations.
A participant works on \icpsrtask and \biostask in a random order.

We recruited 15 participants through mailing lists at the University of Colorado Boulder:
7 were female and 8 were male, with ages ranging from 18 to 40.\footnote{Note that the wide range in age is due to the available choices in our exit survey.
Namely, the first option is 18-25 and the second option is 26-40.
}
To understand the general population that does not have a machine learning background, we sent out emails to computer science and interdisciplinary programs.
Participants included both undergraduate and graduate students with and without machine learning background.
The user study is conducted on Zoom due to the pandemic.
The user study sessions were recorded with the participants' consent.
Participants were compensated for \$10 for every 30 minutes. 
A typical user study session lasted between an hour to an hour and a half.
Participants were assigned in a round-robin manner to interactive and static explanations. 
For instance, if a participant was assigned to 
static explanations in \biostask, the participant would be assigned to 
interactive explanations 
in \icpsrtask.
As the user study sessions were recorded on Zoom cloud, we used the first-hand transcription provided by Zoom and did a second round of transcribing to correct any mistranscriptions.
Subsequently, thematic analysis was conducted to identify common themes in the think-aloud processes, and thematic codes were collectively coded by two researchers.

Next, we summarize the key themes from the pilot studies and the changes to our interface.

\para{Disagreement with AI predictions.}
Participants tend to disagree with AI predictions when the explanations provided by the AI contradict their intuitions.
For instance, 
although AI suggests that the drug offense type is correlated with ``Will violate'', P4 thinks that ``drug offense is not something serious, a minor offense'' and thus disagrees with AI and chooses ``Will not violate''.
With a similar train of thought, P7 asks why 
AI suggests the violent offense type to be correlated with ``Will not violate''
 and thinks that it should be the other way around.
A potential reason is that people are more likely to restrain themselves after serious crimes as the consequence can be dire, but it seemed difficult for the participants to reason about this counterintuitive pattern.
The above comments suggest that some patterns that AI identifies can be counterintuitive and thus challenging for humans to make sense of.

Furthermore, participants disagree with AI predictions 
due to focusing too much on a few patterns they learned from AI.
For instance, if a participant learns that {\em Prior Failure to Appear} positively 
relates to {``Will violate''}, they will apply the same logic on future examples and disagree with the AI when the pattern and prediction disagrees.
Quoting from P9, ``The current example has no for {\em Prior Failure to Appear} and drug offense but the previous examples had yes for {\em Prior Failure to Appear} and drug offense''. P9 then chooses ``Will not violate'' because of these two features.
This observation highlights the importance of paying attention to features globally, which can be challenging for humans.

Finally, 
participants are more confident 
in \biostask than in \icpsrtask as they are able to relate to the task better and understand the explanations provided by the AI better.
They believe that the biography text is sufficient to detect the profession,
but much of the crucial information is missing in \icpsrtask. 
P9 said, ``there was more background on what they did in their lives, and how they got there and whatnot, so it helped me make a more educated decision''.
This observation also extends to their evaluation of AI predictions,
quoting from P12, ``the AI would be more capable of predicting based on a short snippet about someone than predicting something that hasn't happened''.

\para{Strategies in different tasks.}
Different strategies are employed in different tasks.
Since \biostask is a task requiring participants to read a text, most participants look for (highlighted) keywords that distinguish similar professions.
For instance, while both professor and teacher teach, participants look for keywords such as ``phd'' to distinguish them.
Similarly, in the case of surgeon and physician, participants look for keywords such as ``practice'' and ``surgery''.
In \icpsrtask, as there are only seven features, most participants pay extra attention to a few of them, including {\em Prior Failure to Appear}, {\em Prior Convictions}, {\em Prior Arrest}, and {\em Offense Type}.
We also noticed during the interview that most participants tend to avoid discussing or mentioning sensitive features such as {\em Race}.
In \secref{sec:discussion}, we elaborate and discuss findings on an exploratory study on important features identified by participants.

\para{The effect of interactive explanations.}
Participants could be categorized into two groups according to their use of the interactive console, either they do not experiment with it, or they play with it excessively.
Participants in the former group 
interact with the console only when prompted, while the latter group 
result in a prolonged user study session.
Some participants find the additional value of interactive console limited  as compared to static explanations such as highlights.
They are unsure of the `right' way to use it as P12 commented, ``I know how it works, but I don't know what I should do. Maybe a few use cases can be helpful. Like examples of how to use them''.
Other participants do not interact much with it, but still think it is helpful.
With reference to P6, ``I only played with it in the first few examples. I just use them to see the AI's decision boundaries. Once I get it in training, I don't need them when I predict.''

Another interesting finding was that while some participants make decisions due to visual factors, others make decisions due to numerical factors.
P2 said, ``the color and different darkness were really helpful instead of just having numbers''. 
In contrast, P4, who often made decisions by looking at the numbers, commented on one of the many justifications that the defendant ``will not violate because the numbers are low.''
This observation suggests that our dynamic highlights may provide extra benefits to static highlights.

\para{Web application feedback.}
As some participants were unsure of how to use the interactive console and make the most out of it,
we added an animated video that showcased an example of using the interactive console on top of the walk-through tutorial that guides a user through each interactive element (see the supplementary materials).
We also added a nudging component describing how many instances they have used interactive explanations with to remind participants of using the interactive console (see \figref{fig:interactive_screenshot}).

In addition to Zoom sessions, we conducted pilot studies on Mechanical Turk before deploying them in large-scale tasks.
Since some Zoom sessions took longer than we expected, we wanted to investigate the total time taken for completing 10 training and 20 test instances.
We noted from the feedback collected from exit surveys of pilot studies that the training was too time consuming and difficult.
We thus reduced the number of training instances
and improved the attention check questions and instruction interfaces.
See the supplementary materials for details.

\subsection{Large-scale Experiments with Crowdworkers}
\label{sec:mturk}

Finally, we discuss our setup for the large-scale experiments on Amazon Mechanical Turk. First, in order to understand the effect of \ood examples, we consider the performance of humans without any assistance as our control setting.
Second, another focus of our study is on interactive explanations, we thus compares interactive explanations and static explanations.\footnote{A natural question is about the effect of explanations vs. AI assistance without explanations. We refer readers to prior work on this question \citep{lai+liu+tan:20,lai+tan:19,green2019principles}.}

Specifically, participants first go through a training phase to understand the patterned embedded in machine learning models, and then enter the prediction phase where we evaluate the performance of \humanaiteams. We allow different interfaces in the training phase and in the prediction phase because the ideal outcome is that participants can achieve complementary performance without real-time assistance after the training phase.
To avoid scenarios where users experience a completely new interface during prediction,
we consider situations where the assistance in training is more elaborate than the real-time assistance in prediction.
Therefore, we consider the following six conditions to understand the effect of explanation types during training and prediction (the word before and after ``/'' refers to the assistance type during training and prediction respectively):
\begin{itemize}[topsep=0pt,leftmargin=*]
	\item \textbf{None/None.} 
  Participants are not given any form of AI assistance in either the training phase or the prediction phase. In the training phase, there are only two examples instead of 5-6 in other conditions to help participants understand the task. In other words, this condition is a {\em human-only} condition.
	\item \textbf{Static/None.}
  Participants are provided static assistance in the training phase.
  Important features are highlighted in shades of pink/blue ~and AI predictions are provided.
	Participants
  are {\em not} provided any assistance in the prediction phase.
	\item \textbf{Static/Static.} Participants are provided static assistance
  in both training and prediction.
	\item \textbf{Interactive/None.} Participants are provided interactive explanations during the training phase, and no assistance in the prediction phase.
	\item \textbf{Interactive/Static.} Participants are provided interactive explanations in the training phase and static assistance  in the prediction phase.
	\item \textbf{Interactive/Interactive.} Participants are provided interactive explanations in both training and prediction.
\end{itemize}

We refer to these different conditions as \textit{\textbf{explanation type}} in the rest of this paper.
The representative examples are the same during training in
Interactive and Static.
Participants are recruited via Amazon Mechanical Turk and must satisfy three criteria to work on the task: 1) residing in the United States, 2) have completed at least 50 Human Intelligence Tasks (HITs), and 3) have been approved for 99\% of the HITs completed.
Following the evaluation protocol in prior work \citep{green2019disparate,green2019principles}, 
each participant is randomly assigned to one of the explanation types, and their performance is evaluated on 10 random \ind examples and 10 random \ood examples.
We do not allow any repeated participation.
We used the software program G*Power to conduct a power analysis.
Our goal was to obtain .95 power to detect a small effect size of .1 at the standard .01 alpha error probability using F-tests.
As such, we employed 216 participants for each explanation type, which adds up to 1,296 participants per task.
Note that our setup allows us to examine human performance on random samples beyond a fixed set of 20 examples, which alleviates the concern that our findings only hold on a dataset of 20 instances.

The median time taken to complete a HIT is 9 minutes and 22 seconds.
Participants exposed to interactive conditions took 12 minutes, while participants exposed to non-interactive conditions took 7 minutes (see \figref{fig:time_taken}).
Our focus in this work is on human performance,
so we did not limit the amount of time in the experiments. Participants were allowed to spend as much time as they needed so that they were able to explore the full capacities of our interface. 
Participants were paid an average wage of \$11.31 per hour.
We leave consideration of efficiency (i.e., maintaining good performance while reducing duration of interactions) for future work.

\section{RQ1: The Effect of \Ind and \Ood Examples on Human Performance}
\label{sec:performance}

\begin{table}
  \begin{tabular}{lrr>{\raggedright\arraybackslash}p{1.5cm}rr>{\raggedright\arraybackslash}p{1.5cm}}
      \toprule
      \multirow{2}{*}{Task} &
        \multicolumn{3}{c}{IND (typical setup)} &
        \multicolumn{3}{c}{OOD (proposed setup)} \\
        & Human & AI & {Difference between humans and AI} & Human & AI & Difference between humans and AI \\ 
        \midrule
        \icpsrtask & 60.9 & 68.4 & $-$7.5 & 55.9 & 55.0 & 0.9 \\ 
        \compastask & 60.0 & 65.5 & $-$5.5 & 54.5 & 56.1 & $-$1.6 \\
        \biostask & 63.5 & 84.1 & $-$20.6 & 68.4 & 76.6 & $-$8.2 \\ 
        \midrule
        Deception detection \citep{lai+liu+tan:20,lai+tan:19} & $\sim$51 & $\sim$87.0 & $\sim-$36 & --- & --- & --- \\
        LSAT \citep{bansal2021does} & $\sim$58 & 65 & $\sim-$7 & --- & --- & --- \\
        Beer reviews \citep{bansal2021does} & $\sim$82 & 84 & $\sim-$2 & --- & --- & --- \\
        \bottomrule
  \end{tabular}
  \medskip
  \caption{
  Performance comparison between human alone and AI alone.
  We also add numbers from prior work to contextualize these numbers.
  Note that AI performance here is slightly different ($\leq$1.2\%) from that in \figref{fig:ai_acc}, because AI performance in this table is calculated from a subset of examples shown in None/None (human alone) while the AI performance in \figref{fig:ai_acc} is calculated from the \ood test set of 180 examples.
  }
  \label{tb:performance_summary}
\end{table}

Our first research question examines how \ind and \ood examples affect the performance of \humanaiteams.
Recall that  \citet{bansal2021does} hypothesize that 
comparable performance is important to achieve complementary performance.
\tabref{tb:performance_summary} compares the performance of human alone and AI alone in the three prediction tasks both \ind and \ood (we also add tasks from other papers to illustrate the ranges in prior work). 
The performance gap between human alone and AI alone in \icpsrtask and \compastask is similar to tasks considered in \citet{bansal2021does}.
In \biostask, the \ind performance gap between human alone and AI alone is greater than the tasks in \citet{bansal2021does} but much smaller than deception detection, and the \ood performance gap between human alone and AI alone becomes similar to LSAT in \citet{bansal2021does}.
As a result, we believe that our chosen tasks somewhat satisfy the condition of ``comparable performance'' and allow us to study human-AI decision making over a variety of performance gaps between human alone and AI alone.

Note that AI performance here is calculated from the random samples shown in None/None (human alone), and is thus slightly different ($\leq$1.2\%) from AI performance in \figref{fig:ai_acc}, which is calculated from the \ind and \ood test set of 180 examples each.
To account for this sample randomness and compare human performance in different explanation types for these two \domain types, we need to establish a baseline given the random samples
(we show absolute accuracy in the supplementary material as the performance difference without accounting for the baseline is misleading; see \figref{fig:perf_norm}).
Therefore, we calculate the accuracy difference on the same examples between a \humanaiteam and AI, and use {\em accuracy gain} as our main metric.
Accuracy gain is positive if a \humanaiteam outperforms AI.
In the rest of this paper, we will use {\em human performance} and {\em the performance of \humanaiteams} interchangeably.
Since the results are similar between \icpsrtask and \compastask, we show the results for \icpsrtask in the main paper and include the figures for \compastask in the supplementary materials (see \figref{fig:compas_perf}-\figref{fig:compas_top_features}).

\para{Preview of results.} To facilitate the understanding of our complex results across tasks, we provide a preview of results before unpacking the details of each analysis.
Our results indeed replicate existing findings that AI performs better than \humanaiteams in \ind examples. However, \humanaiteams fail to outperform AI in \ood examples.
The silver lining is that the performance gap between \humanaiteams and AI is smaller \ood than \ind.
These results are robust across tasks (see Table~\ref{tb:preview-performance} for a summary).

\begin{table}[t]
  \centering
  \begin{tabular}{p{0.4\linewidth}|c@{\hspace{4pt}}c@{\hspace{4pt}}c|c@{\hspace{4pt}}c@{\hspace{4pt}}c}
      \toprule
      \multirow{2}{*}{} &
          \multicolumn{3}{c|}{IND (typical setup)} &
          \multicolumn{3}{c}{OOD (proposed setup)} \\
          & \icpsrtask & \compastask & \biostask & \icpsrtask & \compastask & \biostask \\ \midrule

       AI performs better than \humanaiteams in \ind examples. & \cmark & \cmark & \cmark & --- & --- & --- \\ \midrule
       \Humanaiteams perform better than AI in \ood examples. & --- & --- & --- & \xmark & \xmark & \xmark \\ \midrule
       The performance difference between \humanaiteams and AI is smaller \ood than \ind. & \multicolumn{3}{c|}{\textcolor{gray}{see the \shortood columns}} & \hcmark & \cmark & \cmark \\ \midrule

      \multicolumn{1}{l}{\cmark: ~holds} & \multicolumn{6}{l}{\hcmark: holds in at least half of the explanation types} \\
      \multicolumn{1}{l}{\xmark: ~rejected} & \multicolumn{6}{l}{\hxmark: rejected in all except one explanation type} \\ 
      \bottomrule
  \end{tabular}
  \medskip
  \caption{Summary of results on human-AI team performance.
  }
  \label{tb:preview-performance}
\end{table}

\begin{figure}[t]
    \centering
    \begin{subfigure}{0.48\textwidth}
        \includegraphics[width=0.9\textwidth]{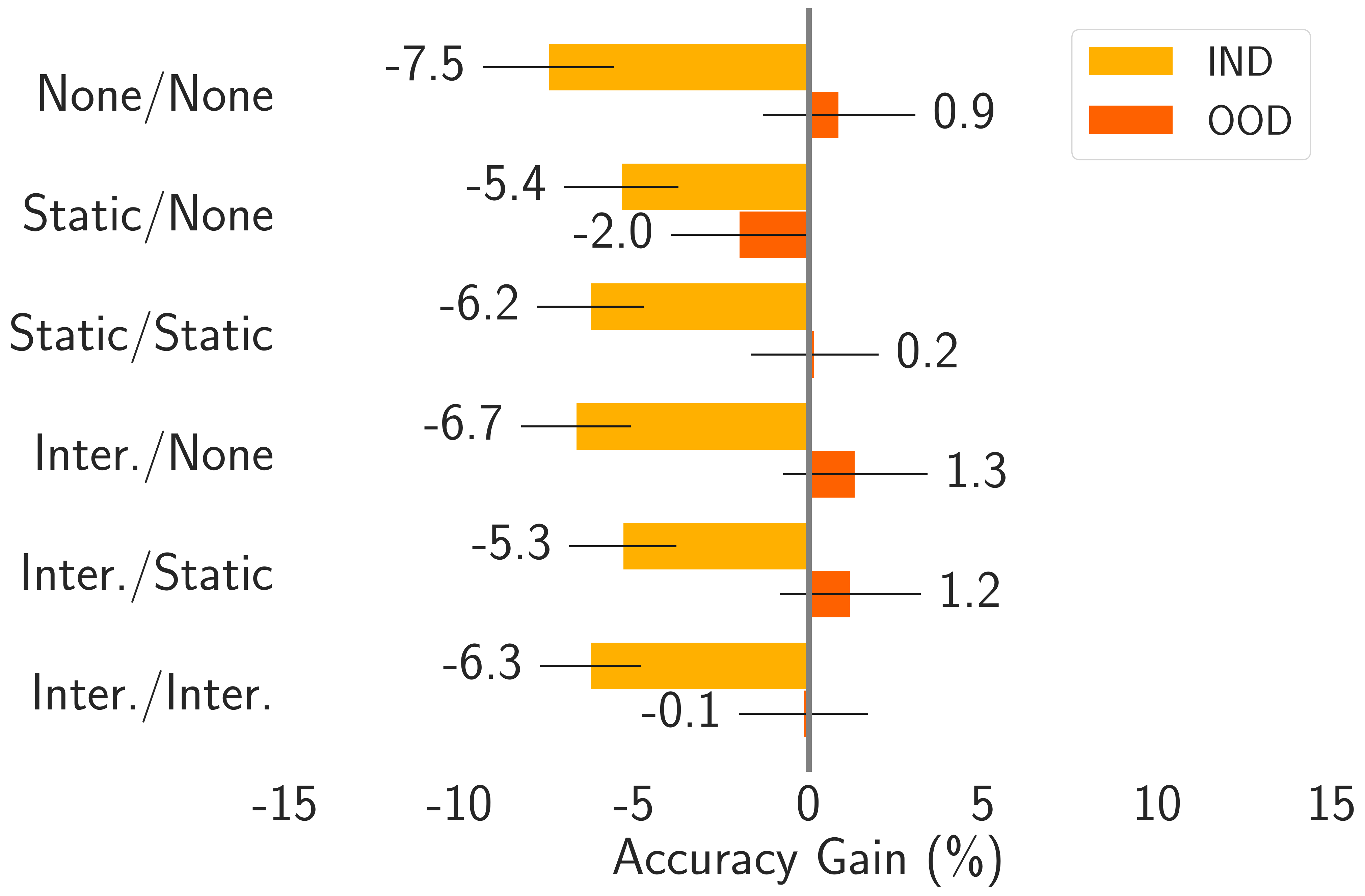}
        \caption{Accuracy gain in \icpsrtask. }
        \label{fig:icpsr_perf}
    \end{subfigure}
    \hfill
    \begin{subfigure}{0.48\textwidth}
		\includegraphics[width=0.9\textwidth]{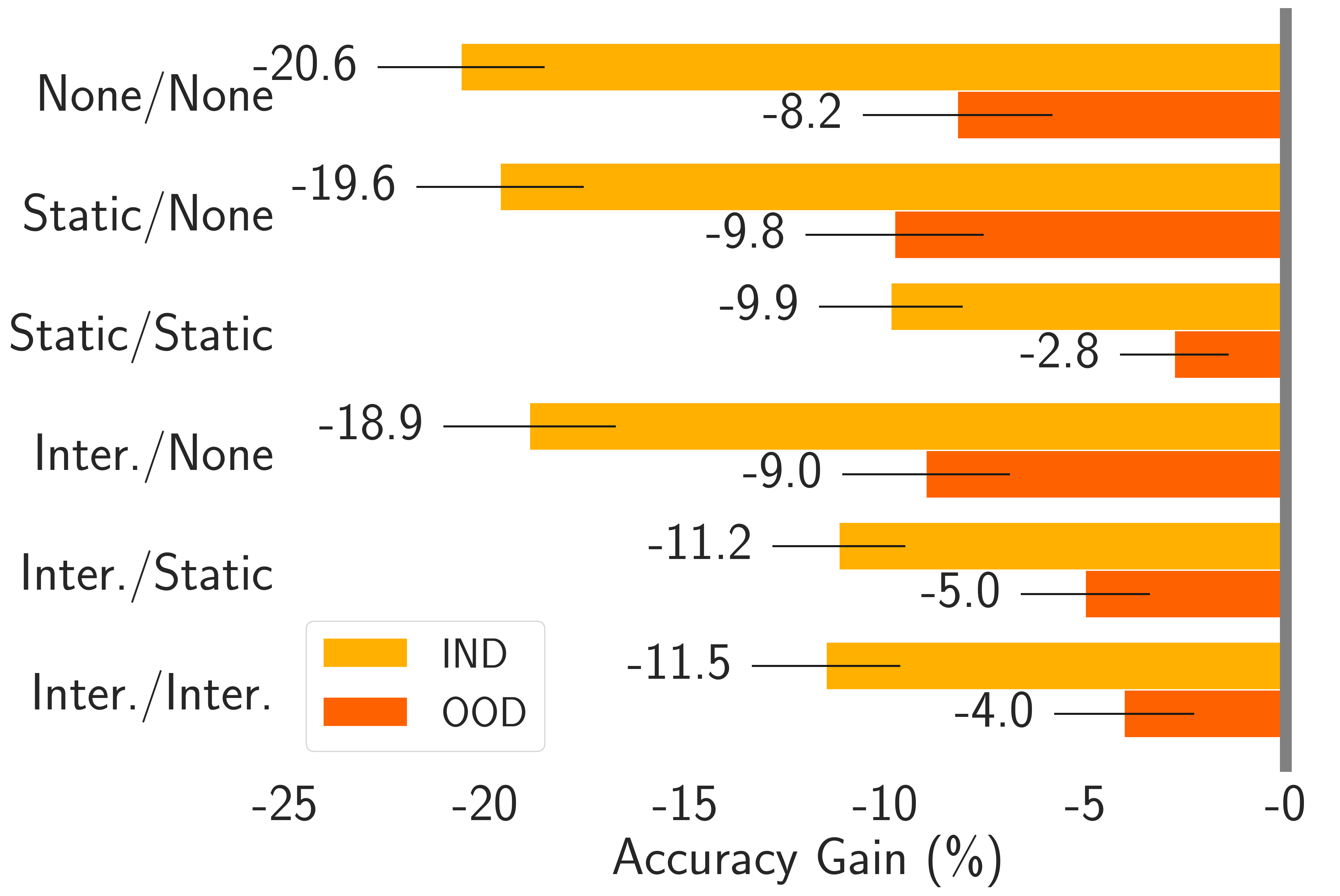}
        \caption{Accuracy gain in \biostask.}
        \label{fig:bios_perf}
    \end{subfigure}
    \caption{
      Accuracy gain in \icpsrtask and \biostask.
      \Domain types are indicated by the color of the bar and error bars represent 95\% confidence intervals.
      All accuracy gains are statistically significantly negative \ind, indicating that Human-AI teams underperform AI based on typical random split of training/test sets.
      However, results are mixed for \ood examples.
      While accuracy gain in \biostask is always negative, accuracy gain in \icpsrtask is sometimes positive (although not statistically significant).
      The performance gap
      between \humanaiteams and AI is generally smaller \ood than \ind, suggesting that humans may have more complementary insights to offer \ood. 
      Results in \compastask are similar to \icpsrtask and can be found in the supplementary materials.
    }
    \label{fig:performance}
    \Description{}
\end{figure}

\para{Human-AI teams underperform AI in \ind examples (see \figref{fig:performance}).}
We use $t$-tests with Bonferroni correction to determine whether the accuracy gain for \ind examples is statistically significant.
Consistent with Proposition~\ref{proposition},
our results show that accuracy gain is negative across all explanation types ($p < 0.001$).
In other words, the performance of \humanaiteams is lower than AI performance for \ind examples.
This observation also holds across all tasks, which means that AI may have an advantage in both challenging (\icpsrtask and \compastask) and relatively simple tasks (\biostask) for humans if the test set follows a similar distribution as the training set (\ind).

\para{Human-AI teams do not outperform AI in \ood examples, although the accuracy gain \ood is sometimes positive (see \figref{fig:performance}).}
Similarly,  we use $t$-tests with Bonferroni correction to determine whether the accuracy gain for \ood examples is statistically significant.
The results are different than what we expected: humans seldom outperform AI in \ood examples.
Interestingly, we observe quite different results across different tasks.
In \biostask, accuracy gain is significantly below 0 across all explanation types ($p < 0.001$).
In \icpsrtask and \compastask, accuracy gain is occasionally positive, including None/None, Static/Static,
Interactive/None, Interactive/Static in \icpsrtask, and Interactive/None in \compastask, although none of them is statistically significant.
The negative accuracy gain (Static/None) in \icpsrtask is not significant either.
These results suggest that 
although AI performs worse \ood than \ind, it remains challenging for \humanaiteams to outperform AI alone \ood.
The performance of \humanaiteams, however, becomes comparable to AI performance in challenging tasks such as recidivism prediction, partly because the performance of AI alone is more comparable to human alone \ood (e.g., 0.9\% in \icpsrtask vs. -8.2\% in \biostask in None/None (human alone) in \figref{fig:performance}).

Interestingly, Interactive/None leads to the highest accuracy gain in \icpsrtask, while Interactive/Interactive leads to a tiny negative gain, suggesting interactive explanations as real-time assistance might hurt human performance in \icpsrtask. 
We will elaborate on this observation in \secref{sec:interactive}.

\para{The performance gap between human-AI teams and AI is smaller in \ood examples than in \ind examples (see \figref{fig:performance}).}
We finally examine 
the difference between \ind and \ood examples.
We use two approaches to determine whether there exists a significant difference.
First, for each explanation type in each task, we test whether the accuracy gain in \ood examples is significantly different from that in \ind examples with $t$-tests after Bonferroni correction.
In both \biostask and \compastask, accuracy gain is significantly greater in \ood examples than in \ind examples across all explanation types ($p < 0.001$).
In \icpsrtask, accuracy gain is significantly greater in \ood examples than in \ind examples in all explanation types ($p < 0.001$) except Static/None.
Second, we conduct two-way ANOVA based on \domain types and explanation types.
We focus on the effect of \domain types here and discuss the effect of explanation types in \secref{sec:interactive}.
We observe a strong effect of \domain type across all tasks ($p<0.001$),
suggesting a clear difference between \ind and \ood.
Note that this reduced performance gap does not necessarily suggest that humans behave differently \ood from \ind, as it is possible that human performance stays the same and the reduced performance gap is simply due to a drop in AI performance. We further examine human agreement with AI predictions to shed light on the reasons behind this reduced performance gap.

In short, our results suggest a significant difference between \ind and \ood,
and \humanaiteams are more likely to perform well in comparison with AI \ood.
These results are robust across different explanation types.
In general, the accuracy gain is greater in recidivism prediction than in \biostask.
After all, the \ind AI performance in \biostask is much stronger than humans without any assistance.
This observation resonates with the hypothesis in
\citet{bansal2021does} 
that comparable performance between humans and AI is related to complementary performance. However,  we do not observe complementary performance in our experiments, which suggests that comparable performance between humans and AI alone is insufficient for complementary performance.

\section{RQ2: Agreement/Trust of Humans with AI}
\label{sec:agreement}

\begin{table}[t]
    \centering
    \begin{tabular}{p{0.4\linewidth}|c@{\hspace{4pt}}c@{\hspace{4pt}}c|c@{\hspace{4pt}}c@{\hspace{4pt}}c}
        \toprule
        \multirow{2}{*}{} &
            \multicolumn{3}{c|}{IND (typical setup)} &
            \multicolumn{3}{c}{OOD (proposed setup)} \\
            & \icpsrtask & \compastask & \biostask & \icpsrtask & \compastask & \biostask \\ \midrule

         Agreement is higher \ind than \ood. & \multicolumn{3}{c|}{\textcolor{gray}{see the \shortood columns}} & \cmark & \hcmark & \hreversedmark \\ \midrule
         Agreement is higher when AI predictions are correct (appropriate agreement) than when AI predictions are wrong (overtrust). & \hreversedmark & \xmark & \cmark & \hcmark & \hcmark & \cmark \\ \midrule
         When AI predictions are correct, agreement (appropriate agreement) is higher \ind than \ood. & \multicolumn{3}{c|}{\textcolor{gray}{see the \shortood columns}} & \hcmark & \xmark & \reversedmark \\ \midrule
         When AI predictions are wrong, agreement (overtrust) is higher \ind than \ood. & \multicolumn{3}{c|}{\textcolor{gray}{see the \shortood columns}} & \cmark & \hcmark & \hxmark \\ \midrule
        \multicolumn{1}{l}{\cmark: ~holds} & \multicolumn{6}{l}{\hcmark: holds in at least half of the explanation types} \\
        \multicolumn{1}{l}{\xmark: ~rejected} &
      \multicolumn{6}{l}{\hxmark: rejected in all except one explanation type}\\
        \multicolumn{1}{p{0.4\linewidth}}{\reversedmark: ~mostly supported in the reverse direction except one explanation type} & \multicolumn{6}{l}{\hreversedmark: reversed only in one explanation type} \\ 
        \bottomrule
    \end{tabular}
    \medskip
    \caption{Summary of results on agreement with AI. Recall that appropriate agreement refers to humans agreeing with correct AI predictions, and overtrust refers to humans agreeing with incorrect AI predictions.
    }
    \label{tb:preview-agreement}
\end{table}

Our second research question examines
how
well human predictions agree with AI predictions
depending on the \domain type.
Agreement is defined as the percentage of examples where the human gives the same prediction as AI.
Humans have access to AI predictions in Static/Static, Interactive/Static, Interactive/Interactive, so agreement in these explanation types may be interpreted as how much {\em trust} humans place in AI predictions (we use {\em overtrust} to refer to agreement with incorrect predictions in all explanation types).
Since both \icpsrtask and \compastask yield similar results, we show \icpsrtask results in the main paper and \compastask in the supplementary materials (see \figref{fig:compas_perf}-\figref{fig:compas_top_features}).

\para{Preview of results.}
Different from results in performance, we observe intriguing differences across tasks.
Our results show that humans tend to show higher agreement with AI predictions in \ind examples than \ood examples in \icpsrtask and \compastask, but not in \biostask.
When it comes to appropriate agreement vs. overtrust, 
the results depend on \domain types.
We first compare the extent of  appropriate agreement and overtrust in the same \domain type.
In \ood examples, human agreement with AI predictions is higher when AI predictions are correct than when AI predictions are wrong (appropriate agreement exceeds overtrust).
But for \ind examples, this is only true for \biostask, but false in \icpsrtask and \compastask.
To further understand these results, we compare appropriate agreement and overtrust \ind to \ood.
We find that both appropriate agreement and overtrust are stronger \ind than \ood in \icpsrtask, but in \biostask, the main statistical significant results are that appropriate agreement is stronger \ood than \ind.
See Table~\ref{tb:preview-agreement} for a summary.

\begin{figure}[t]
    \centering
    \begin{subfigure}{0.48\textwidth}
        \includegraphics[width=0.9\textwidth]{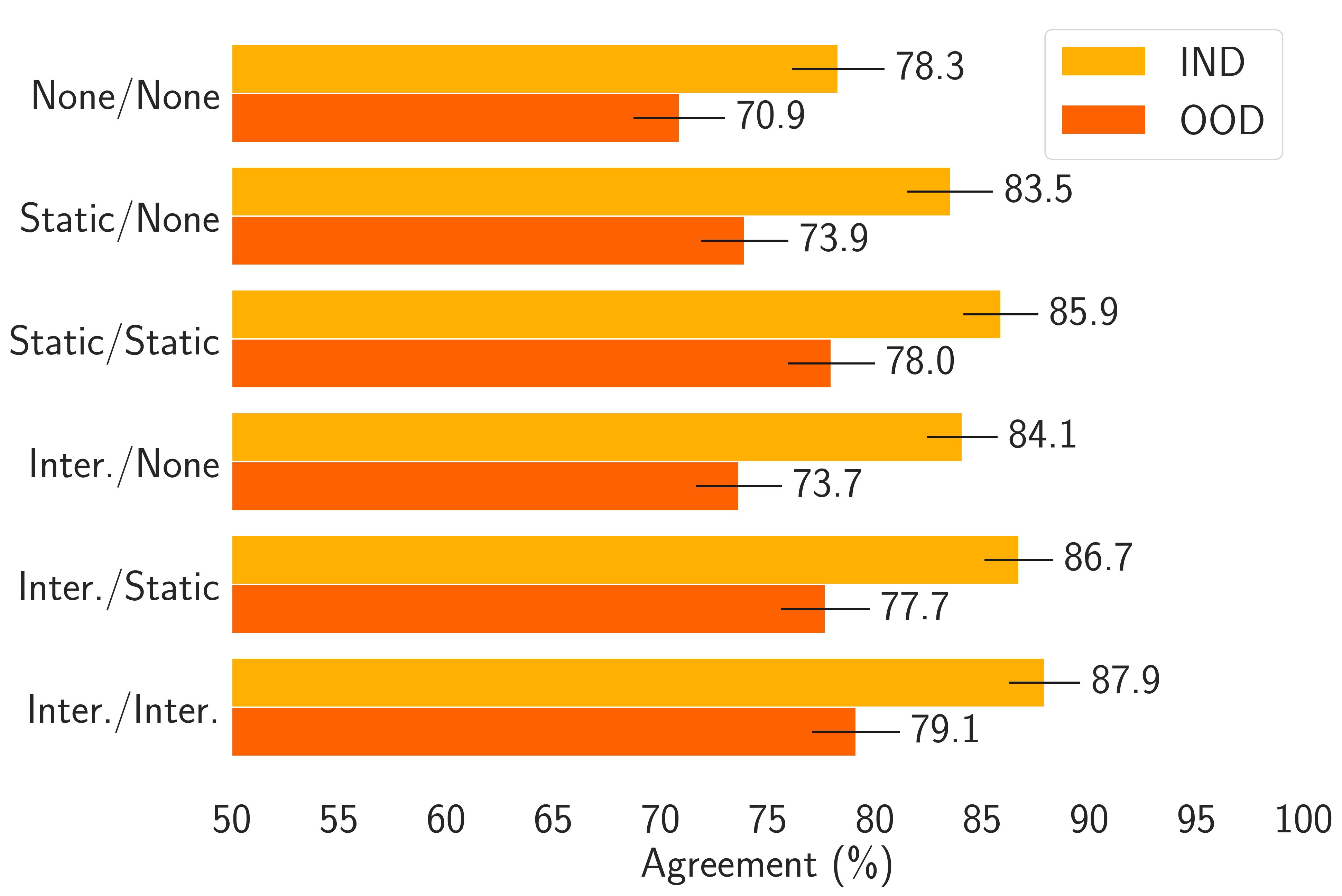}
        \caption{Agreement with AI predictions in \icpsrtask.}
        \label{fig:icpsr_agree}
    \end{subfigure}
    \hfill
    \begin{subfigure}{0.48\textwidth}
		\includegraphics[width=0.9\textwidth]{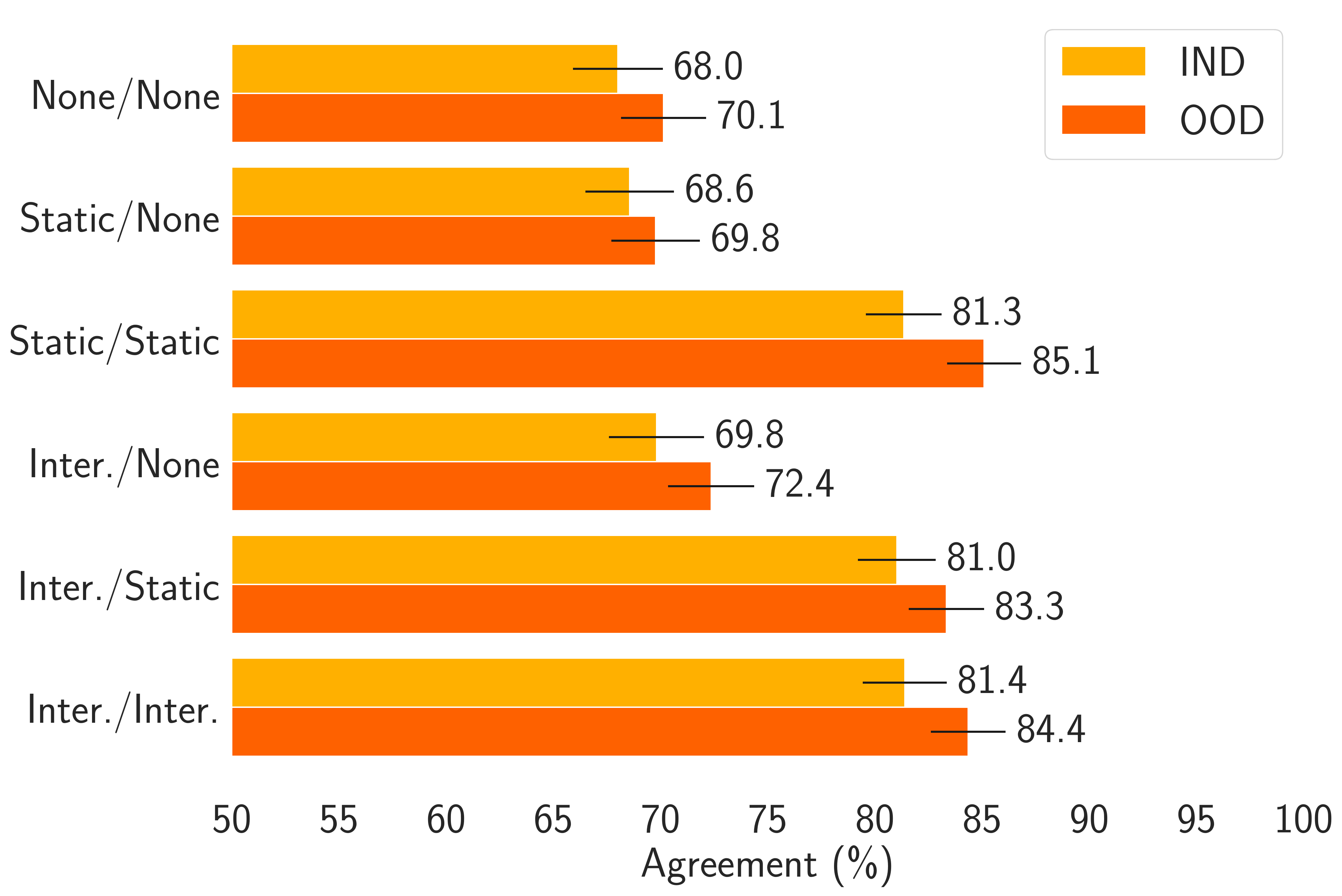}
        \caption{Agreement with AI predictions in \biostask.}
        \label{fig:bios_agree}
    \end{subfigure}
    \Description{}
    \caption{Agreement with AI predictions in \icpsrtask and \biostask. \Domain types are indicated by the color of the bar and error bars represent 95\% confidence intervals.
    In \icpsrtask and \compastask, agreement with AI predictions is much higher \ind than \ood. However, this trend is reversed in \biostask.
    In \biostask, agreement is generally higher in Static/Static, Interactive/Static, and Interactive/Interactive, where AI predictions and explanations are shown.
    We will discuss the effect of explanation type in \secref{sec:interactive}.
    }
    \label{fig:agreement}
\end{figure}

\para{Humans are more likely to agree with AI on \ind examples than \ood examples in \icpsrtask and \compastask, but not in \biostask(see \figref{fig:agreement}).}
As AI performance is typically
better \ind than \ood, we expect humans to agree with AI predictions more often \ind than \ood.
To determine whether the difference is significant, we use $t$-test with Bonferroni correction for each explanation type in \figref{fig:agreement}.
In \icpsrtask, agreement is indeed significantly greater \ind than \ood in all explanation types ($p<0.001$).
In \compastask, \ind agreement is significantly higher in all explanation types ($p<0.05$ in None/None, $p<0.01$ in Static/None and Interactive/Interactive, $p<0.001$ in Interactive/None) except Static/Static and Interactive/Static (see \figref{fig:compas_agree}).
These results suggest that in \icpsrtask and \compastask, humans
indeed behave more differently from AI \ood.
However,
in \biostask, we find the agreement is generally higher for \ood examples than for \ind examples, and the difference is statistically significant
in Static/Static
($p<0.05$).
Note that the agreement difference between \ind and \ood is much smaller in \biostask ($<$4\%, usually within 2\%) than in \icpsrtask and \compastask ($\sim$10\%).

These results echo observations in our virtual pilot studies
that humans are more confident in themselves when detecting professions and are
less affected by \ind vs. \ood differences, and may turn to AI predictions \ood because the text is too short for them to determine the label confidently.
In comparison, the fact that humans agree with AI predictions less \ood than \ind in recidivism prediction suggests that humans seem to recognize that AI predictions are more likely to be wrong \ood than \ind in \icpsrtask and \compastask. To further unpack this observation, we analyze human agreement with correct AI predictions vs. incorrect AI predictions.

\begin{figure}[t]
    \centering
    \begin{subfigure}{0.44\textwidth}
        \includegraphics[trim=0 0 390 0, clip, width=\textwidth]{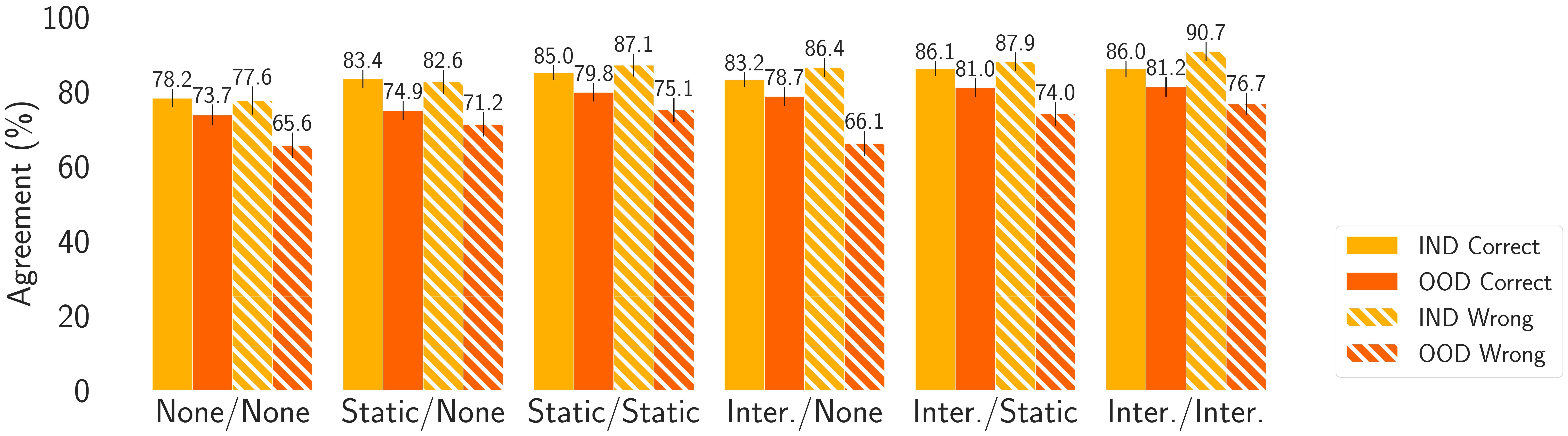}
        \caption{\icpsrtask agreement by correctness.}
        \label{fig:icpsr_correct}
    \end{subfigure}
    \hfill
    \begin{subfigure}{0.54\textwidth}
        \includegraphics[trim=0 0 0 0, clip, width=\textwidth]{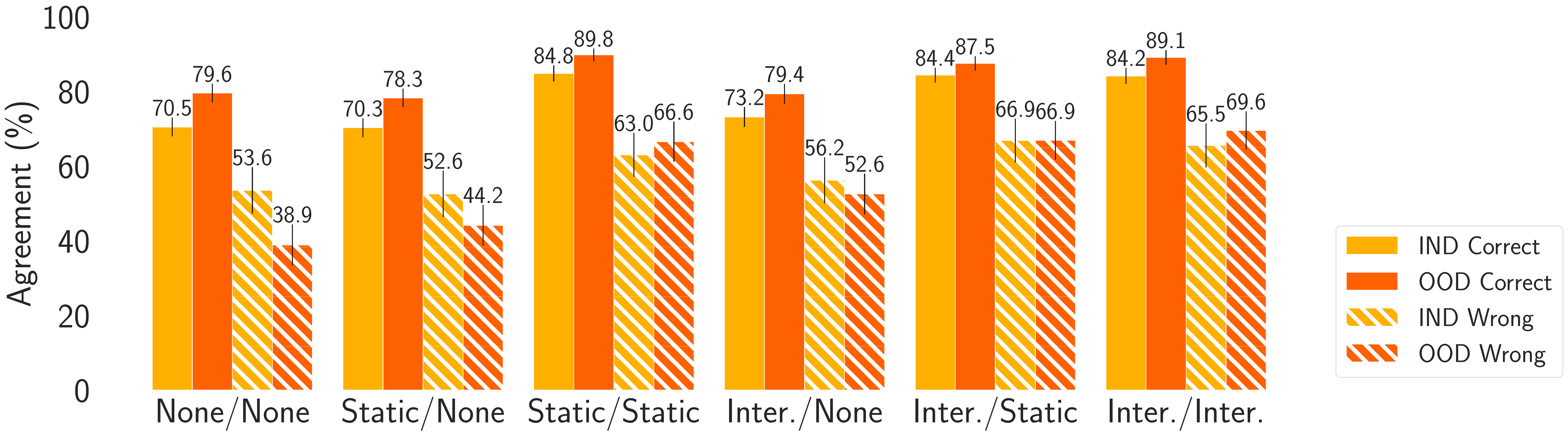}
        \caption{\biostask agreement by correctness.}
        \label{fig:bios_correct}
    \end{subfigure}
    \caption{Agreement with AI grouped by \domain type and whether AI predictions are correct.
    \Domain types are indicated by the color of the bar, bars with stripes represent wrong AI predictions, and error bars represent 95\% confidence intervals.
    A notable observation is that when AI is wrong, humans are significantly less likely to agree with AI predictions \ood than \ind in \icpsrtask and \compastask, but it is not the case in \biostask.
    }
    \Description{}
    \label{fig:agreement_correct}
\end{figure}

\para{%
\Ood appropriate agreement {\em mostly exceeds} \ood overtrust in all of the three tasks; \ind appropriate agreement exceeds \ind overtrust {\em only} in \biostask%
(%
    see \figref{fig:agreement_correct}).}
We next examine the role of \domain type in whether humans can somehow distinguish when AI is correct from when AI is wrong.
First, for each \domain type,
we use $t$-test with Bonferroni correction
to determine if humans agree with AI more when AI predictions are correct.
Consistent with prior work \citep{lai+tan:19,bansal2021does},
we find that \humanaiteams are more likely to agree with AI when AI predictions are correct than when AI predictions are wrong in most explanation types.
This is 
true both \ind and \ood in \biostask ($p<0.001$):
the agreement gap between correct and incorrect AI predictions is close to 20\%, and even reaches 30\%-40\% \ood with some explanation types (\figref{fig:bios_correct}).
In \icpsrtask and \compastask,
we mostly find significantly greater appropriate agreement than overtrust \ood.
In fact, \shortind appropriate agreement tends to be lower than \shortind overtrust, though only significantly in Interactive/Interactive ($p<0.05$) in \icpsrtask.
In comparison, for \ood examples,
appropriate agreement is significantly higher than overtrust in three explanation types in \icpsrtask($p<0.01$ in None/None, Interactive/None, and Interactive/Static).
In \compastask, appropriate agreement is also significantly higher than overtrust in \ood examples ($p<0.05$ in None/None and Interactive/Static, $p<0.01$ in Static/None and Interactive/None) except Static/Static and Interactive/Interactive (see \figref{fig:compas_correct}).
These results are especially intriguing as they suggest that although the performance of human alone and AI alone is worse \ood than \ind in recidivism prediction, humans can more accurately detect AI mistakes, which explains the small positive accuracy gain in \figref{fig:performance}.
\para{\Ind and \ood appropriate agreement comparison shows different results in each of the three tasks
     (%
see \figref{fig:agreement_correct}).}
We further compare human agreement between \ind and \ood when AI is correct. Similarly, we use $t$-tests with Bonferroni corrections for each explanation type.
Different from our expectation, appropriate agreement ~is significantly higher \ood than \ind in all explanation types in \biostask except Interactive/Static ($p<0.001$ in None/None and Static/None; $p<0.01$ in Static/Static, Interactive/None, and Interactive/Interactive).
This is consistent with the observation of higher overall agreement \ood than \ind in \biostask 
in \figref{fig:agreement}.
In \icpsrtask, appropriate agreement for \ind examples is significantly higher than for \ood examples in %
all explanation types except None/None ($p<0.01$ in Interactive/None, Interactive/Static, and Interactive/Interactive, $p<0.05$ in Static/None and Static/Static).
In \compastask, no significant difference is found between \ind and \ood.

These results suggest that appropriate agreement is stronger \ood than \ind in \biostask. In other words, humans can recognize correct AI predictions better \ood than \ind.
This could relate to 
that humans have higher confidence 
in their own predictions when the text is longer. As a result, they are more likely to overrule correct AI predictions.
However, appropriate agreement is stronger \ind than \ood in \icpsrtask, which relatively weakens the performance of \humanaiteams compared to AI alone \ood, and suggests that a reduced overtrust is the main contributor to the aforementioned reduced performance gap.
In comparison, it seems that in \compastask, humans simply tend to agree with AI predictions more \ind than \ood, without the ability to recognize when AI predictions are correct.

\para{Overtrust is lower \ood than \ind in \icpsrtask and \compastask, but not in \biostask (see Fig.~\ref{fig:agreement_correct}).}
In comparison, when AI predictions are wrong,
human agreement is significantly lower for \ood examples than \ind examples in all explanation types ($p<0.001$) in \icpsrtask.
This also holds for
some explanation types ($p<0.01$ in Static/None, Interactive/None, and Interactive/Static) in \compastask.
However, overtrust in \ind examples has no significant difference from \ood examples in \biostask
except for None/None ($p<0.01$).
These results suggest that in recidivism prediction, human decisions contradict wrong AI predictions \ood more accurately than \ind, but it is not the case in \biostask.
In summary, 
 the contrast between appropriate agreement and overtrust is interesting as it explains the different stories behind the reduced performance gap \ood compared to \ind in \icpsrtask and in \biostask:
the reduced performance gap in \biostask is mainly attributed to the higher appropriate agreement \ood,
while the reduced performance gap in \icpsrtask is driven by the lower overtrust \ood. 
These results may relate to the task difficulty for humans.
Recidivism prediction is more challenging for humans and the advantage of humans may lie in the ability to recognize obvious AI mistakes.
In constrast, as humans are more confident in their predictions in \biostask, it is useful that they avoid overruling correct AI predictions.
Such asymmetric shifts in agreement rates highlight the complementary insights that humans can offer when working with AI assistance and suggest interesting design opportunities to leverage human expertise in detecting AI mistakes.
\section{RQ3: The Effect of Interactive Explanations}
\label{sec:interactive}

In this section, we focus on the effect of interactive explanations in human decision making.
We revisit human performance and human agreement and then examine human perception of AI assistance's usefulness collected in our exit survey.
Finally, for \icpsrtask and \compastask, we
take a deep look at the most important features
reported by humans in the exit survey to
understand the limited improvement in the performance of \humanaiteams.

\para{Preview of results.}
In general, we do not find significant impact from interactive explanations with respect to the performance of \humanaiteam or human agreement with wrong AI predictions, compared to static explanations.
However, humans are more likely to find AI assistance useful with interactive explanations than static explanations in \icpsrtask and \compastask, but not in \biostask.
Table~\ref{tb:preview-interaction} summarizes the results.

\begin{table}[t]
  \centering
  \begin{tabular}{p{0.4\linewidth}|c@{\hspace{4pt}}c@{\hspace{4pt}}c|c@{\hspace{4pt}}c@{\hspace{4pt}}c}
      \toprule
      \multirow{2}{*}{} &
          \multicolumn{3}{c|}{IND (typical setup)} &
          \multicolumn{3}{c}{OOD (proposed setup)} \\
          & \icpsrtask & \compastask & \biostask & \icpsrtask & \compastask & \biostask \\ \midrule

       Interactive explanations lead to better \humanaiteam performance. & \xmark & \xmark & \xmark & \xmark & \xmark & \xmark \\ \midrule
       Interactive explanations lead to lower human agreement with wrong
      AI predictions (overtrust). & \xmark & \xmark & \xmark & \xmark & \xmark & \xmark \\ \midrule
       \Humanaiteams are more likely to find AI assistance useful with interactive explanations. & \multicolumn{3}{c|}{\textcolor{gray}{see the \shortood columns}} & \hcmark & \hcmark & \xmark \\ \midrule

      \multicolumn{1}{l}{\cmark: ~holds} & \multicolumn{6}{l}{\hcmark: holds in at least half of the explanation types} \\
      \multicolumn{1}{l}{\xmark: ~rejected} &
      \multicolumn{6}{l}{\hxmark: rejected in all except one explanation type}\\
      \bottomrule
  \end{tabular}
  \medskip
  \caption{Summary of results on the effect of interactive explanations.
  }
  \label{tb:preview-interaction}
\end{table}

\para{Real-time assistance leads to better performance than no assistance in \biostask, but interactive explanations do not lead to better human-AI performance than AI alone (see \figref{fig:performance}).}
We conduct one-way ANOVA on explanation type for \ind and \ood separately on human performance due to the clear difference between \ind and \ood.
We find that explanation type affects human performance in both \domain types significantly in \biostask ($p<0.001$),
but not in \icpsrtask ($p=0.432$ \shortind, $p=0.184$ \shortood) nor in \compastask ($p=0.274$ \shortind, $p=0.430$ \shortood).
We further use Tukey's HSD test to see if differences between explanation types are significant.
In \biostask, we find Static/Static, Interactive/Static, and Interactive/Interactive have significantly better performance than None/None, Static/None, and Interactive/None for \ind examples ($p<0.001$). For \ood examples, we have almost the same observation ($p<0.05$) except that the difference between Interactive/Static and None/None is no longer significant.
These results suggest that
real-time assistance in the prediction phase improves human performance in \biostask, consistent with \citep{lai+tan:19,bansal2021does},
although there is no significant difference between
static and interactive explanations.
In \icpsrtask and \compastask, no significant difference exists
between any pair of explanation types.
In other words, no explanation type leads to better nor worse \humanaiteam performance in recidivism prediction.

\begin{figure}[t]
    \centering
    \begin{subfigure}{0.4185\textwidth} %
      \includegraphics[width=\textwidth]{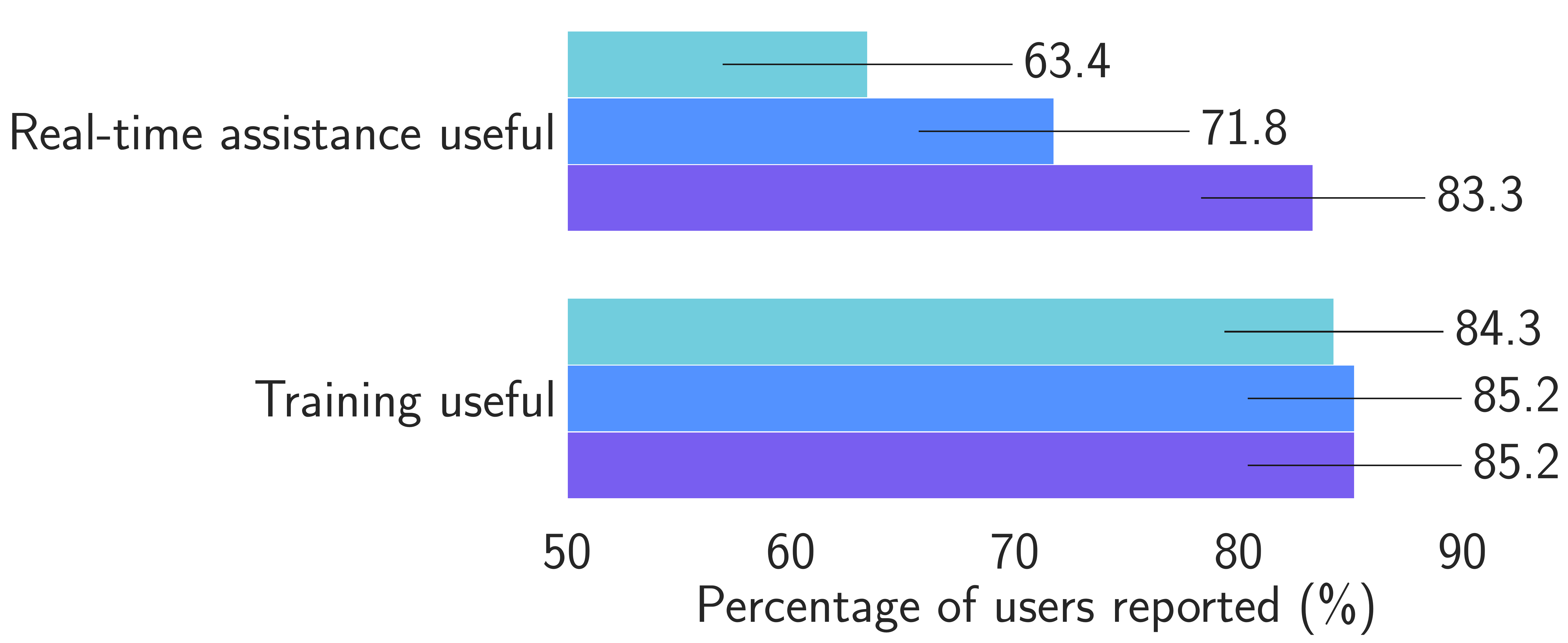}
      \caption{\icpsrtask. }
      \label{fig:icpsr_subjective}
    \end{subfigure}
    \begin{subfigure}{0.405\textwidth} %
        \includegraphics[width=\textwidth]{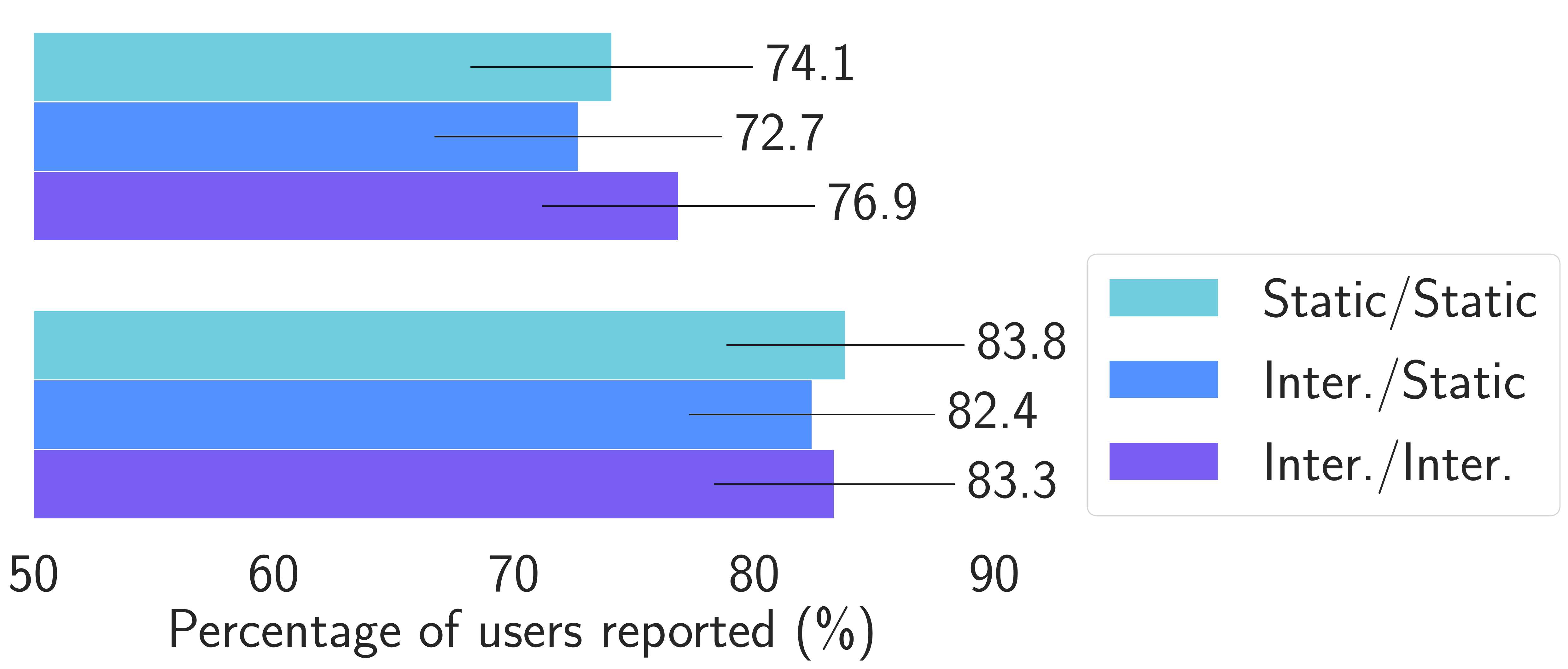}
        \caption{\biostask.}
        \label{fig:bios_subjective}
    \end{subfigure}
    \caption{Human perception on whether real-time assistance is useful and whether training is useful.
    $x$-axis shows the percentage of users that answered affirmatively.
    Error bars represent 95\% confidence interval.
    }
    \label{fig:subjective}
    \Description{}
\end{figure}

\para{Interactive explanations do not lead to significantly lower overtrust (see \figref{fig:agreement_correct}).}
We use one-way ANOVA to determine whether significant differences in overtrust exist between different explanation types. We also do this separately for \ind and \ood examples.
We observe a strong effect in all tasks in both \domains ($p<0.001$).
However, Tukey's HSD test shows overtrust in Interactive/Interactive is not statistically different from Static/Static; similarly, Interactive/None is not statistically different from Static/None either.
The strong effect comes from the significant differences between explanation types with real-time assistance and those without, likely because predicted labels are shown in real-time assistance.
For example, in \ood examples in \biostask, three explanation types without real-time assistance (None/None, Static/None, Interactive/None) have significantly lower overtrust than the three with real-time assistance (Static/Static, Interactive/Static, Interactive/Interactive) ($p < 0.001$ for most pairs; $p < 0.01$ for Interactive/None vs. Static/Static and Interactive/None vs. Interactive/Static). %
Similarly, in \ood examples in \icpsrtask, None/None and Interactive/None has significantly lower overtrust than Static/Static, Interactive/Static, and Interactive/Interactive 
($p < 0.001$ for most pairs; $p < 0.01$ for None/None vs. Interstatic/Static, Interactive/None vs. Static/Static, and Interactive/None vs. Interactive/Static).
In fact, Interactive/Interactive has the highest overtrust in both \ind and \ood examples in \icpsrtask.
Results in \compastask are qualitatively similar (see \figref{fig:compas_agree}).\footnote{For \ind overtrust, None/None is significantly lower than explanation types with real-time assistance ($p<0.05$ in Static/Static; $p<0.001$ in Interactive/Static and Interactive/Interactive). For \ood overtrust, all explanation types without real-time assistance are significantly lower than Static/Static ($p<0.05$) and Interactive/Interactive ($p<0.01$).
  However, similarly to \icpsrtask, we do not see significantly lower overtrust in interactive explanations than in static explanations either \ind or \ood.}

These results are contrary to our expectation: interactive explanations do not lead to lower overtrust.
In fact, they lead to the highest overtrust in \icpsrtask, so they may not encourage users to critique incorrect AI predictions.
Our observations also resonate with prior work that shows higher overall agreement with AI predictions when predicted labels are shown \citep{lai+liu+tan:20,lai+tan:19}.

\para{Human-AI teams are more likely to find AI assistance useful with interactive explanations in \icpsrtask and \compastask, but not in \biostask(see \figref{fig:subjective}).}
We ask participants whether they find training and real-time assistance useful when applicable.
Since only Static/Static, Interactive/Static, and Interactive/Interactive have real-time assistance, we focus our analysis here on these three explanation types.
We use one-way ANOVA to test the effect of explanation type for the usefulness of training and real-time AI assistance separately.
For training, the effect of explanation type is significant only in \compastask ($p<0.05$). 
With Tukey's HSD test, we find the perception of training usefulness is significantly higher in Interactive/Interactive than in Static/Static ($p<0.05$).
These results show that \humanaiteam with interactive explanations are more likely to find training useful in \compastask.

\begin{figure}[t]
    \centering
    \begin{subfigure}{0.52\textwidth}
      \includegraphics[width=0.85\textwidth]{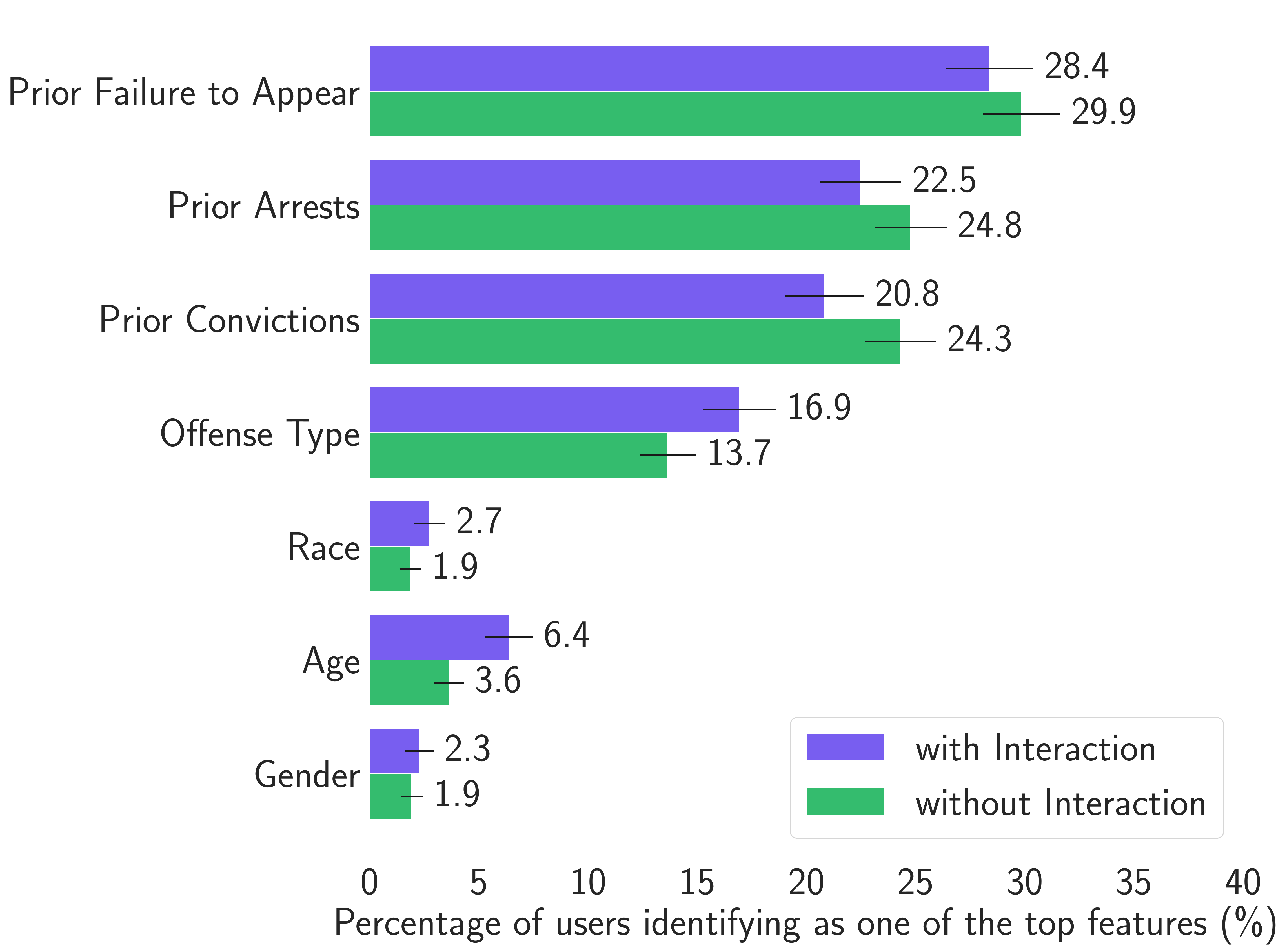}
      \caption{Percentage of participants finding a feature important in \icpsrtask. }
      \label{fig:icpsr_top_features}
    \end{subfigure}
        \hfill
        \begin{subfigure}{0.45\textwidth}
                \includegraphics[width=0.95\textwidth]{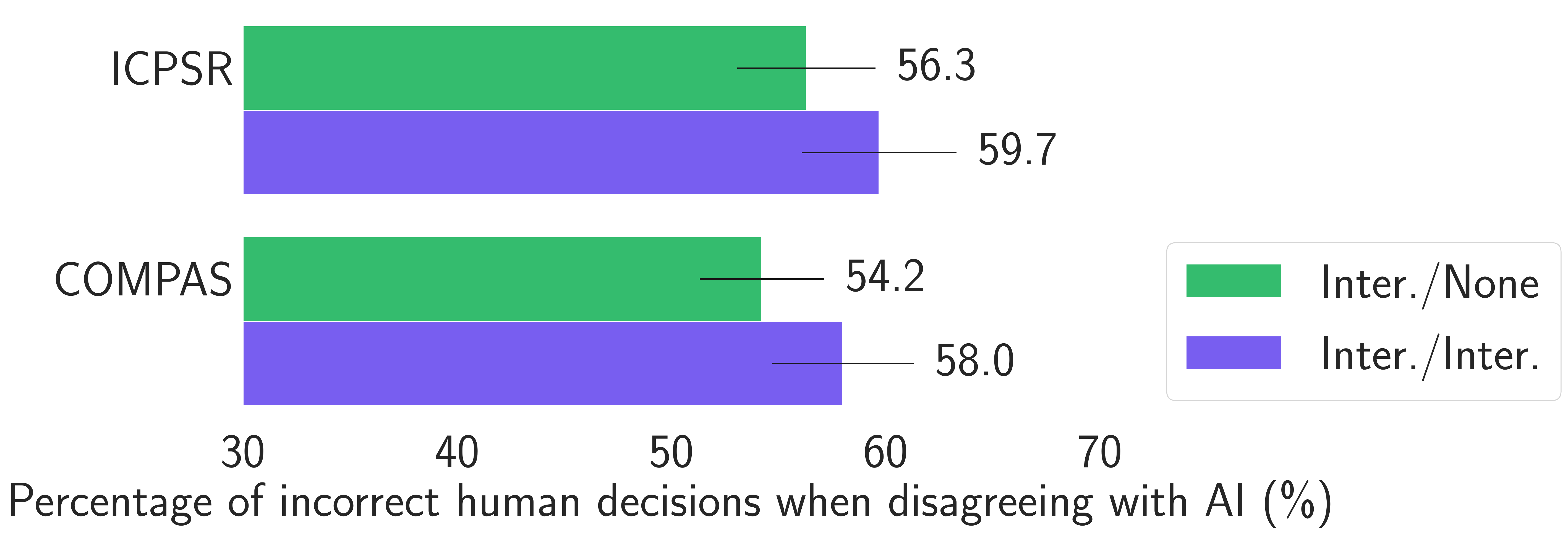}
                \caption{The percentage of examples answered wrongly by participants when they disagree with AI predictions. }
                \label{fig:conf_mat}
        \end{subfigure}
    \Description{}
    \caption{
    The features in \ref{fig:icpsr_top_features} are sorted in descending order from top to bottom by their Spearman correlation with groundtruth labels.
    }
    \label{fig:top_features}
\end{figure}

For perception of real-time assistance, explanation type has a significant effect
in \compastask ($p<0.001$) and \icpsrtask ($p<0.001$), 
but not in \biostask ($p=0.6$).
We also use Tukey's HSD test to determine whether there is a pairwise difference among explanation types.
In \compastask, Interactive/Interactive achieves a significantly higher human perception of real-time assistance usefulness than both Static/Static ($p<0.001$) and Interactive/Static ($p<0.05$) (see \figref{fig:compas_subjective}).
Perception of Interactive/Static is also significantly higher than that of Static/Static ($p<0.001$).
We find similar results in \icpsrtask except that the difference between Static/Static and Interactive/Static is not significant.
In \biostask, Interactive/Interactive
has the highest human perception of AI assistance usefulness, but no significant difference is found.
These results suggest that with interactive explanations, \humanaiteams perceive real-time assistance as more useful, especially in recidivism prediction.
A possible reason is that human perception of usefulness 
depends on the difficulty of tasks.
\compastask is more challenging than \biostask to humans as recidivism
prediction is not an average person's experience, thus interactive
explanations may have decreased the difficulty of the task in perception.
\para{Exploratory study on important features.}
Finally, since there are only seven features in \icpsrtask and \compastask, we asked participants to identify the top three most important features that made the biggest influence on their own predictions in the exit survey (see \figref{fig:survey} for the wording of all survey questions).
We also identify important features based on Spearman correlation
as a comparison point.
The top three are (``Prior Failure to Appear'', ``Prior Arrests'', ``Prior Convictions'') in \icpsrtask, and
(``Prior Crimes'', ``Age'', and ``Race'') in \compastask.
By comparing these computationally important features with human-perceived important features, we can identify potential biases in human perception to better understand the limited performance improvement.

\figref{fig:icpsr_top_features} shows the percentage of participants that choose
each feature as an important feature for their decisions in \icpsrtask.
We group participants based on explanation types: 1) without interactions (Static/None and Static/Static) and 2) with interactions (Interactive/None, Interactive/Static, and Interactive/Interactive).
Humans largely choose the top computationally important features in both groups in \icpsrtask.
We use $t$-test with Bonferroni correction to test whether there is a difference between the two groups.
In \icpsrtask, we find participants with interaction choose significantly more ``Age'' and ``Offense Type'', but less ``Prior Convictions'' (all $p<0.01$).
In fact, participants with interaction are less likely to choose all of the top three features than those without.
In \compastask (see \figref{fig:compas_top_features}), we find participants with interaction choose significantly more ``Race'' and ``Sex'', but less ``Charge Degree'' ($p<0.001$ in ``Race'', $p<0.05$ in ``Sex'' and ``Charge Degree'').
These results suggest that participants with interaction are
more likely to fixate on demographic features and potentially reinforce human biases,\footnote{Race is indeed important in \compastask, so this might be justified to a certain extent.}
but are less likely to identify computationally important features in \icpsrtask and \compastask.

This observation may also relate to why interactive explanations do not lead to better performance of \humanaiteams.
We thus hypothesize that participants with interaction make more mistakes when they disagree with AI predictions, which can explain the performance difference between Interactive/None and Interactive/Interactive in \figref{fig:performance}.
\figref{fig:agreement} shows that
users disagree with AI predictions less frequently in Interactive/Interactive than in Interactive/None, 
and \figref{fig:conf_mat} 
further shows that they are indeed more likely to be wrong
when they disagree (not statistically significant).

\section{Discussion}
\label{sec:discussion}

In this work, we investigate the effect of \ood examples and interactive explanations on human-AI decision making through both virtual pilot studies and large-scale, randomized human subject experiments.
Consistent with prior work, our results show that the performance of \humanaiteams is lower than AI alone \ind.
This performance gap becomes smaller \ood, suggesting a clear difference between \ind and \ood, although complementary performance is not yet achieved.
We also observe intriguing differences between tasks with respect to human agreement with AI predictions.
For instance, participants in \icpsrtask and \compastask agree with AI predictions more \ind than \ood, which is consistent with AI performance differences \ind and \ood, but it is not the case in \biostask.
As for the effect of interactive explanations, although they 
fail to improve the performance of \humanaiteams,
they tend to improve human perception of AI assistance's usefulness, 
with an important caveat of potentially reinforcing  human biases.
Our work highlights the promise and importance of exploring \ood examples.
The  performance gap between \humanaiteams and AI alone is smaller \ood than \ind both in recidivism prediction, where the task is challenging and humans show comparable performance with AI,
and in \biostask, where the task is easier for both humans and AI but AI demonstrates a bigger advantage than humans.
However, complementary performance is not achieved in our experiments, suggesting that \ood examples and interactive explanations (as we approach them) are not the only missing ingredients.
Similarly, comparable performance alone might not be a sufficient condition for complementary performance.
While results with respect to \humanaiteam performance and the effect of interactive explanations are relatively stable across tasks,
the intriguing differences in human agreement with AI predictions between tasks demonstrate the important role of tasks and the complexity of interpreting findings in this area.
We group our discussion of implications by \ood experiment design, interactive explanations, and choice of tasks, and then conclude with other limitations.
\para{\Ood experimental design.}
The clear differences between \ind and \ood suggest that \domain type should be an important factor when designing experimental studies on human-AI decision making.
Our results also indicate that it is promising to reduce the performance gap between \humanaiteams and AI for \ood examples, as AI is more likely to suffer from distribution shift.
\Ood examples, together with typical \ind examples, provide a more realistic examination of human-AI decision making and represent an important direction to examine how humans and AI complement each other.

However, it remains an open question of what the best practice is for evaluating the performance of \humanaiteams~\ood.%
\footnote{Concurrently with this work, \citet{chiang2021you} investigates human reliance on machine predictions when humans are aware of \domain shifts.}
To simulate \ood examples, we use separate bins based on an attribute (age for \icpsrtask and \compastask; length for \biostask).
Our setup is realistic in the sense that it is possible that age distribution in the training data differs from the testing data and leads to worse generalization performance in \ood examples in recidivism prediction.
Similarly, length is a sensible dimension for distributon mistach in text classification.
That said, our choice of separate bins leads to non-overlapping \ood and \ind examples. In practice, the
difference between \ood and \ind can be continuous and subtle to quantify~\cite{koh2021wilds}.
From an experimental point of view, it is challenging to investiage the effect of \ood examples on a continuous spectrum, and  \ood examples that are very close to \ind examples may not be interesting to study. As a result, it makes sense to zoom in on the challenging \ood examples and have a clear separation between \ind and \ood. We believe that our design represents a reasonable first attempt in understanding the effect of \ood examples and future work is required to address the spectrum of \ood.

Notably, a side effect of our split is that \ood examples are more difficult than \ind examples for humans in recidivism prediction (but not in \biostask; see \figref{fig:perf_norm}).
We encourage future work to examine to what extent this is true in practice and how this shift affects human decision making. 
Furthermore, \ood examples might benefit from new feature representations, which humans can extract, pointing to novel interaction with AI.
Overall, many research questions emerge in designing experiments and  interfaces to effectively integrate humans and AI under distribution shift.

\para{Interactive explanations and appropriate trust in AI predictions.}
We find that interactive explanations improve human perception of AI assistance but fail to improve the performance of \humanaiteams.
While the idea of interactive explanations is exciting, our implementation of interactive explanations seems insufficient.
That said, our results suggest future directions for interactive explanations:
1) detecting \ood examples and helping users calibrate their trust \ind and \ood (e.g., by suggesting how similar an example is to the training set);
2) automatic counterfactual suggestions~\citep{wachter2017counterfactual} to help users navigate the decision boundary as it might be difficult for decision makers to come up with counterfactuals on their own;
3) disagreement-driven assistance that frames the decision as to whether to agree with AI predictions or not and help decision makers explore features accordingly.

Meanwhile, we show that interactive explanations may reinforce human biases.
While this observation is preliminary and further work is required to understand the effect of interactive explanations on human biases, this concern is consistent with prior work showing that explanations, including random ones, may improve people's trust in AI predictions~\citep{lai+tan:19,bansal2021does,green2019disparate,green2019principles}.
Therefore, it is important to stay cautious about the potential drawback of interactive explanations and help humans not only detect issues in AI predictions but also reflect biases from themselves.
Future work is required to justify these interactive explanations to be deployed to support human decision making.

\para{Choice of tasks and the complexity of interpreting findings in human-AI decision making.} Our work suggests tasks can play an important role and it can be challenging to understand the generalizability of findings across tasks.
We observe intriguing differences with respect to human agreement with AI predictions between recidivism prediction and \biostask.
A surprising finding is that humans agree with AI predictions more \ood than \ind in \biostask, despite that AI performs worse \ood than \ind.
Furthermore, there exists an asymmetry of human agreement with AI predictions when comparing \shortood with \shortind: the reduced performance gap \ood in recidivism prediction is because humans are less likely to agree with {\bf incorrect} predictions \shortood than \shortind, 
but the reduced performance gap in \biostask is due to that humans are more likely to agree with {\bf correct} AI predictions \shortood than \shortind.
This asymmetry indicates that humans perform better relatively with AI \shortood than \shortind for different reasons in different tasks. 
One possible interpretation of this observation is that humans can complement AI in different ways in different tasks.
To best leverage human insights, it may be useful to design appropriate interfaces that guide humans to find reasons to respectively reject AI predictions or accept AI predictions.

Moreover, 
by exploring tasks with different performance gaps, our results
suggest that comparable performance alone might not be sufficient for complementary performance, echoing the discussion in \citet{bansal2021does}. 
These differences could be driven by many possible factors related to tasks, including difficulty levels, performance gap, and human expertise/confidence.
Although these factors render it difficult to assess the generalizability of findings across tasks,
it is important to explore the diverse space and understand how the choice of tasks may induce different results in the emerging area of human-AI interaction.
We hope that our experiments provide valuable samples for future studies to explore the question of what tasks should be used
and how findings would generalize in the context of human-AI decision making.

Our choice of tasks is aligned with the discovering mode proposed in \citet{lai+liu+tan:20}, where AI can identify counterintuitive patterns and humans may benefit from AI assistance beyond efficiency.
In contrast, humans define the labels in tasks such as question answering and object recognition in the emulating mode, in which case improving performance is essentially improving the quality of data annotation.
We argue that improvement in these two cases can be qualitatively different.

We include recidivism prediction because of its societal importance.
One might argue that complementary performance is not achieved because crowdworkers are not representative of decision makers in this task (i.e., judges) and recidivism prediction might be too difficult for humans.
Indeed, crowdworkers are not the best demographic for recidivism prediction and lack relevant experieince compared to judges.
That said, we hypothesized that complementary performance is possible in recidivism prediction 
because
1) humans and AI show comparable performance, in fact $<$1\% \ood (as a result, the bar to exceed AI performance \ood is quite low and the absolute performance is similar to LSAT in \citet{bansal2021does});
2) prior studies have developed valuable insights on this task with mechanical turkers \citep{green2019disparate,green2019principles} and mechanical turkers outperform random guessing, indicating that they can potentially offer valuable insights, despite their lack of experience compared to judges.
Therefore, we believe that this was a reasonable attempt, although it is possible that the performance of judge-AI teams would differ.
As for the difficulty of this task, it is useful to note that this task is challenging for judges as well.
This difficulty might have contribued to the elusiveness of complementary performance, but is also why it is
especially important to improve human performance in these challenging tasks where human performance is low, ideally while preserving human agency.

To complement recidivism prediction, we chose \biostask because humans including mechanical turkers have strong intuitions about this task and can potentially provide complementary insights from AI.
Indeed, mechanical turkers are more likely to override wrong AI predictions in \biostask than in recidivism prediction.
However, the performance gap between AI and humans in \biostask might be too big to count as ``comparable''.
As ``comparable performance'' is a new term, it is difficult to quantify and decide what performance gap constitutes comparable performance.

\para{Model complexity and other limitations.} In this work, we have focused on linear models because they are relatively simp{}le to ``explain''.
However, a growing body of work has shown that ``explaining'' linear models is non-trivial in a wide variety of tasks \citep{lai+liu+tan:20,poursabzi2021manipulating}.
We speculate that the reason is that the relatively simple patterns in linear models are still challenging for humans to make sense of, e.g., why violent crimes are associated with ``will not violate pretrial terms''.
Humans need to infer the reason might be that the consequence is substantial in that scenario.
We expect such challenges to be even more salient for complex deep learning models.
We leave it to future work for examining the role of model complexity in human-AI decision making.

Our limitations in samples of human subjects also apply to our virtual pilot studies.
University students are not necessarily representative of decision makers for each task.
Our findings may depend on the sample population, although it is reassuring that both virtual pilot studies and large-scale, randomized experiments show that humans may not identify important features or effectively use patterns identified by AI.

\section*{Acknowledgments}

We thank anonymous reviewers for their insightful suggestions and comments. 
We thank all members of the Chicago Human+AI Lab for feedbacks on early versions of our website interface. 
All experiments were approved by the University of Colorado IRB (20-0012). 
This work was supported in part by NSF grants IIS-1837986, 2040989, 2125116, and 2125113.

\appendix

\section{Human Performance in Absolute Accuracy}

\begin{figure}[ht]
    \centering
    \begin{subfigure}{0.32\textwidth}
        \includegraphics[width=\textwidth]{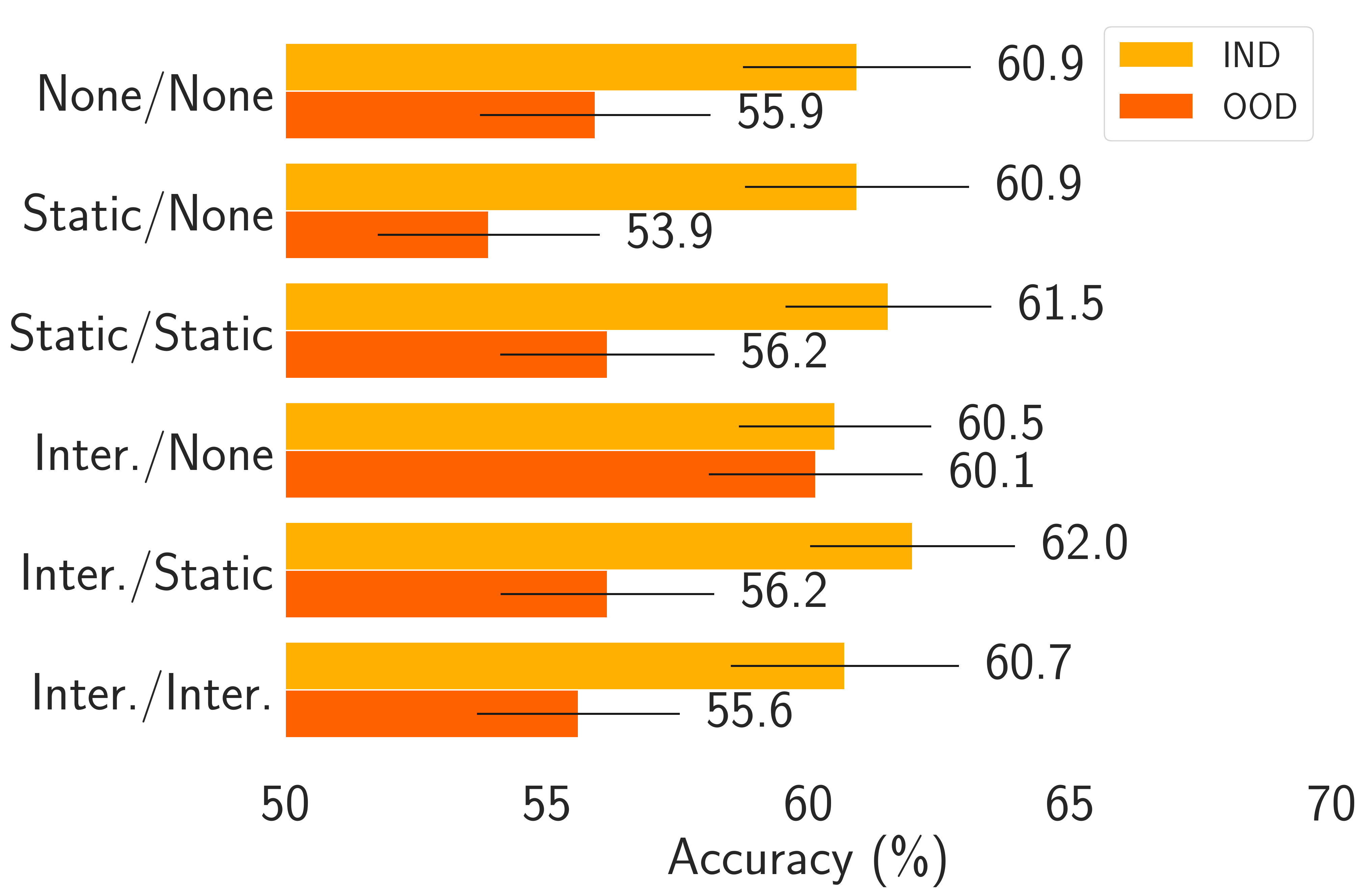}
        \caption{\icpsrtask}
        \label{fig:icpsr_sperf_norm}
    \end{subfigure}
    \begin{subfigure}{0.32\textwidth}
        \includegraphics[width=\textwidth]{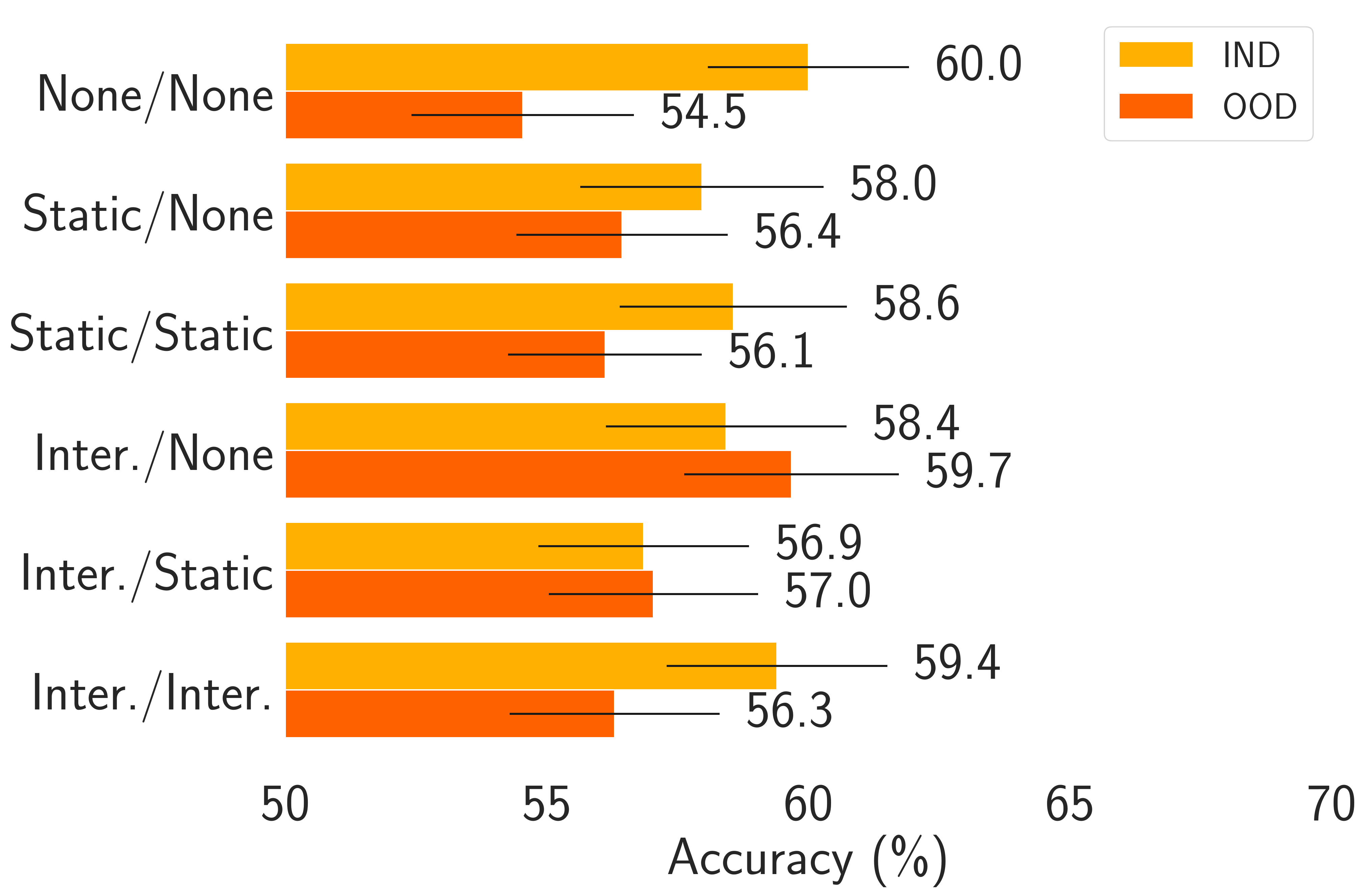}
        \caption{\compastask}
        \label{fig:compas_sperf_norm}
    \end{subfigure}
    \begin{subfigure}{0.32\textwidth}
        \includegraphics[width=\textwidth]{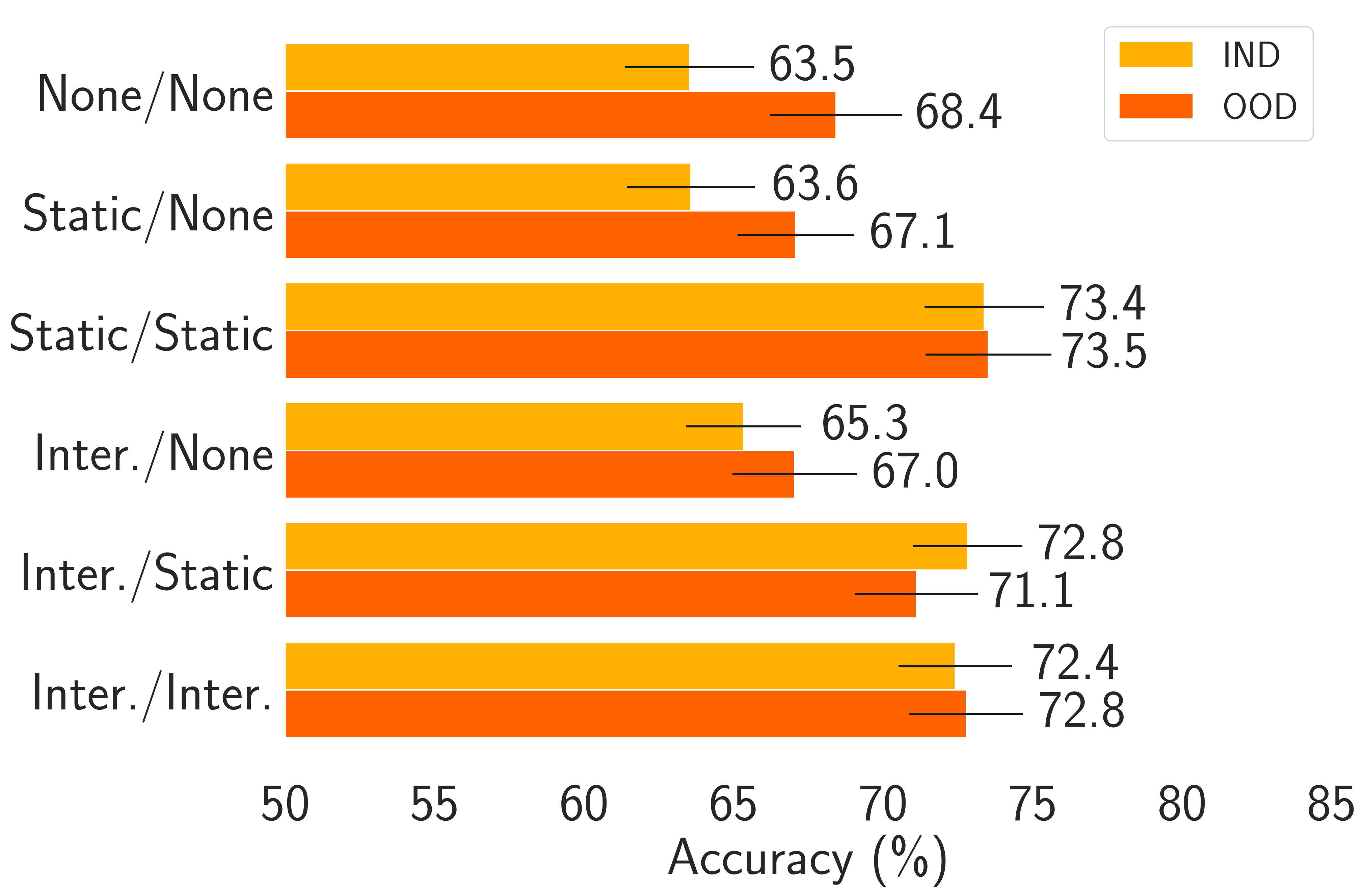}
        \caption{\biostask}
        \label{fig:bios_sperf_norm}
    \end{subfigure}
    \caption{\Humanaiteam performance of different explanation types. \Domain types are indicated by color of the bar and error bars represent 95\% confidence intervals. 
    In \icpsrtask, \humanaiteam performance is significantly higher \ind than \ood in all explanation types ($p<0.01$) except Interactive/None. 
    In \compastask, \ind performance is significantly higher only in None/None ($p<0.005$).
    In \biostask, \ood performance is significantly higher only in None/None ($p<0.01$).
    }
    \label{fig:perf_norm}
    \Description{}
\end{figure}

\figref{fig:perf_norm} shows human performance in absolute accuracy.

\section{COMPAS Figures}

\begin{figure}[ht]
    \centering
        \includegraphics[width=0.4\linewidth]{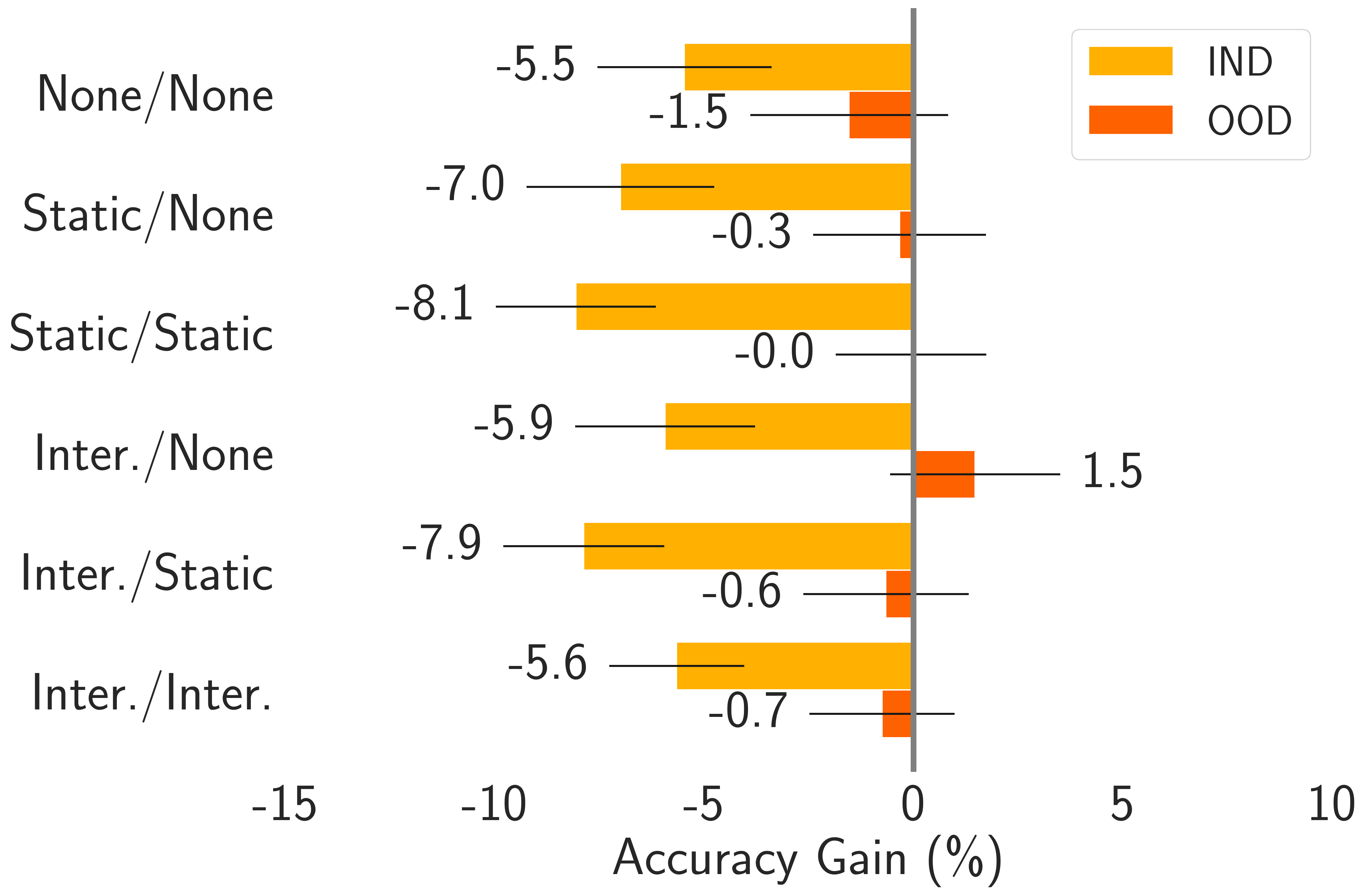}
    \caption{Accuracy gain of different conditions in \compastask.
      \Domain types are indicated by color of the bar and error bars represent 95\% confidence intervals.
      Accuracy gain is only sometimes positive (although not statistically significant). Performance gap between \humanaiteams and AI is significantly smaller in all explanation types except None/None.}
    \label{fig:compas_perf}
    \Description{}
\end{figure}

\begin{figure}[ht]
    \centering
        \includegraphics[width=0.4\linewidth]{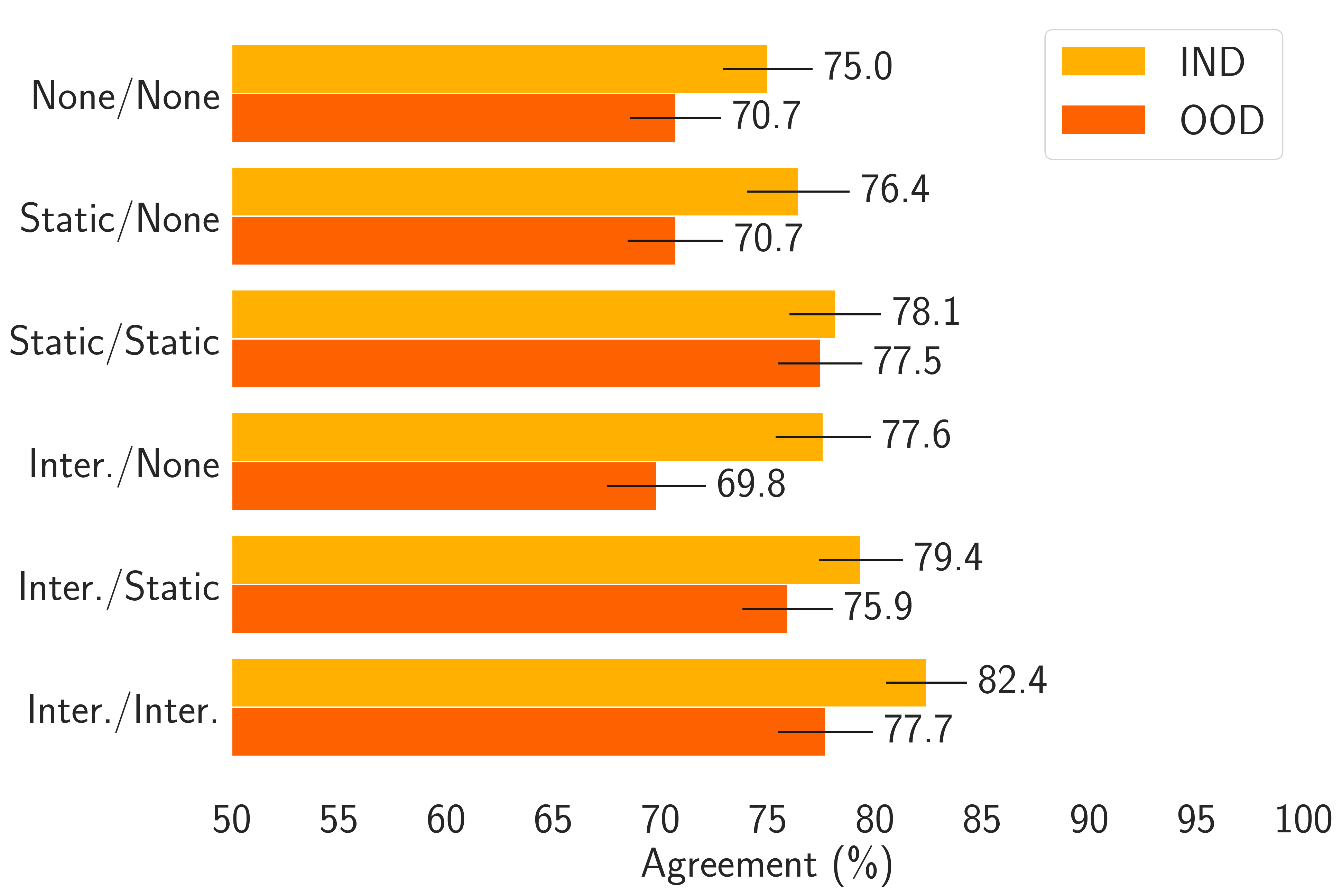}
    \caption{Agreement with AI predictions of different conditions in \compastask. \Domain types are indicated by color of the bar and error bars represent 95\% confidence intervals.
    As compared to \biostask, agreement with AI predictions is much higher \ind than \ood in all explanation types except Static/Static and Interactive/Static.}
    \label{fig:compas_agree}
    \Description{}
\end{figure}

\begin{figure}[ht]
    \centering
        \includegraphics[trim=0 0 0 0, clip, width=0.6\textwidth]{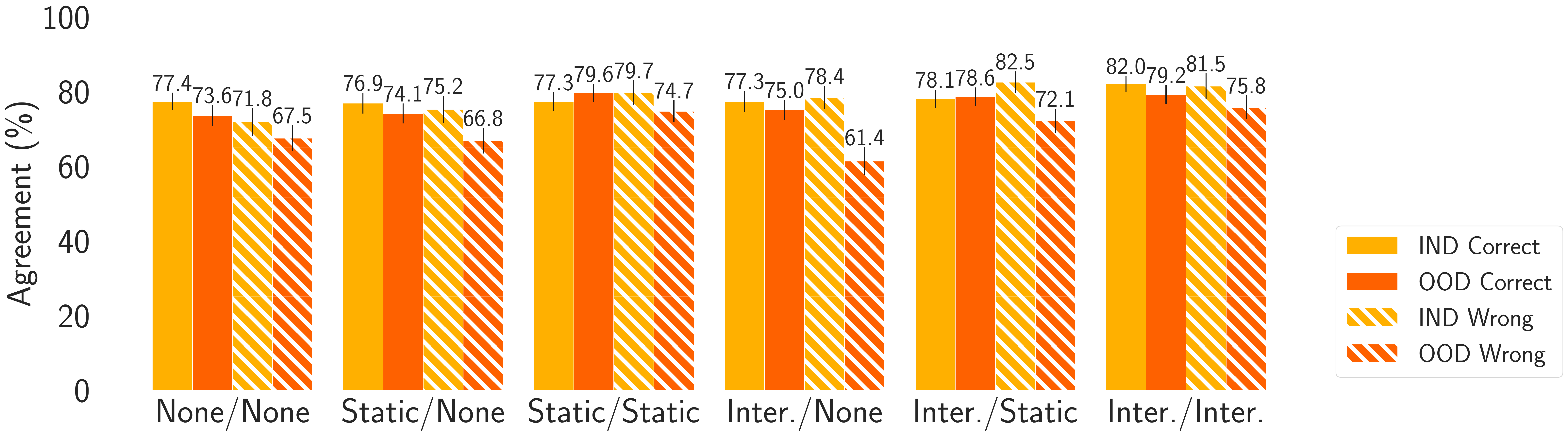}
    \caption{Agreement with AI grouped by \domain type and whether AI predictions are correct in \compastask.
    \Domain types are indicated by color of the bar, bars with stripes represent wrong AI predictions, and error bars represent 95\% confidence intervals.
    \humanaiteams are only more likely to agree with correct AI predictions \ood for all explanation types except None/None.}
    \label{fig:compas_correct}
    \Description{}
\end{figure}

\begin{figure}[ht]
    \centering
        \includegraphics[trim=0 0 0 0, clip, width=0.6\textwidth]{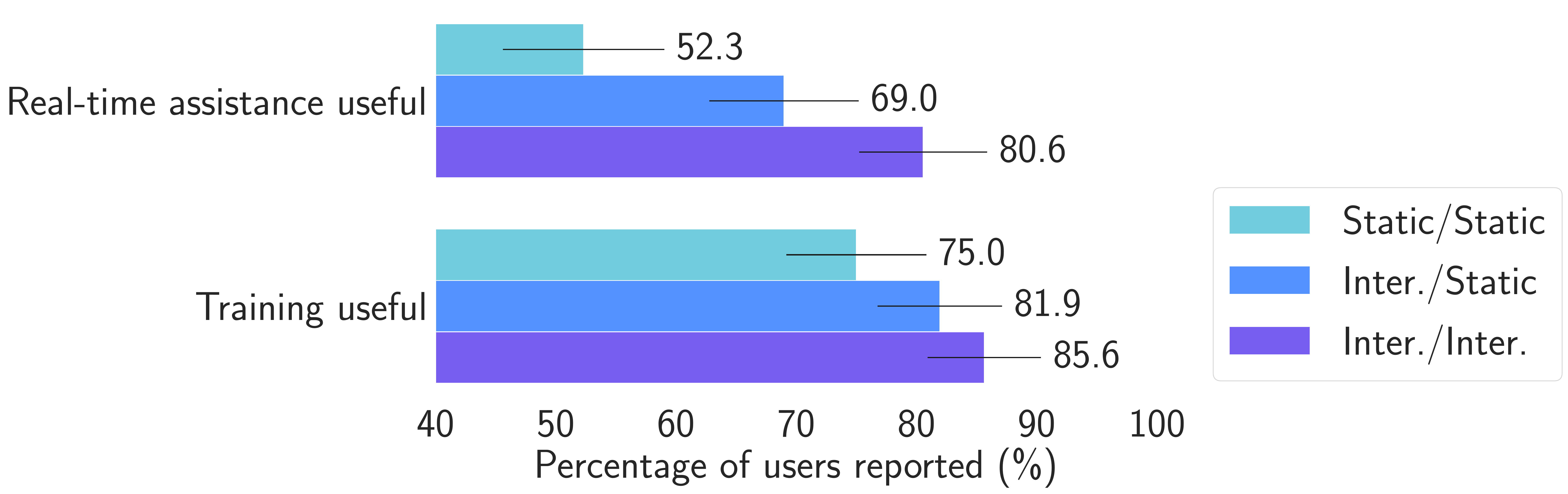}
    \caption{Subjective perception on whether real-time assistance is useful and whether training is useful. $x$-axis shows the percentage of users that answered affirmatively.}
    \label{fig:compas_subjective}
    \Description{}
\end{figure}

\begin{figure}[ht]
    \includegraphics[width=0.48\textwidth]{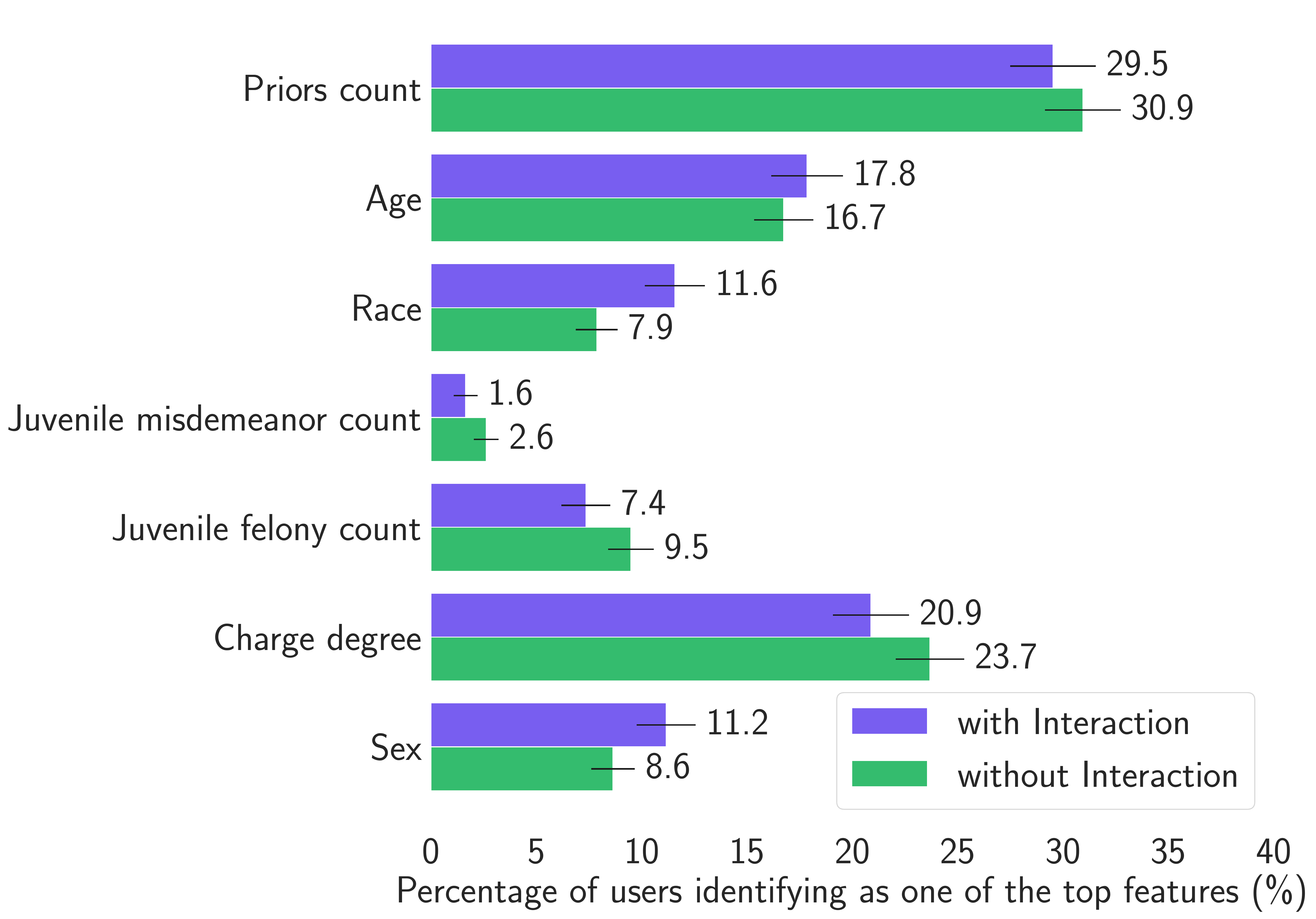}
      \caption{Percentage of users finding a feature important in \compastask. The features are sorted in decreasing order from top to bottom by their Spearman correlation with groundtruth labels.}
      \label{fig:compas_top_features}
      \Description{}
\end{figure}

We also present the figures related to our hypotheses and results for \compastask. The accuracy gain in \compastask is shown in \figref{fig:compas_perf}. The agreement and agreement by correctness are shown in \figref{fig:compas_agree} and \figref{fig:compas_correct}. The subjective perception on whether real-time assistance is useful and whether training is useful is shown in \figref{fig:compas_subjective}.
\figref{fig:compas_top_features} shows the percentage of participants who rate a feature important.

\clearpage

\section{\Ind vs. \Ood Setup}

\begin{figure}[t]
  \centering
  \begin{subfigure}{0.6\textwidth}
    \includegraphics[width=.48\textwidth]{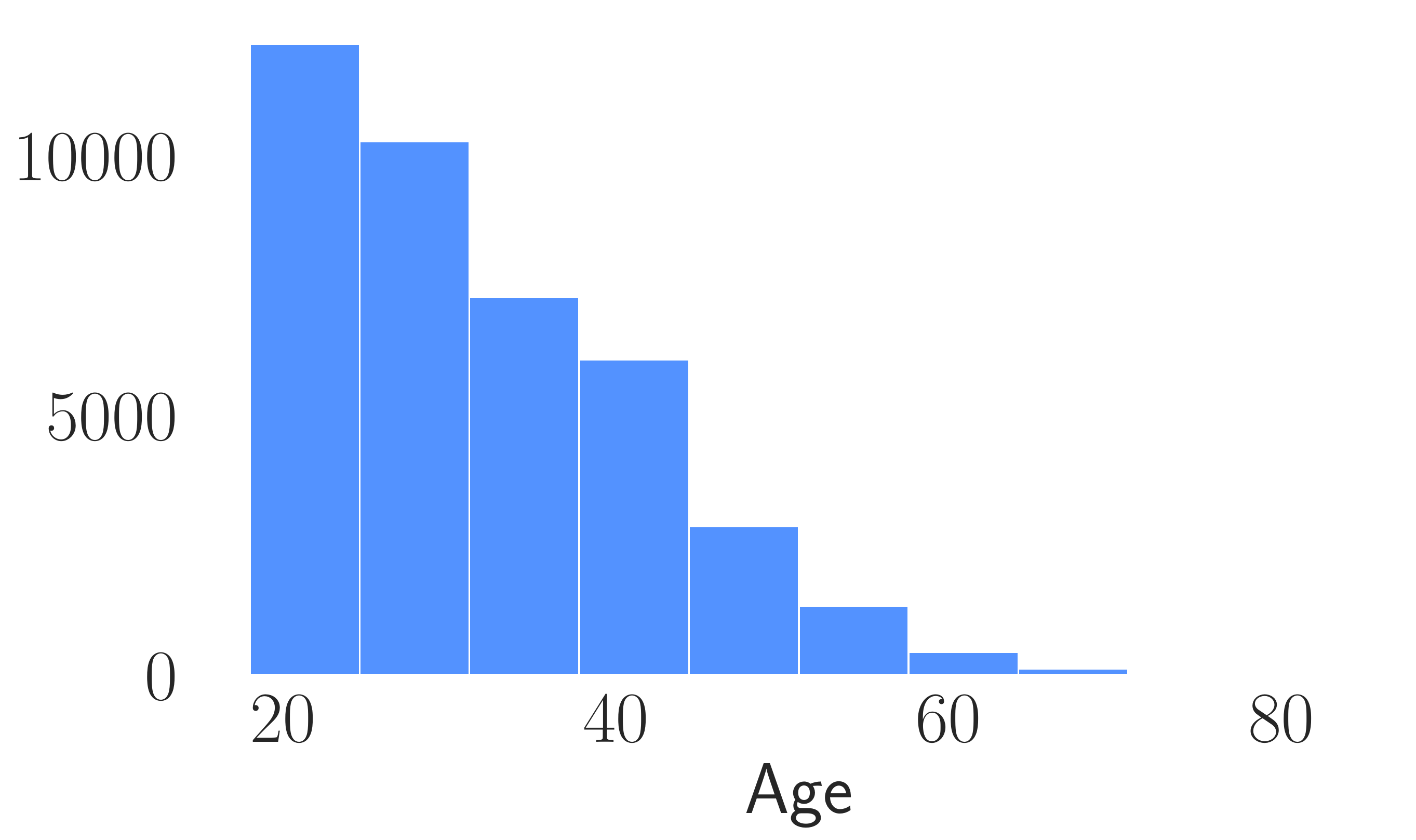}
    \includegraphics[width=.48\textwidth]{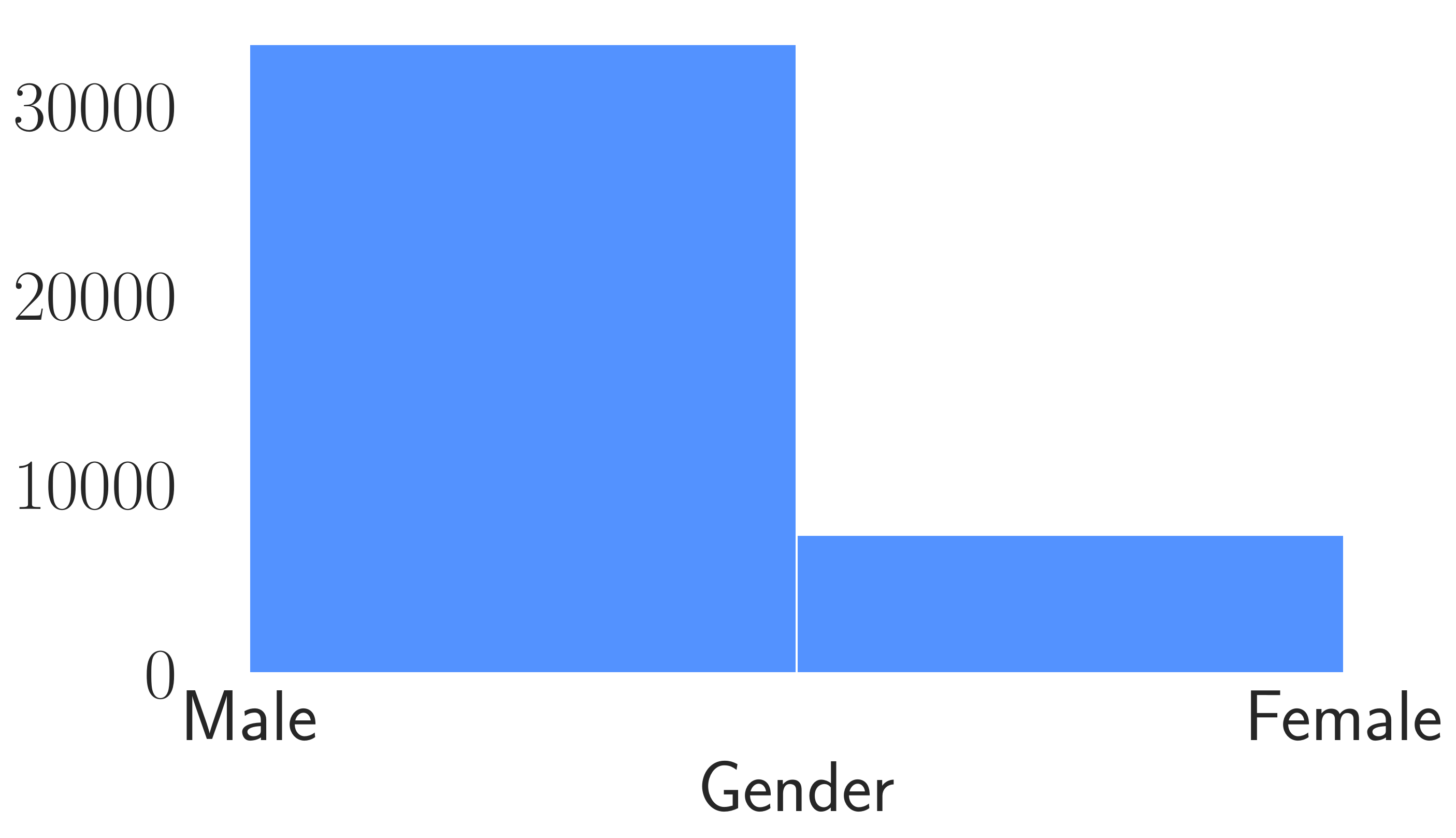}
    \includegraphics[width=.48\textwidth]{figs/icpsr_feature_hist_3.pdf}
    \includegraphics[width=.48\textwidth]{figs/icpsr_feature_hist_4.pdf}
    \caption{Histograms of a subset of features in \icpsrtask. We choose Age as the attribute to split the \domain types because it has a relatively uniform distribution.}
    \label{fig:icpsr_feature_hist}
  \end{subfigure}
  \hfill
  \begin{subfigure}{0.35\textwidth}
      \includegraphics[width=\textwidth]{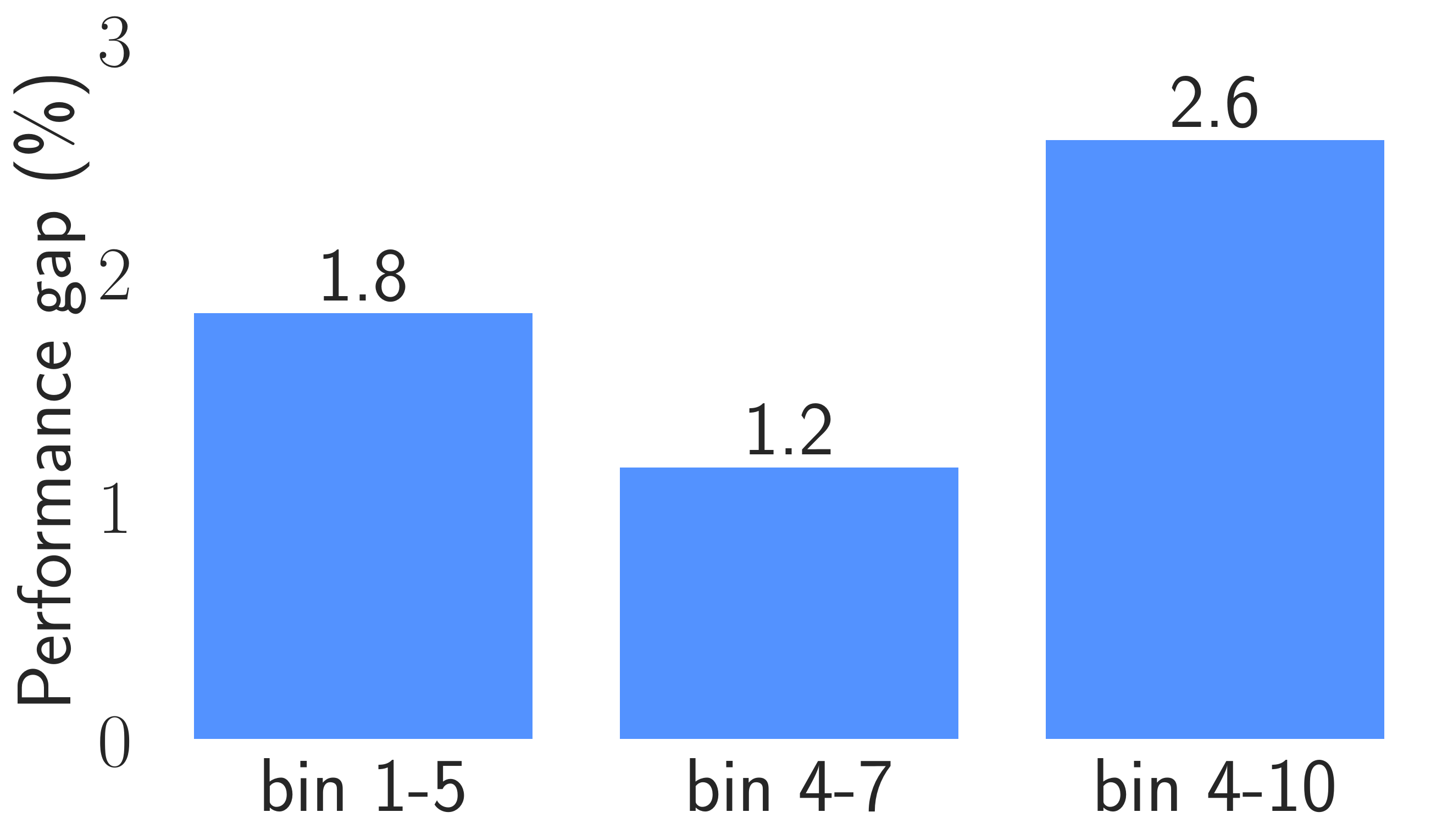}
      \caption{Performance gaps between \ind and \ood examples for each option of \shortind bins in \icpsrtask. Using bin 4-10 as \shortind gives the largest performance gap.}
      \Description{The performance gap for bin 1-5 is 1.8, for bin 4-7 is i.2, and for bin 4-10 is 2.6.}
      \label{fig:icpsr_perf_gap}
  \end{subfigure}
  \caption{Figures for \icpsrtask\ \ind vs. \ood setup.}
  \Description{}
\end{figure}

In this section, we will explain how we split \ind examples and \ood examples in \icpsrtask as an demonstration of the \ind vs. \ood setup procedures.
First, we need to select an attribute for splitting. 
For each candidate attribute, we split the data into 10 bins of equal size based on this attribute. 
We do this because we want to explore different settings of splitting, e.g. different ranges of bins to use for training. 
In other words, we hope to have as much control as possible when we consider which bins are \shortind and which are \shortood. 
For example in \figref{fig:icpsr_feature_hist} we show the histogram of four candidate attributes that we can use to split the examples. 
The distribution is so extreme in Gender and Prior Arrests (too many ``Male'' in Gender and too many ``10'' in Prior Arrests) that if we choose any of these two attributes, we would have no choice but to use nearly half of our data as either \shortind or \shortood, because we want to avoid having the same value in both \domain types. 
Similarly Prior Convictions also limits our choices of bins due to its extreme distribution. 
Since there are too many instances with value ``$0$,'' bin 1 and bin 2 would both consist of defendants who have $0$ prior convictions after binning. 
If we were to use a splitting where bin 1 is \shortind and bin 2 is \shortood, then this splitting does not make sense (one \domain type falls into the other's distribution). Therefore we finally choose Age as the attribute. We also design desiderata 3) for the \ind vs. \ood setup to avoid these situations. 

After selecting the attribute, we also need to decide which bins we use as \ind examples and which bins as \ood examples. In \icpsrtask, the options we explore are:
1) bin 1-5 as \shortind: age $\geq$ 30 as \shortind and age $>$ 30 as \shortood. 
2) bin 4-7 as \shortind: age between 25-36 as \shortind and age $<$ 25 or age $>$ 36 as \shortood; 
3) bin 4-10 as \shortind: age $\geq$ 25 as \shortind and age $<$ 25 as \shortood; 
We finally settled on option 3) because it gives us the largest performance gap between \ind examples and \ood examples (\figref{fig:icpsr_perf_gap}). Note that this performance gap looks different from what we present in Fig. 2 in the main paper because here we use the entire testset (after balancing labels) for evaluation, instead of the 360 randomly sampled examples we prepare for the user study. The \ind examples in the random samples are easier for AI, therefore giving us an even larger performance gap between \ind and \ood.

\section{User Interface Designs}

\begin{figure}[t]
  \centering
  \includegraphics[width=0.8\textwidth]{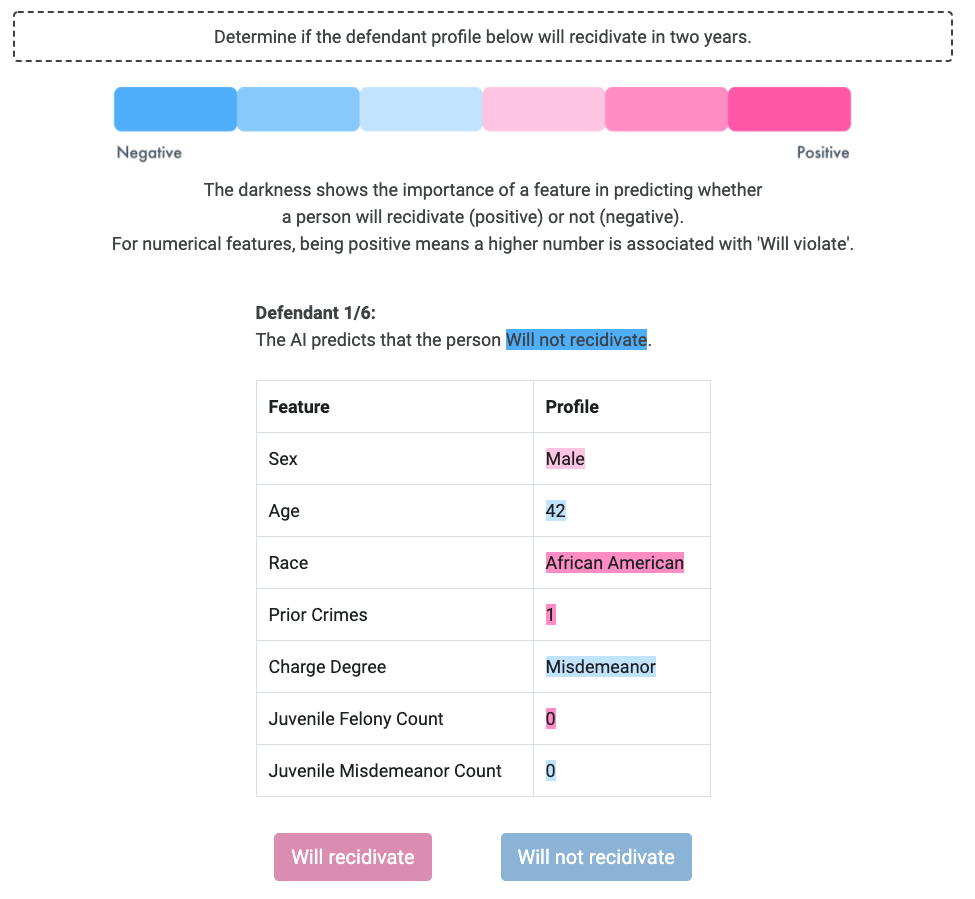}
  \caption{
    Static assistance for \compastask.
  }
  \label{fig:compas_static}
  \Description{}
\end{figure}

\para{Screenshots for static assistance for ~\compastask. }
\figref{fig:compas_static} shows the static assistance for \compastask.

\clearpage

\begin{figure}[t]
  \centering
  \includegraphics[width=0.75\textwidth,page=1,trim=80 0 100 0,clip]{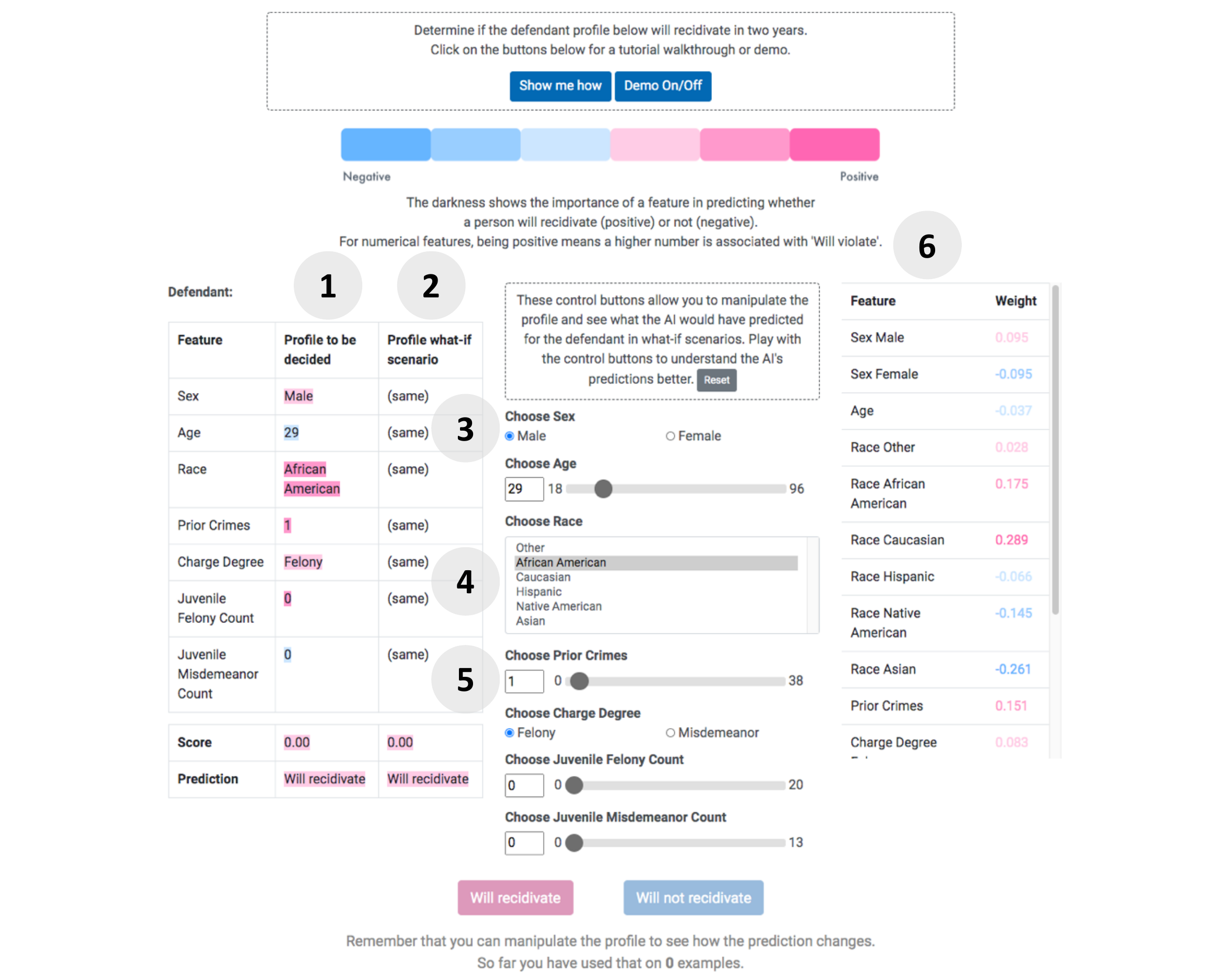}
  \caption{In addition to static assistance such as feature highlights and showing AI predictions, users are able to manipulate the features of defendant's profile to see any changes in the AI prediction. Illustration of interactive console for \compastask: 1) actual defendant's profile; 2) edited defendant's profile if user manipulates any features; 3) user is able to edit the value of {\em Sex} and {\em Charge Degree} with radio buttons; 4) user is able to edit the value of {\em Race} with dropdown; 5) user is able to edit the value {\em Age}, {\em Prior Crimes}, {\em Juvenile Felony Count}, and {\em Juvenile Misdemeanor Count} with sliders; 6) a table displaying features and respective coefficients, the color and darkness of the color shows the importance of a feature in predicting whether a person will recidivate or not.
  }
  \label{fig:compas_screenshot}
  \Description{Picture shows an interactive console that allows user to manipulate feature values with sliders and radio buttons. Changes in feature values may result in a change of the AI prediction. A table on the right of the picture displays feature coefficients.}
\end{figure}

\para{Interactive interface for \compastask.}
\figref{fig:compas_screenshot} shows the interactive interface for \compastask.

\clearpage

\para{Attention check.}
In the recidivism prediction task, many participants found one of the attention-check questions to be very tricky.
As the purpose of the attention-check questions was not to intentionally trick users into answering the wrong answer, we made edits to one of the attention-check questions to remove any confusion.
In addition, many participants felt that it was better if they could refer to the definitions of certain terminology.
As such, we combined the instructions and attention-check questions step in one page so participants are able to look up on the definitions if they had forgotten.
\figref{fig:attention} shows screenshots of attention check questions in all the three tasks. 

\begin{figure}[t]
  \centering
  \begin{subfigure}{\textwidth}
    \centering
    \includegraphics[width=0.8\textwidth]{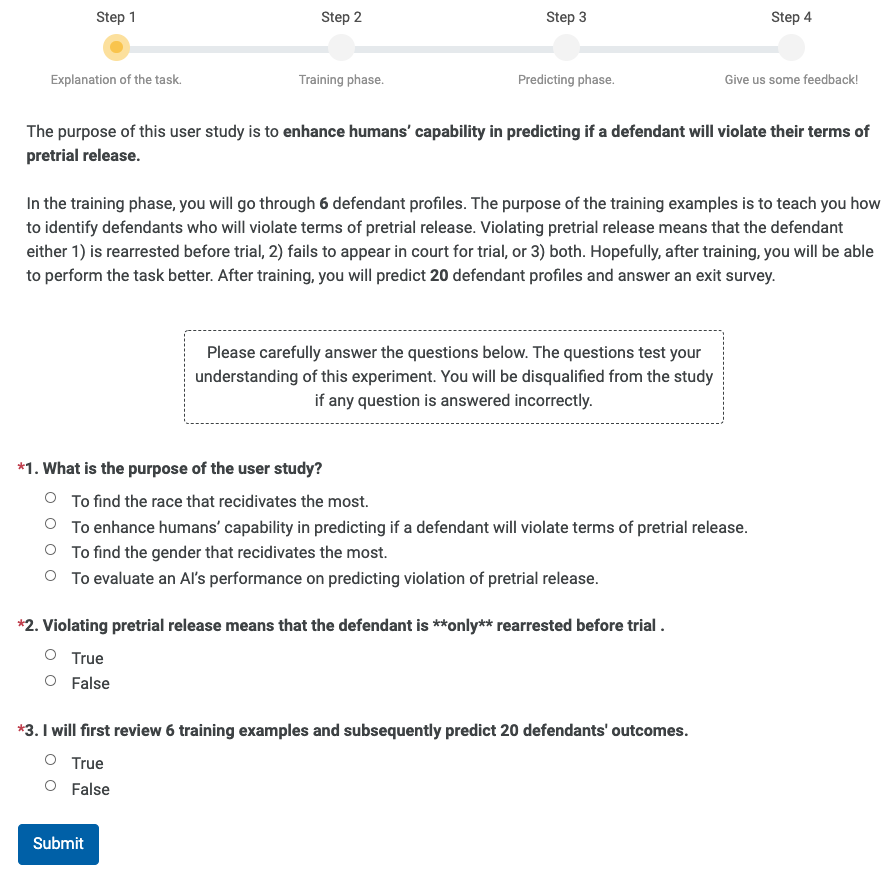}
    \caption{Attention check for \icpsrtask. The user is required to select the correct answers before they are allowed to proceed to the training phase. The answers to the attention check questions can be found in the same page.}
    \label{fig:icpsr_attention_check}
    \Description{}
  \end{subfigure}
\end{figure}

\begin{figure}[t]\ContinuedFloat
    \centering
  \begin{subfigure}{\textwidth}
    \centering
    \includegraphics[width=0.8\textwidth]{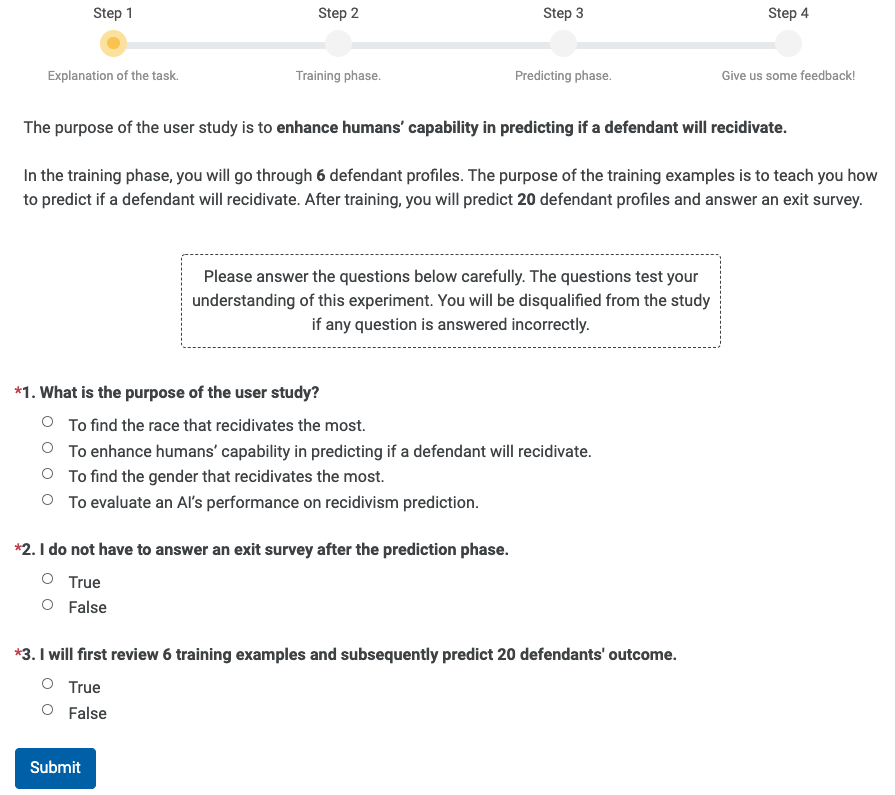}
    \caption{Attention check for \compastask. The user is required to select the correct answers before they are allowed to proceed to the training phase. The answers to the attention check questions can be found in the same page.}
    \label{fig:compas_attention_check}
    \Description{}
  \end{subfigure}
\end{figure}

\begin{figure}[t]\ContinuedFloat
  \centering
  \begin{subfigure}{\textwidth}
    \centering
    \includegraphics[width=0.8\textwidth]{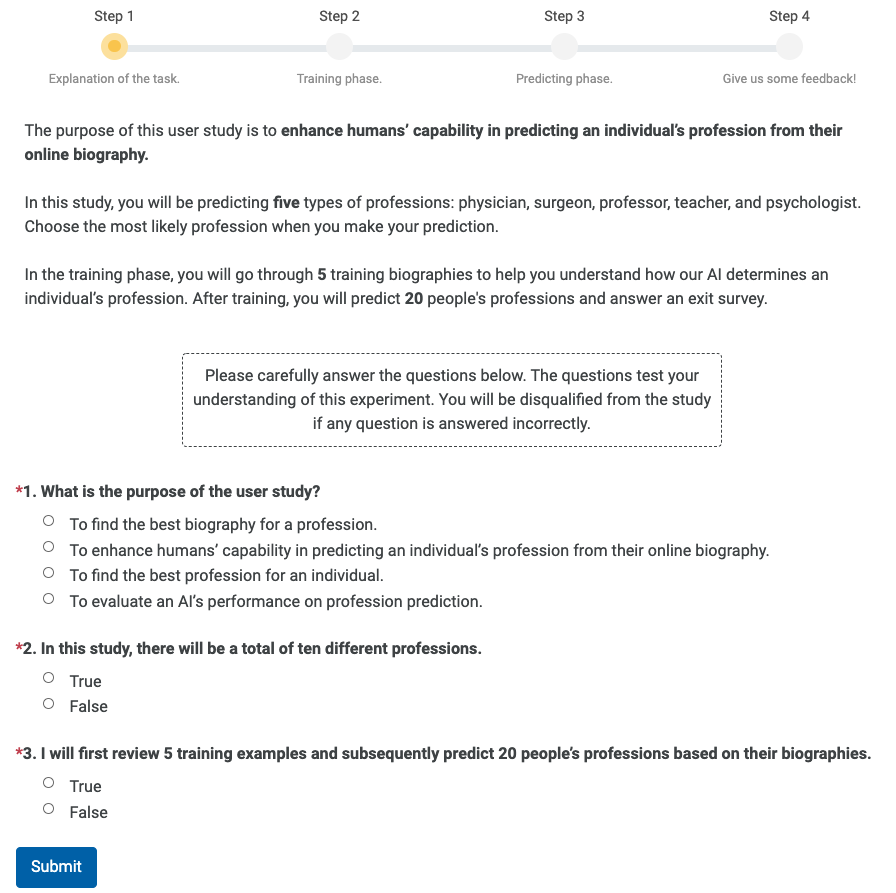}
    \caption{Attention check for \biostask. The user is required to select the correct answers before they are allowed to proceed to the training phase. The answers to the attention check questions can be found in the same page.}
    \label{fig:bios_attention_check}
    \Description{}
  \end{subfigure}
  \caption{Attention check questions.}
  \label{fig:attention}
\end{figure}

\clearpage

\para{Feature quiz.}
In the training phase of each task, for all explanation types except None/None, we also design a feature quiz to see if users understand the association between features and labels correctly.
For each training instance in the training phase, we prompt users the quiz as in \figref{fig:feature_quiz} after they make the prediction.
We ask users to identify the positive and negative feature from two candidate features.
The correct candidate is prepared by a random sampling from all the features that are currently shown in the interface, while the incorrect candidate is sampled from all features that do not have the correct polarity as prompted.
The submit button is disabled for five seconds starting from the appearance of the check to refrain users from submitting a random answer.

\begin{figure}[t]
  \centering
  \begin{subfigure}{\textwidth}
    \centering
    \includegraphics[width=0.8\textwidth]{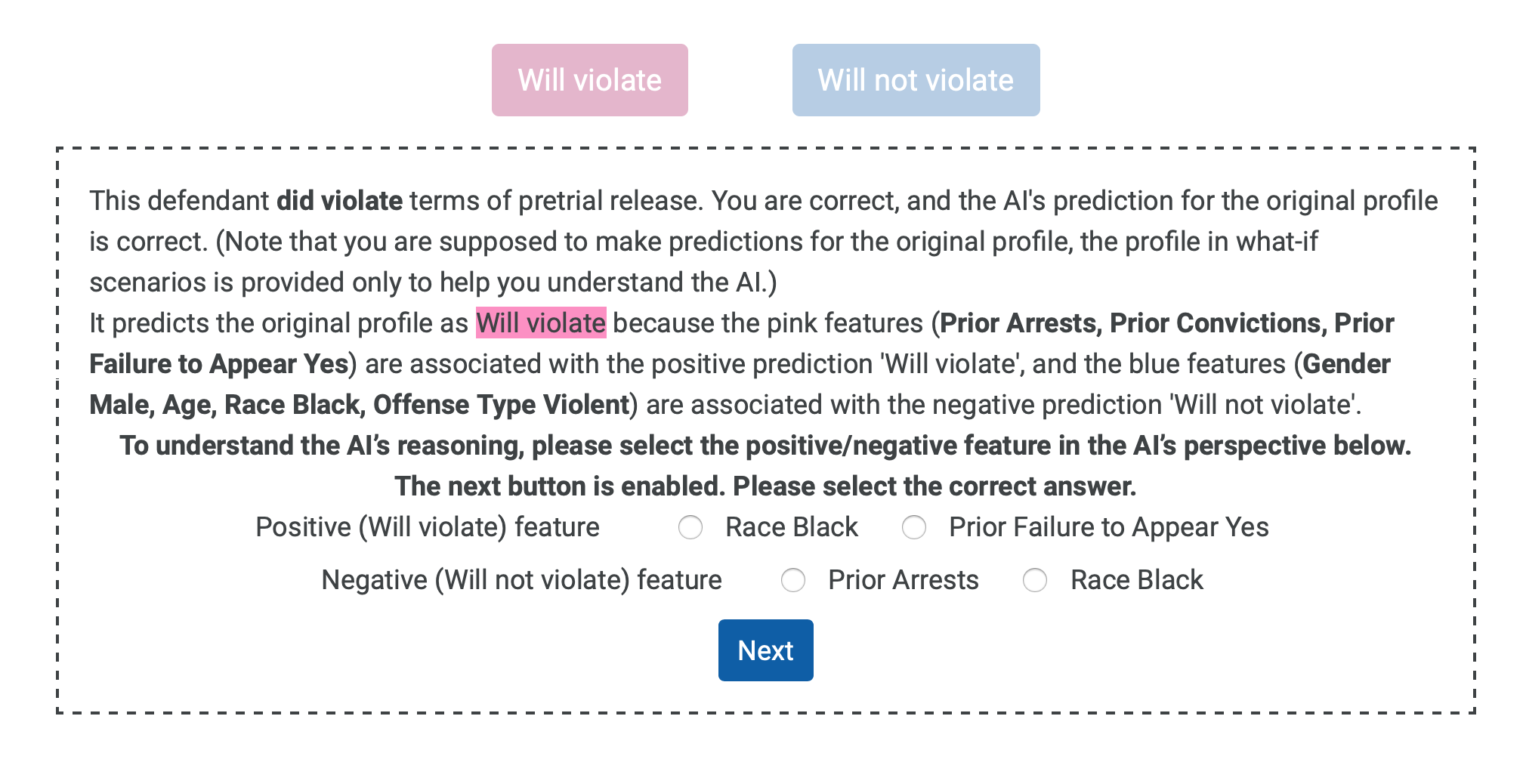}
    \caption{Features quiz for \icpsrtask.  The user is required to select the correct positive and negative feature before they are allowed to proceed to the next instance. In this example, the correct answer for positive feature is {\em Prior Failure to Appear Yes}, and the correct answer for negative feature is {\em Race Black}.}
    \label{fig:icpsr_feature_check}
    \Description{}
  \end{subfigure}
  \hfill
  \begin{subfigure}{\textwidth}
    \centering
    \includegraphics[width=0.8\textwidth]{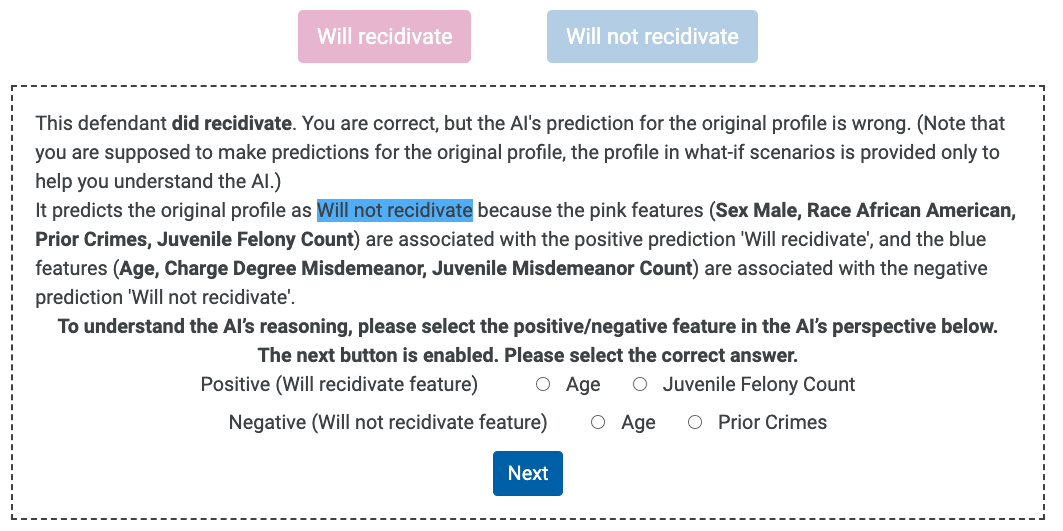}
    \caption{Features quiz for \compastask. The user is required to select the correct positive and negative feature before they are allowed to proceed to the next instance. In this example, the correct answer for positive feature is {\em Juvenile Felony Count}, and the correct answer for negative feature is {\em Age}.}
    \label{fig:compas_feature_check}
    \Description{}
  \end{subfigure}
\end{figure}

\begin{figure}\ContinuedFloat
  \begin{subfigure}{\textwidth}
    \centering
    \includegraphics[width=0.8\textwidth]{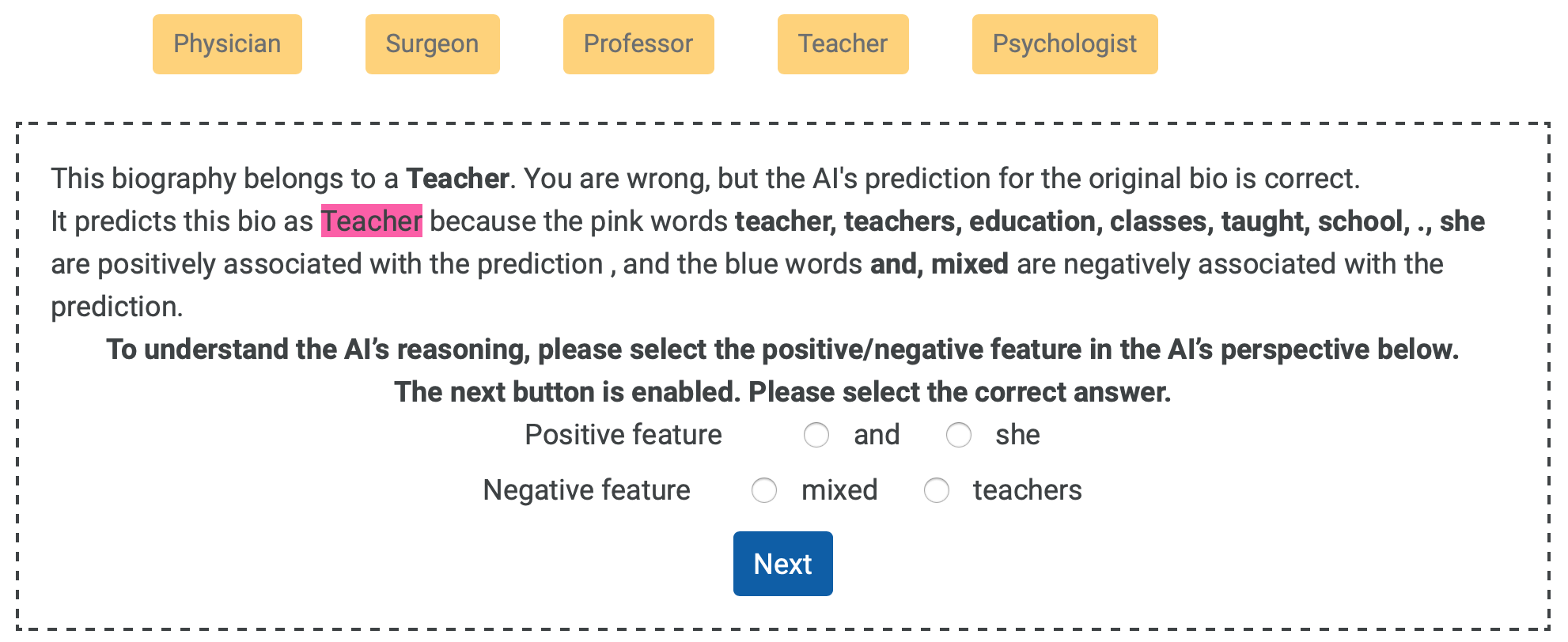}
    \caption{Features quiz for \biostask. The user is required to select the correct positive and negative feature before they are allowed to proceed to the next instance. In this example, the correct answer for positive feature is {\em she}, and the correct answer for negative feature is {\em mixed}.}
    \label{fig:bios_feature_check}
    \Description{}
  \end{subfigure}
  \caption{Feature quiz.}
  \label{fig:feature_quiz}
\end{figure}

\clearpage

\begin{figure}[t]
    \centering
    \begin{subfigure}{0.32\textwidth}
        \includegraphics[width=\textwidth]{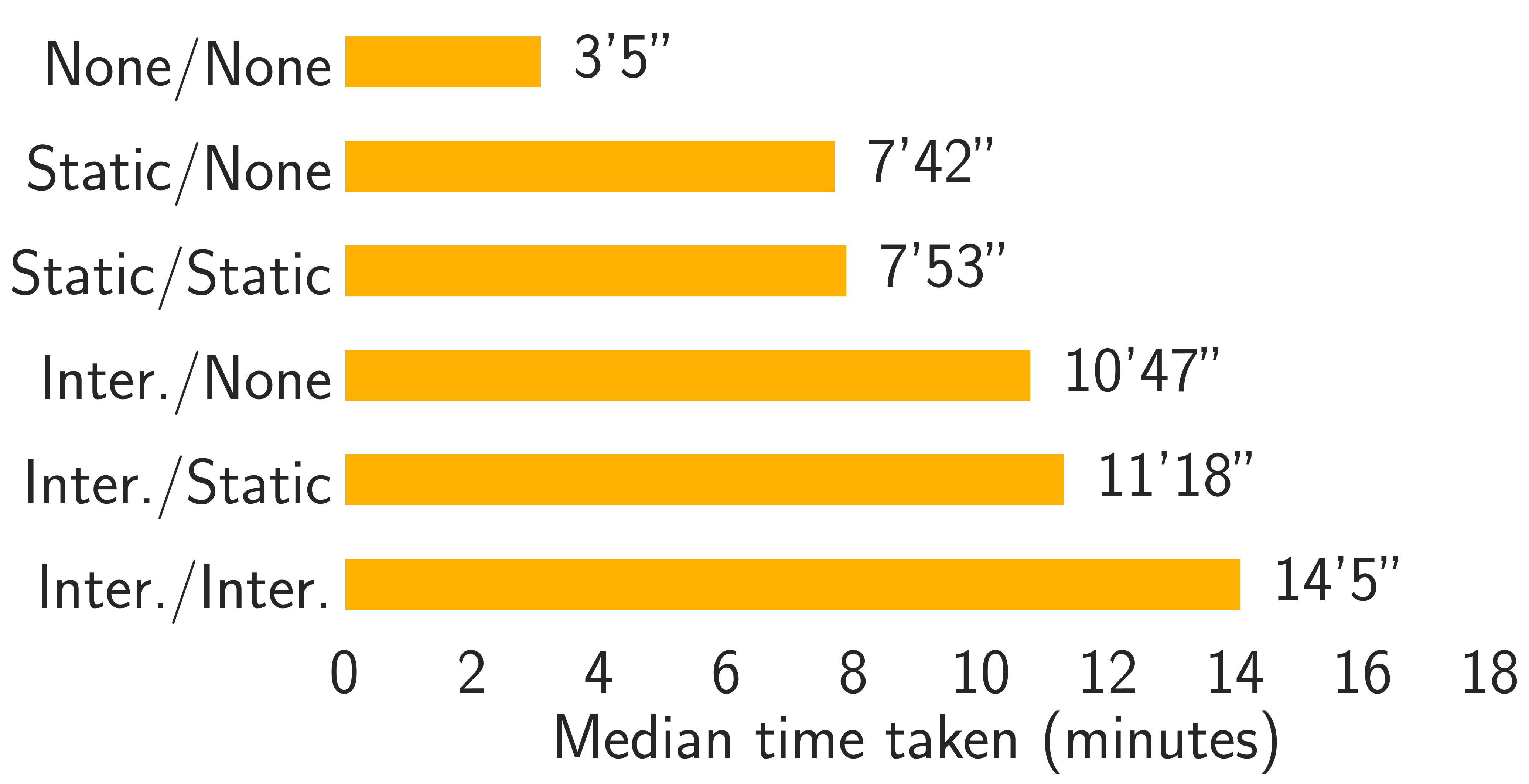}
        \caption{Median time taken by users in \icpsrtask.}
        \Description{Median time taken for each explanation type are 3'5'', 7'42'', 7'53'', 10'47'', 11'18'', and 14'5''.}
        \label{fig:icpsr_time_taken}
    \end{subfigure}
    \hfill
    \begin{subfigure}{0.32\textwidth}
        \includegraphics[width=\textwidth]{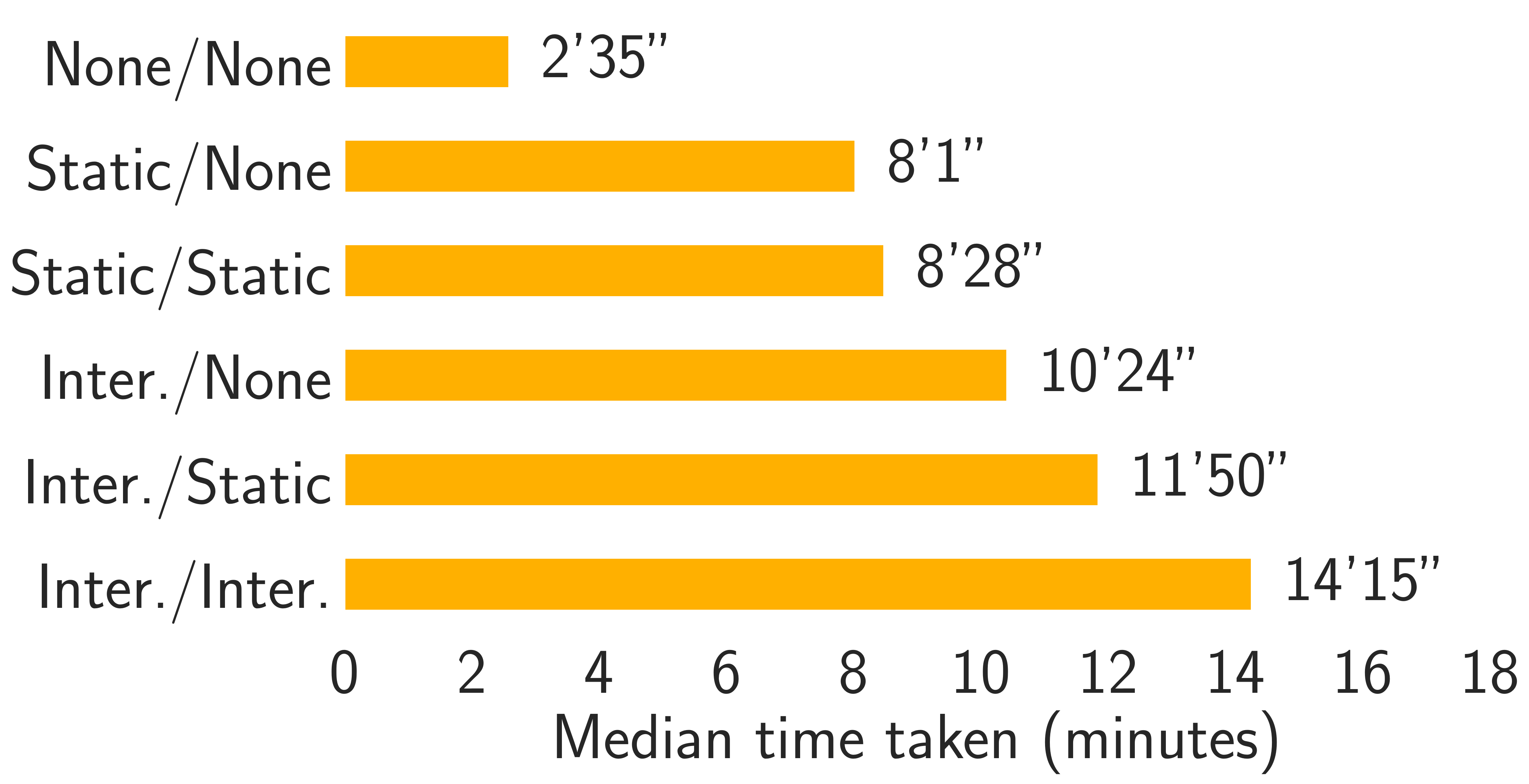}
        \caption{Median time taken by users in \compastask.}
        \Description{Median time taken for each explanation type are 2'35'', 8'1'', 8'28'', 10'24'', 11'50'', and 14'15''.}
        \label{fig:compas_time_taken}
    \end{subfigure}
    \begin{subfigure}{0.32\textwidth}
        \includegraphics[width=\textwidth]{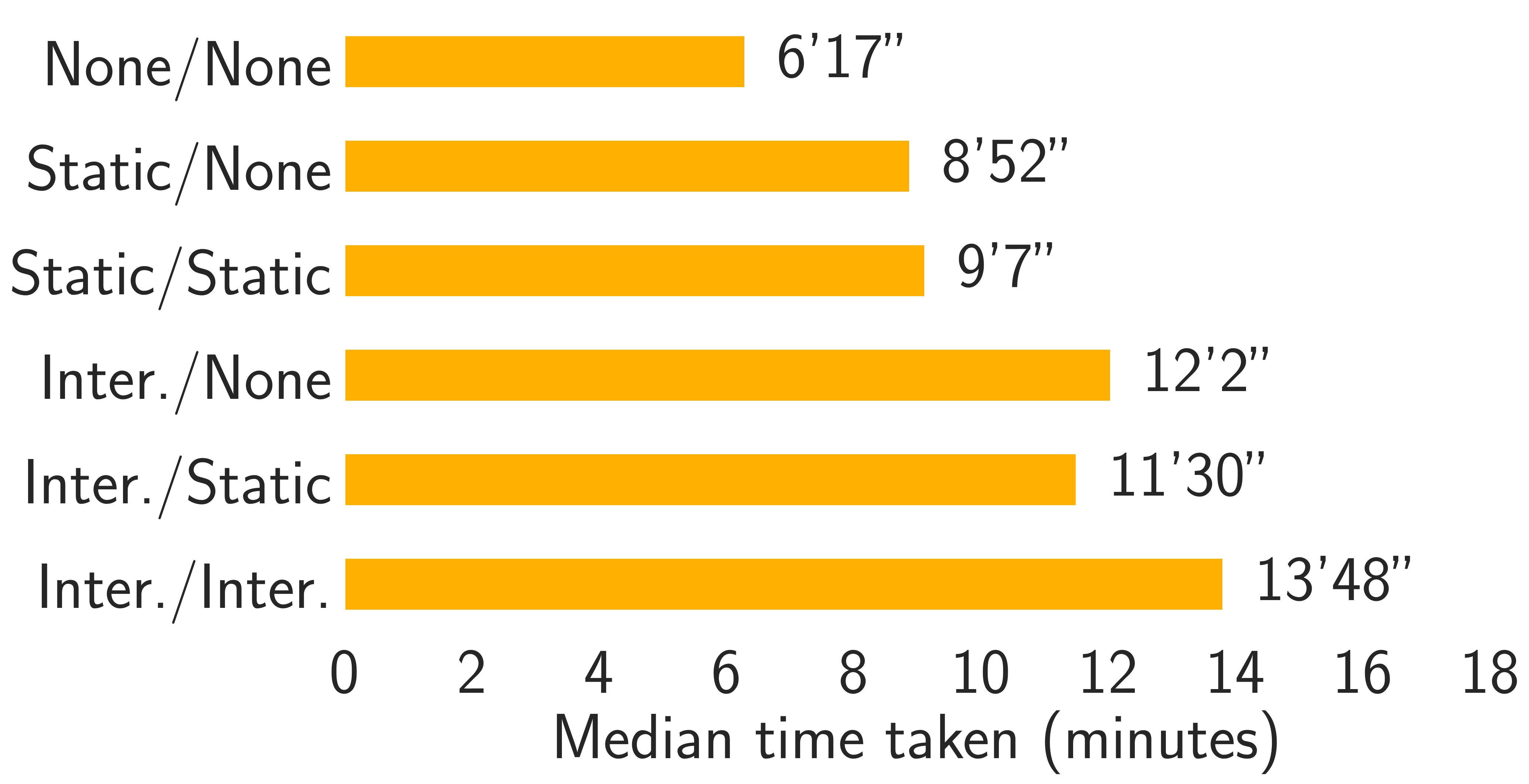}
        \caption{Median time taken by users in \biostask.}
        \Description{Median time taken for each explanation type are 6'17'', 8'52'', 9'7'', 12'2'', 11'30'', and 13'48''.}
        \label{fig:bios_time_taken}
    \end{subfigure}
    \caption{Median of time taken by MTurk users in each explanation type.}
    \label{fig:time_taken}
    \Description{}
\end{figure}

\para{Details for experiments on Mechanical Turk.}
We report the median time taken by the users to complete each task.
The median time taken for \icpsrtask, \compastask, and \biostask are 9'55'', 9'16'', and 8'59'' respectively.
In \figref{fig:time_taken}, we show the median time taken for each explanation type.
We are reporting the median time taken due to a few outliers in the data collected where user is inactive for a long period of time during the study.

\section{Survey Questions}

\begin{figure}[t]
  \centering
  \begin{subfigure}{\textwidth}
    \centering
    \includegraphics[width=\textwidth]{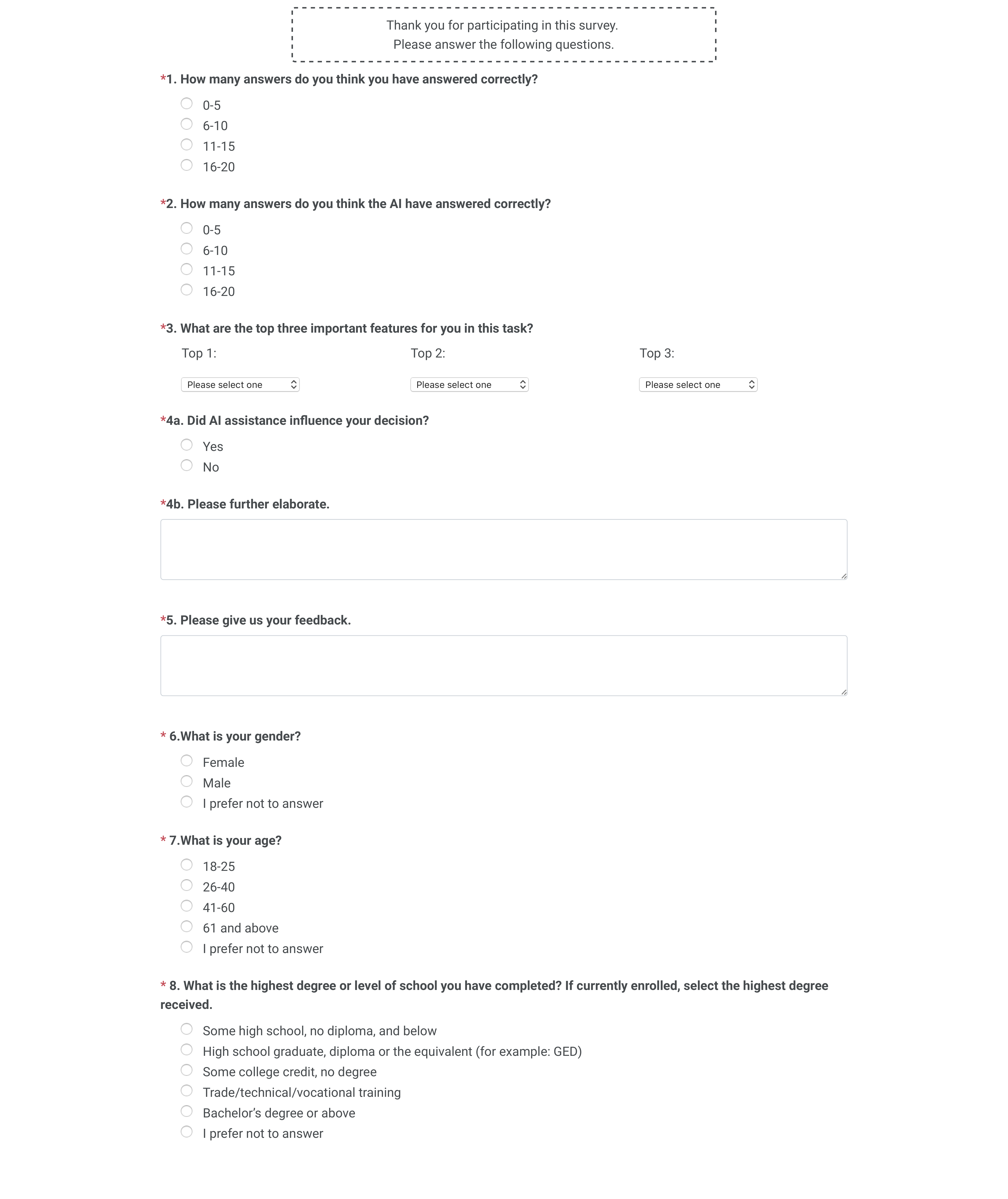}
    \caption{Survey questions for \icpsrtask and \compastask.}
    \label{fig:compas_icpsr_survey}
    \Description{}
  \end{subfigure}
\end{figure}

\begin{figure}[t]\ContinuedFloat
  \centering
  \begin{subfigure}{\textwidth}
    \centering
    \includegraphics[width=\textwidth]{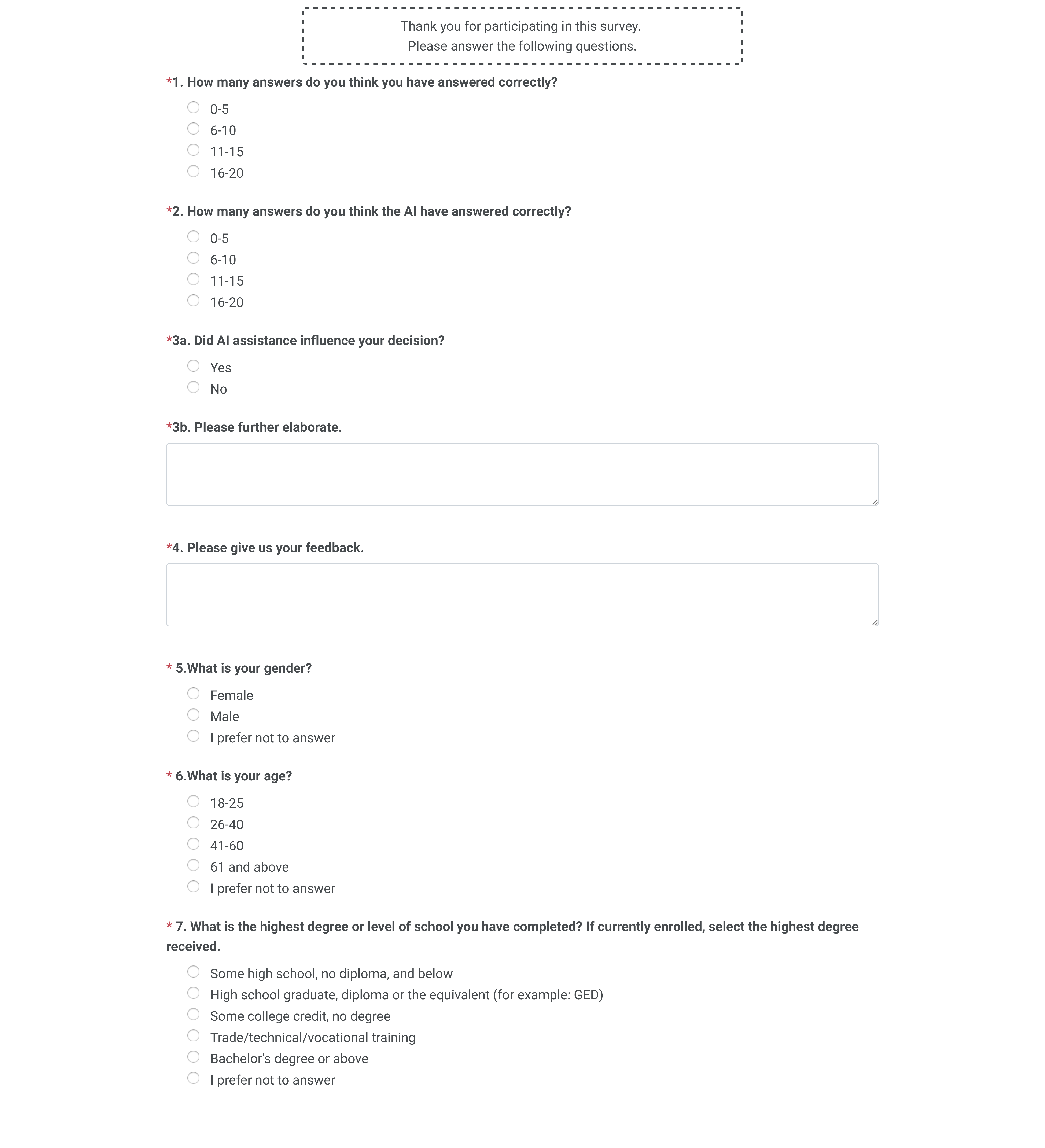}
    \caption{Survey questions for \biostask.}
    \label{fig:bios_survey}
    \Description{}
  \end{subfigure}
  \caption{Survey questions.}
  \label{fig:survey}
\end{figure}

\clearpage

\bibliographystyle{ACM-Reference-Format}
\bibliography{refs}


\begin{thebibliography}{74}


\ifx \showCODEN    \undefined \def \showCODEN     #1{\unskip}     \fi
\ifx \showDOI      \undefined \def \showDOI       #1{#1}\fi
\ifx \showISBNx    \undefined \def \showISBNx     #1{\unskip}     \fi
\ifx \showISBNxiii \undefined \def \showISBNxiii  #1{\unskip}     \fi
\ifx \showISSN     \undefined \def \showISSN      #1{\unskip}     \fi
\ifx \showLCCN     \undefined \def \showLCCN      #1{\unskip}     \fi
\ifx \shownote     \undefined \def \shownote      #1{#1}          \fi
\ifx \showarticletitle \undefined \def \showarticletitle #1{#1}   \fi
\ifx \showURL      \undefined \def \showURL       {\relax}        \fi
\providecommand\bibfield[2]{#2}
\providecommand\bibinfo[2]{#2}
\providecommand\natexlab[1]{#1}
\providecommand\showeprint[2][]{arXiv:#2}

\bibitem[\protect\citeauthoryear{Angwin, Larson, Mattu, and Kirchner}{Angwin
  et~al\mbox{.}}{2016}]%
        {angwin2016machine}
\bibfield{author}{\bibinfo{person}{Julia Angwin}, \bibinfo{person}{Jeff
  Larson}, \bibinfo{person}{Surya Mattu}, {and} \bibinfo{person}{Lauren
  Kirchner}.} \bibinfo{year}{2016}\natexlab{}.
\newblock \bibinfo{title}{Machine Bias}.
\newblock
\newblock


\bibitem[\protect\citeauthoryear{Bansal, Wu, Zhou, Fok, Nushi, Kamar, Ribeiro,
  and Weld}{Bansal et~al\mbox{.}}{2021}]%
        {bansal2021does}
\bibfield{author}{\bibinfo{person}{Gagan Bansal}, \bibinfo{person}{Tongshuang
  Wu}, \bibinfo{person}{Joyce Zhou}, \bibinfo{person}{Raymond Fok},
  \bibinfo{person}{Besmira Nushi}, \bibinfo{person}{Ece Kamar},
  \bibinfo{person}{Marco~Tulio Ribeiro}, {and} \bibinfo{person}{Daniel Weld}.}
  \bibinfo{year}{2021}\natexlab{}.
\newblock \showarticletitle{Does the whole exceed its parts? the effect of ai
  explanations on complementary team performance}. In
  \bibinfo{booktitle}{\emph{Proceedings of the 2021 CHI Conference on Human
  Factors in Computing Systems}}. \bibinfo{pages}{1--16}.
\newblock


\bibitem[\protect\citeauthoryear{Beede, Baylor, Hersch, Iurchenko, Wilcox,
  Ruamviboonsuk, and Vardoulakis}{Beede et~al\mbox{.}}{2020}]%
        {beede2020human}
\bibfield{author}{\bibinfo{person}{Emma Beede}, \bibinfo{person}{Elizabeth
  Baylor}, \bibinfo{person}{Fred Hersch}, \bibinfo{person}{Anna Iurchenko},
  \bibinfo{person}{Lauren Wilcox}, \bibinfo{person}{Paisan Ruamviboonsuk},
  {and} \bibinfo{person}{Laura~M Vardoulakis}.}
  \bibinfo{year}{2020}\natexlab{}.
\newblock \showarticletitle{A Human-Centered Evaluation of a Deep Learning
  System Deployed in Clinics for the Detection of Diabetic Retinopathy}. In
  \bibinfo{booktitle}{\emph{Proceedings of the 2020 CHI Conference on Human
  Factors in Computing Systems}}. \bibinfo{pages}{1--12}.
\newblock


\bibitem[\protect\citeauthoryear{Brown and Sandholm}{Brown and
  Sandholm}{2019}]%
        {brown2019superhuman}
\bibfield{author}{\bibinfo{person}{Noam Brown} {and} \bibinfo{person}{Tuomas
  Sandholm}.} \bibinfo{year}{2019}\natexlab{}.
\newblock \showarticletitle{Superhuman AI for multiplayer poker}.
\newblock \bibinfo{journal}{\emph{Science}} \bibinfo{volume}{365},
  \bibinfo{number}{6456} (\bibinfo{year}{2019}), \bibinfo{pages}{885--890}.
\newblock


\bibitem[\protect\citeauthoryear{Bu{\c{c}}inca, Lin, Gajos, and
  Glassman}{Bu{\c{c}}inca et~al\mbox{.}}{2020}]%
        {buccinca2020proxy}
\bibfield{author}{\bibinfo{person}{Zana Bu{\c{c}}inca}, \bibinfo{person}{Phoebe
  Lin}, \bibinfo{person}{Krzysztof~Z Gajos}, {and} \bibinfo{person}{Elena~L
  Glassman}.} \bibinfo{year}{2020}\natexlab{}.
\newblock \showarticletitle{Proxy tasks and subjective measures can be
  misleading in evaluating explainable AI systems}. In
  \bibinfo{booktitle}{\emph{Proceedings of the 25th International Conference on
  Intelligent User Interfaces}}. \bibinfo{pages}{454--464}.
\newblock


\bibitem[\protect\citeauthoryear{Cai, Reif, Hegde, Hipp, Kim, Smilkov,
  Wattenberg, Viegas, Corrado, Stumpe, et~al\mbox{.}}{Cai
  et~al\mbox{.}}{2019}]%
        {cai2019human}
\bibfield{author}{\bibinfo{person}{Carrie~J Cai}, \bibinfo{person}{Emily Reif},
  \bibinfo{person}{Narayan Hegde}, \bibinfo{person}{Jason Hipp},
  \bibinfo{person}{Been Kim}, \bibinfo{person}{Daniel Smilkov},
  \bibinfo{person}{Martin Wattenberg}, \bibinfo{person}{Fernanda Viegas},
  \bibinfo{person}{Greg~S Corrado}, \bibinfo{person}{Martin~C Stumpe},
  {et~al\mbox{.}}} \bibinfo{year}{2019}\natexlab{}.
\newblock \showarticletitle{Human-centered tools for coping with imperfect
  algorithms during medical decision-making}. In
  \bibinfo{booktitle}{\emph{Proceedings of the 2019 CHI Conference on Human
  Factors in Computing Systems}}. ACM, \bibinfo{pages}{4}.
\newblock


\bibitem[\protect\citeauthoryear{Carton, Mei, and Resnick}{Carton
  et~al\mbox{.}}{2020}]%
        {carton2020feature}
\bibfield{author}{\bibinfo{person}{Samuel Carton}, \bibinfo{person}{Qiaozhu
  Mei}, {and} \bibinfo{person}{Paul Resnick}.} \bibinfo{year}{2020}\natexlab{}.
\newblock \showarticletitle{Feature-Based Explanations Don't Help People Detect
  Misclassifications of Online Toxicity}. In
  \bibinfo{booktitle}{\emph{Proceedings of the International AAAI Conference on
  Web and Social Media}}, Vol.~\bibinfo{volume}{14}. \bibinfo{pages}{95--106}.
\newblock


\bibitem[\protect\citeauthoryear{Cheng, Wang, Zhang, O'Connell, Gray, Harper,
  and Zhu}{Cheng et~al\mbox{.}}{2019}]%
        {cheng2019explaining}
\bibfield{author}{\bibinfo{person}{Hao-Fei Cheng}, \bibinfo{person}{Ruotong
  Wang}, \bibinfo{person}{Zheng Zhang}, \bibinfo{person}{Fiona O'Connell},
  \bibinfo{person}{Terrance Gray}, \bibinfo{person}{F~Maxwell Harper}, {and}
  \bibinfo{person}{Haiyi Zhu}.} \bibinfo{year}{2019}\natexlab{}.
\newblock \showarticletitle{Explaining Decision-Making Algorithms through UI:
  Strategies to Help Non-Expert Stakeholders}. In
  \bibinfo{booktitle}{\emph{Proceedings of the 2019 CHI Conference on Human
  Factors in Computing Systems}}. ACM, \bibinfo{pages}{559}.
\newblock


\bibitem[\protect\citeauthoryear{Chiang and Yin}{Chiang and Yin}{2021}]%
        {chiang2021you}
\bibfield{author}{\bibinfo{person}{Chun-Wei Chiang} {and} \bibinfo{person}{Ming
  Yin}.} \bibinfo{year}{2021}\natexlab{}.
\newblock \showarticletitle{You'd Better Stop! Understanding Human Reliance on
  Machine Learning Models under Covariate Shift}. In
  \bibinfo{booktitle}{\emph{13th ACM Web Science Conference 2021}}.
  \bibinfo{pages}{120--129}.
\newblock


\bibitem[\protect\citeauthoryear{Clark, Yatskar, and Zettlemoyer}{Clark
  et~al\mbox{.}}{2019}]%
        {clark2019don}
\bibfield{author}{\bibinfo{person}{Christopher Clark}, \bibinfo{person}{Mark
  Yatskar}, {and} \bibinfo{person}{Luke Zettlemoyer}.}
  \bibinfo{year}{2019}\natexlab{}.
\newblock \showarticletitle{Don't Take the Easy Way Out: Ensemble Based Methods
  for Avoiding Known Dataset Biases}. In \bibinfo{booktitle}{\emph{Proceedings
  of the 2019 Conference on Empirical Methods in Natural Language Processing
  and the 9th International Joint Conference on Natural Language Processing
  (EMNLP-IJCNLP)}}. \bibinfo{pages}{4069--4082}.
\newblock


\bibitem[\protect\citeauthoryear{De-Arteaga, Romanov, Wallach, Chayes, Borgs,
  Chouldechova, Geyik, Kenthapadi, and Kalai}{De-Arteaga et~al\mbox{.}}{2019}]%
        {de2019bias}
\bibfield{author}{\bibinfo{person}{Maria De-Arteaga}, \bibinfo{person}{Alexey
  Romanov}, \bibinfo{person}{Hanna Wallach}, \bibinfo{person}{Jennifer Chayes},
  \bibinfo{person}{Christian Borgs}, \bibinfo{person}{Alexandra Chouldechova},
  \bibinfo{person}{Sahin Geyik}, \bibinfo{person}{Krishnaram Kenthapadi}, {and}
  \bibinfo{person}{Adam~Tauman Kalai}.} \bibinfo{year}{2019}\natexlab{}.
\newblock \showarticletitle{Bias in bios: A case study of semantic
  representation bias in a high-stakes setting}. In
  \bibinfo{booktitle}{\emph{Proceedings of the Conference on Fairness,
  Accountability, and Transparency}}. \bibinfo{pages}{120--128}.
\newblock


\bibitem[\protect\citeauthoryear{de~Visser, Cohen, Freedy, and
  Parasuraman}{de~Visser et~al\mbox{.}}{2014}]%
        {de2014design}
\bibfield{author}{\bibinfo{person}{Ewart~J de Visser}, \bibinfo{person}{Marvin
  Cohen}, \bibinfo{person}{Amos Freedy}, {and} \bibinfo{person}{Raja
  Parasuraman}.} \bibinfo{year}{2014}\natexlab{}.
\newblock \showarticletitle{A design methodology for trust cue calibration in
  cognitive agents}. In \bibinfo{booktitle}{\emph{International conference on
  virtual, augmented and mixed reality}}. Springer, \bibinfo{pages}{251--262}.
\newblock


\bibitem[\protect\citeauthoryear{Doshi-Velez and Kim}{Doshi-Velez and
  Kim}{2017}]%
        {doshi2017towards}
\bibfield{author}{\bibinfo{person}{Finale Doshi-Velez} {and}
  \bibinfo{person}{Been Kim}.} \bibinfo{year}{2017}\natexlab{}.
\newblock \showarticletitle{Towards a rigorous science of interpretable machine
  learning}.
\newblock \bibinfo{journal}{\emph{arXiv preprint arXiv:1702.08608}}
  (\bibinfo{year}{2017}).
\newblock


\bibitem[\protect\citeauthoryear{Feng and Boyd-Graber}{Feng and
  Boyd-Graber}{2019}]%
        {feng2019can}
\bibfield{author}{\bibinfo{person}{Shi Feng} {and} \bibinfo{person}{Jordan
  Boyd-Graber}.} \bibinfo{year}{2019}\natexlab{}.
\newblock \showarticletitle{What can ai do for me? evaluating machine learning
  interpretations in cooperative play}. In
  \bibinfo{booktitle}{\emph{Proceedings of the 24th International Conference on
  Intelligent User Interfaces}}. \bibinfo{pages}{229--239}.
\newblock


\bibitem[\protect\citeauthoryear{Frid, Gomes, and Jin}{Frid
  et~al\mbox{.}}{2020}]%
        {frid2020music}
\bibfield{author}{\bibinfo{person}{Emma Frid}, \bibinfo{person}{Ceslo Gomes},
  {and} \bibinfo{person}{Zeyu Jin}.} \bibinfo{year}{2020}\natexlab{}.
\newblock \showarticletitle{Music Creation by Example}. In
  \bibinfo{booktitle}{\emph{Proceedings of the 2020 CHI Conference on Human
  Factors in Computing Systems}}.
\newblock


\bibitem[\protect\citeauthoryear{Gaur, Lasecki, Metze, and Bigham}{Gaur
  et~al\mbox{.}}{2016}]%
        {gaur2016effects}
\bibfield{author}{\bibinfo{person}{Yashesh Gaur}, \bibinfo{person}{Walter~S
  Lasecki}, \bibinfo{person}{Florian Metze}, {and} \bibinfo{person}{Jeffrey~P
  Bigham}.} \bibinfo{year}{2016}\natexlab{}.
\newblock \showarticletitle{The effects of automatic speech recognition quality
  on human transcription latency}. In \bibinfo{booktitle}{\emph{Proceedings of
  the 13th Web for All Conference}}. \bibinfo{pages}{1--8}.
\newblock


\bibitem[\protect\citeauthoryear{Ghai, Liao, Zhang, Bellamy, and Mueller}{Ghai
  et~al\mbox{.}}{2020}]%
        {ghai2020explainable}
\bibfield{author}{\bibinfo{person}{Bhavya Ghai}, \bibinfo{person}{Q~Vera Liao},
  \bibinfo{person}{Yunfeng Zhang}, \bibinfo{person}{Rachel Bellamy}, {and}
  \bibinfo{person}{Klaus Mueller}.} \bibinfo{year}{2020}\natexlab{}.
\newblock \showarticletitle{Explainable Active Learning (XAL): An Empirical
  Study of How Local Explanations Impact Annotator Experience}.
\newblock \bibinfo{journal}{\emph{arXiv preprint arXiv:2001.09219}}
  (\bibinfo{year}{2020}).
\newblock


\bibitem[\protect\citeauthoryear{Gilpin, Bau, Yuan, Bajwa, Specter, and
  Kagal}{Gilpin et~al\mbox{.}}{2018}]%
        {gilpin2018explaining}
\bibfield{author}{\bibinfo{person}{Leilani~H Gilpin}, \bibinfo{person}{David
  Bau}, \bibinfo{person}{Ben~Z Yuan}, \bibinfo{person}{Ayesha Bajwa},
  \bibinfo{person}{Michael Specter}, {and} \bibinfo{person}{Lalana Kagal}.}
  \bibinfo{year}{2018}\natexlab{}.
\newblock \showarticletitle{Explaining explanations: An overview of
  interpretability of machine learning}. In \bibinfo{booktitle}{\emph{2018 IEEE
  5th International Conference on data science and advanced analytics (DSAA)}}.
  IEEE, \bibinfo{pages}{80--89}.
\newblock


\bibitem[\protect\citeauthoryear{Goodfellow, Bengio, and Courville}{Goodfellow
  et~al\mbox{.}}{2016}]%
        {goodfellow2016machine}
\bibfield{author}{\bibinfo{person}{Ian Goodfellow}, \bibinfo{person}{Y Bengio},
  {and} \bibinfo{person}{A Courville}.} \bibinfo{year}{2016}\natexlab{}.
\newblock \showarticletitle{Machine learning basics}.
\newblock In \bibinfo{booktitle}{\emph{Deep learning}}.
  Vol.~\bibinfo{volume}{1}. \bibinfo{publisher}{MIT press},
  \bibinfo{pages}{98--164}.
\newblock


\bibitem[\protect\citeauthoryear{Green and Chen}{Green and Chen}{2019a}]%
        {green2019disparate}
\bibfield{author}{\bibinfo{person}{Ben Green} {and} \bibinfo{person}{Yiling
  Chen}.} \bibinfo{year}{2019}\natexlab{a}.
\newblock \showarticletitle{Disparate interactions: An algorithm-in-the-loop
  analysis of fairness in risk assessments}. In
  \bibinfo{booktitle}{\emph{Proceedings of the Conference on Fairness,
  Accountability, and Transparency}}. ACM, \bibinfo{pages}{90--99}.
\newblock


\bibitem[\protect\citeauthoryear{Green and Chen}{Green and Chen}{2019b}]%
        {green2019principles}
\bibfield{author}{\bibinfo{person}{Ben Green} {and} \bibinfo{person}{Yiling
  Chen}.} \bibinfo{year}{2019}\natexlab{b}.
\newblock \showarticletitle{The principles and limits of algorithm-in-the-loop
  decision making}.
\newblock \bibinfo{journal}{\emph{Proceedings of the ACM on Human-Computer
  Interaction}} \bibinfo{volume}{3}, \bibinfo{number}{CSCW}
  (\bibinfo{year}{2019}), \bibinfo{pages}{50}.
\newblock


\bibitem[\protect\citeauthoryear{Gurari, Li, Stangl, Guo, Lin, Grauman, Luo,
  and Bigham}{Gurari et~al\mbox{.}}{2018}]%
        {gurari2018vizwiz}
\bibfield{author}{\bibinfo{person}{Danna Gurari}, \bibinfo{person}{Qing Li},
  \bibinfo{person}{Abigale~J Stangl}, \bibinfo{person}{Anhong Guo},
  \bibinfo{person}{Chi Lin}, \bibinfo{person}{Kristen Grauman},
  \bibinfo{person}{Jiebo Luo}, {and} \bibinfo{person}{Jeffrey~P Bigham}.}
  \bibinfo{year}{2018}\natexlab{}.
\newblock \showarticletitle{Vizwiz grand challenge: Answering visual questions
  from blind people}. In \bibinfo{booktitle}{\emph{Proceedings of the IEEE
  Conference on Computer Vision and Pattern Recognition}}.
  \bibinfo{pages}{3608--3617}.
\newblock


\bibitem[\protect\citeauthoryear{He, Zhang, Ren, and Sun}{He
  et~al\mbox{.}}{2015}]%
        {he2015delving}
\bibfield{author}{\bibinfo{person}{Kaiming He}, \bibinfo{person}{Xiangyu
  Zhang}, \bibinfo{person}{Shaoqing Ren}, {and} \bibinfo{person}{Jian Sun}.}
  \bibinfo{year}{2015}\natexlab{}.
\newblock \showarticletitle{Delving deep into rectifiers: Surpassing
  human-level performance on imagenet classification}. In
  \bibinfo{booktitle}{\emph{Proceedings of ICCV}}.
\newblock


\bibitem[\protect\citeauthoryear{Hohman, Head, Caruana, DeLine, and
  Drucker}{Hohman et~al\mbox{.}}{2019}]%
        {hohman2019gamut}
\bibfield{author}{\bibinfo{person}{Fred Hohman}, \bibinfo{person}{Andrew Head},
  \bibinfo{person}{Rich Caruana}, \bibinfo{person}{Robert DeLine}, {and}
  \bibinfo{person}{Steven~M Drucker}.} \bibinfo{year}{2019}\natexlab{}.
\newblock \showarticletitle{Gamut: A design probe to understand how data
  scientists understand machine learning models}. In
  \bibinfo{booktitle}{\emph{Proceedings of the 2019 CHI Conference on Human
  Factors in Computing Systems}}. \bibinfo{pages}{1--13}.
\newblock


\bibitem[\protect\citeauthoryear{Jia and Liang}{Jia and Liang}{2017}]%
        {jia2017adversarial}
\bibfield{author}{\bibinfo{person}{Robin Jia} {and} \bibinfo{person}{Percy
  Liang}.} \bibinfo{year}{2017}\natexlab{}.
\newblock \showarticletitle{Adversarial Examples for Evaluating Reading
  Comprehension Systems}. In \bibinfo{booktitle}{\emph{Proceedings of the 2017
  Conference on Empirical Methods in Natural Language Processing}}.
  \bibinfo{pages}{2021--2031}.
\newblock


\bibitem[\protect\citeauthoryear{Kaur, Nori, Jenkins, Caruana, Wallach, and
  Wortman~Vaughan}{Kaur et~al\mbox{.}}{2020}]%
        {kaur2020interpreting}
\bibfield{author}{\bibinfo{person}{Harmanpreet Kaur}, \bibinfo{person}{Harsha
  Nori}, \bibinfo{person}{Samuel Jenkins}, \bibinfo{person}{Rich Caruana},
  \bibinfo{person}{Hanna Wallach}, {and} \bibinfo{person}{Jennifer
  Wortman~Vaughan}.} \bibinfo{year}{2020}\natexlab{}.
\newblock \showarticletitle{Interpreting Interpretability: Understanding Data
  Scientists' Use of Interpretability Tools for Machine Learning}. In
  \bibinfo{booktitle}{\emph{Proceedings of the 2020 CHI Conference on Human
  Factors in Computing Systems}}. \bibinfo{pages}{1--14}.
\newblock


\bibitem[\protect\citeauthoryear{Kleinberg, Lakkaraju, Leskovec, Ludwig, and
  Mullainathan}{Kleinberg et~al\mbox{.}}{2018}]%
        {kleinberg2018human}
\bibfield{author}{\bibinfo{person}{Jon Kleinberg}, \bibinfo{person}{Himabindu
  Lakkaraju}, \bibinfo{person}{Jure Leskovec}, \bibinfo{person}{Jens Ludwig},
  {and} \bibinfo{person}{Sendhil Mullainathan}.}
  \bibinfo{year}{2018}\natexlab{}.
\newblock \showarticletitle{Human decisions and machine predictions}.
\newblock \bibinfo{journal}{\emph{The quarterly journal of economics}}
  \bibinfo{volume}{133}, \bibinfo{number}{1} (\bibinfo{year}{2018}),
  \bibinfo{pages}{237--293}.
\newblock


\bibitem[\protect\citeauthoryear{Kleinberg, Ludwig, Mullainathan, and
  Obermeyer}{Kleinberg et~al\mbox{.}}{2015}]%
        {kleinberg2015prediction}
\bibfield{author}{\bibinfo{person}{Jon Kleinberg}, \bibinfo{person}{Jens
  Ludwig}, \bibinfo{person}{Sendhil Mullainathan}, {and} \bibinfo{person}{Ziad
  Obermeyer}.} \bibinfo{year}{2015}\natexlab{}.
\newblock \showarticletitle{Prediction policy problems}.
\newblock \bibinfo{journal}{\emph{American Economic Review}}
  \bibinfo{volume}{105}, \bibinfo{number}{5} (\bibinfo{year}{2015}),
  \bibinfo{pages}{491--95}.
\newblock


\bibitem[\protect\citeauthoryear{Koh and Liang}{Koh and Liang}{2017}]%
        {koh2017understanding}
\bibfield{author}{\bibinfo{person}{Pang~Wei Koh} {and} \bibinfo{person}{Percy
  Liang}.} \bibinfo{year}{2017}\natexlab{}.
\newblock \showarticletitle{Understanding black-box predictions via influence
  functions}. In \bibinfo{booktitle}{\emph{Proceedings of the 34th
  International Conference on Machine Learning-Volume 70}}. JMLR. org,
  \bibinfo{pages}{1885--1894}.
\newblock


\bibitem[\protect\citeauthoryear{Koh, Sagawa, Xie, Zhang, Balsubramani, Hu,
  Yasunaga, Phillips, Gao, Lee, et~al\mbox{.}}{Koh et~al\mbox{.}}{2021}]%
        {koh2021wilds}
\bibfield{author}{\bibinfo{person}{Pang~Wei Koh}, \bibinfo{person}{Shiori
  Sagawa}, \bibinfo{person}{Sang~Michael Xie}, \bibinfo{person}{Marvin Zhang},
  \bibinfo{person}{Akshay Balsubramani}, \bibinfo{person}{Weihua Hu},
  \bibinfo{person}{Michihiro Yasunaga}, \bibinfo{person}{Richard~Lanas
  Phillips}, \bibinfo{person}{Irena Gao}, \bibinfo{person}{Tony Lee},
  {et~al\mbox{.}}} \bibinfo{year}{2021}\natexlab{}.
\newblock \showarticletitle{Wilds: A benchmark of in-the-wild distribution
  shifts}. In \bibinfo{booktitle}{\emph{International Conference on Machine
  Learning}}. PMLR, \bibinfo{pages}{5637--5664}.
\newblock


\bibitem[\protect\citeauthoryear{Krause, Perer, and Ng}{Krause
  et~al\mbox{.}}{2016}]%
        {krause2016interacting}
\bibfield{author}{\bibinfo{person}{Josua Krause}, \bibinfo{person}{Adam Perer},
  {and} \bibinfo{person}{Kenney Ng}.} \bibinfo{year}{2016}\natexlab{}.
\newblock \showarticletitle{Interacting with predictions: Visual inspection of
  black-box machine learning models}. In \bibinfo{booktitle}{\emph{Proceedings
  of the 2016 CHI Conference on Human Factors in Computing Systems}}. ACM,
  \bibinfo{pages}{5686--5697}.
\newblock


\bibitem[\protect\citeauthoryear{Lai, Liu, and Tan}{Lai et~al\mbox{.}}{2020}]%
        {lai+liu+tan:20}
\bibfield{author}{\bibinfo{person}{Vivian Lai}, \bibinfo{person}{Han Liu},
  {and} \bibinfo{person}{Chenhao Tan}.} \bibinfo{year}{2020}\natexlab{}.
\newblock \showarticletitle{``Why is `Chicago' deceptive?'' Towards Building
  Model-Driven Tutorials for Humans}. In \bibinfo{booktitle}{\emph{Proceedings
  of the 2020 CHI Conference on Human Factors in Computing Systems}}.
  \bibinfo{pages}{1--13}.
\newblock


\bibitem[\protect\citeauthoryear{Lai and Tan}{Lai and Tan}{2019}]%
        {lai+tan:19}
\bibfield{author}{\bibinfo{person}{Vivian Lai} {and} \bibinfo{person}{Chenhao
  Tan}.} \bibinfo{year}{2019}\natexlab{}.
\newblock \showarticletitle{On human predictions with explanations and
  predictions of machine learning models: A case study on deception detection}.
  In \bibinfo{booktitle}{\emph{Proceedings of the conference on fairness,
  accountability, and transparency}}. \bibinfo{pages}{29--38}.
\newblock


\bibitem[\protect\citeauthoryear{Lakkaraju, Bach, and Leskovec}{Lakkaraju
  et~al\mbox{.}}{2016}]%
        {lakkaraju2016interpretable}
\bibfield{author}{\bibinfo{person}{Himabindu Lakkaraju},
  \bibinfo{person}{Stephen~H Bach}, {and} \bibinfo{person}{Jure Leskovec}.}
  \bibinfo{year}{2016}\natexlab{}.
\newblock \showarticletitle{Interpretable decision sets: A joint framework for
  description and prediction}. In \bibinfo{booktitle}{\emph{Proceedings of the
  22nd ACM SIGKDD international conference on knowledge discovery and data
  mining}}. \bibinfo{pages}{1675--1684}.
\newblock


\bibitem[\protect\citeauthoryear{Lasecki, Miller, Naim, Kushalnagar, Sadilek,
  Gildea, and Bigham}{Lasecki et~al\mbox{.}}{2017}]%
        {lasecki2017scribe}
\bibfield{author}{\bibinfo{person}{Walter~S Lasecki},
  \bibinfo{person}{Christopher~D Miller}, \bibinfo{person}{Iftekhar Naim},
  \bibinfo{person}{Raja Kushalnagar}, \bibinfo{person}{Adam Sadilek},
  \bibinfo{person}{Daniel Gildea}, {and} \bibinfo{person}{Jeffrey~P Bigham}.}
  \bibinfo{year}{2017}\natexlab{}.
\newblock \showarticletitle{Scribe: deep integration of human and machine
  intelligence to caption speech in real time}.
\newblock \bibinfo{journal}{\emph{Commun. ACM}} \bibinfo{volume}{60},
  \bibinfo{number}{9} (\bibinfo{year}{2017}), \bibinfo{pages}{93--100}.
\newblock


\bibitem[\protect\citeauthoryear{Liang and Zheng}{Liang and Zheng}{2020}]%
        {liang2020transfer}
\bibfield{author}{\bibinfo{person}{Gaobo Liang} {and} \bibinfo{person}{Lixin
  Zheng}.} \bibinfo{year}{2020}\natexlab{}.
\newblock \showarticletitle{A transfer learning method with deep residual
  network for pediatric pneumonia diagnosis}.
\newblock \bibinfo{journal}{\emph{Computer methods and programs in
  biomedicine}}  \bibinfo{volume}{187} (\bibinfo{year}{2020}),
  \bibinfo{pages}{104964}.
\newblock


\bibitem[\protect\citeauthoryear{Lin, Jung, Goel, Skeem, et~al\mbox{.}}{Lin
  et~al\mbox{.}}{2020}]%
        {jung2020limits}
\bibfield{author}{\bibinfo{person}{Zhiyuan~``Jerry'' Lin},
  \bibinfo{person}{Jongbin Jung}, \bibinfo{person}{Sharad Goel},
  \bibinfo{person}{Jennifer Skeem}, {et~al\mbox{.}}}
  \bibinfo{year}{2020}\natexlab{}.
\newblock \showarticletitle{The limits of human predictions of recidivism}.
\newblock \bibinfo{journal}{\emph{Science advances}} \bibinfo{volume}{6},
  \bibinfo{number}{7} (\bibinfo{year}{2020}), \bibinfo{pages}{eaaz0652}.
\newblock


\bibitem[\protect\citeauthoryear{Liptak}{Liptak}{2017}]%
        {nytimes}
\bibfield{author}{\bibinfo{person}{Adam Liptak}.}
  \bibinfo{year}{2017}\natexlab{}.
\newblock \bibinfo{title}{Sent to Prison by a Software Program's Secret
  Algorithms}.
\newblock
\newblock


\bibitem[\protect\citeauthoryear{Lipton}{Lipton}{2016}]%
        {lipton2016mythos}
\bibfield{author}{\bibinfo{person}{Zachary~C Lipton}.}
  \bibinfo{year}{2016}\natexlab{}.
\newblock \showarticletitle{The mythos of model interpretability}.
\newblock \bibinfo{journal}{\emph{arXiv preprint arXiv:1606.03490}}
  (\bibinfo{year}{2016}).
\newblock


\bibitem[\protect\citeauthoryear{Lombrozo}{Lombrozo}{2006}]%
        {lombrozo2006structure}
\bibfield{author}{\bibinfo{person}{Tania Lombrozo}.}
  \bibinfo{year}{2006}\natexlab{}.
\newblock \showarticletitle{The structure and function of explanations}.
\newblock \bibinfo{journal}{\emph{Trends in cognitive sciences}}
  \bibinfo{volume}{10}, \bibinfo{number}{10} (\bibinfo{year}{2006}),
  \bibinfo{pages}{464--470}.
\newblock


\bibitem[\protect\citeauthoryear{Louie, Coenen, Huang, Terry, and Cai}{Louie
  et~al\mbox{.}}{2020}]%
        {louie2020novice}
\bibfield{author}{\bibinfo{person}{Ryan Louie}, \bibinfo{person}{Andy Coenen},
  \bibinfo{person}{Cheng~Zhi Huang}, \bibinfo{person}{Michael Terry}, {and}
  \bibinfo{person}{Carrie~J Cai}.} \bibinfo{year}{2020}\natexlab{}.
\newblock \showarticletitle{Novice-AI Music Co-Creation via AI-Steering Tools
  for Deep Generative Models}. In \bibinfo{booktitle}{\emph{Proceedings of the
  2020 CHI Conference on Human Factors in Computing Systems}}.
  \bibinfo{pages}{1--13}.
\newblock


\bibitem[\protect\citeauthoryear{Lundberg and Lee}{Lundberg and Lee}{2017}]%
        {lundberg2017unified}
\bibfield{author}{\bibinfo{person}{Scott~M Lundberg} {and}
  \bibinfo{person}{Su-In Lee}.} \bibinfo{year}{2017}\natexlab{}.
\newblock \showarticletitle{A unified approach to interpreting model
  predictions}. In \bibinfo{booktitle}{\emph{Proceedings of the 31st
  international conference on neural information processing systems}}.
  \bibinfo{pages}{4768--4777}.
\newblock


\bibitem[\protect\citeauthoryear{Lundberg, Nair, Vavilala, Horibe, Eisses,
  Adams, Liston, Low, Newman, Kim, et~al\mbox{.}}{Lundberg
  et~al\mbox{.}}{2018}]%
        {lundberg2018explainable}
\bibfield{author}{\bibinfo{person}{Scott~M Lundberg}, \bibinfo{person}{Bala
  Nair}, \bibinfo{person}{Monica~S Vavilala}, \bibinfo{person}{Mayumi Horibe},
  \bibinfo{person}{Michael~J Eisses}, \bibinfo{person}{Trevor Adams},
  \bibinfo{person}{David~E Liston}, \bibinfo{person}{Daniel King-Wai Low},
  \bibinfo{person}{Shu-Fang Newman}, \bibinfo{person}{Jerry Kim},
  {et~al\mbox{.}}} \bibinfo{year}{2018}\natexlab{}.
\newblock \showarticletitle{Explainable machine-learning predictions for the
  prevention of hypoxaemia during surgery}.
\newblock \bibinfo{journal}{\emph{Nature biomedical engineering}}
  \bibinfo{volume}{2}, \bibinfo{number}{10} (\bibinfo{year}{2018}),
  \bibinfo{pages}{749--760}.
\newblock


\bibitem[\protect\citeauthoryear{McBride and Morgan}{McBride and
  Morgan}{2010}]%
        {mcbride2010trust}
\bibfield{author}{\bibinfo{person}{Maranda McBride} {and}
  \bibinfo{person}{Shona Morgan}.} \bibinfo{year}{2010}\natexlab{}.
\newblock \showarticletitle{Trust calibration for automated decision aids}.
\newblock \bibinfo{journal}{\emph{Institute for Homeland Security Solutions}}
  (\bibinfo{year}{2010}), \bibinfo{pages}{1--11}.
\newblock


\bibitem[\protect\citeauthoryear{McCormack, Gifford, Hutchings,
  Llano~Rodriguez, Yee-King, and d'Inverno}{McCormack et~al\mbox{.}}{2019}]%
        {mccormack2019silent}
\bibfield{author}{\bibinfo{person}{Jon McCormack}, \bibinfo{person}{Toby
  Gifford}, \bibinfo{person}{Patrick Hutchings}, \bibinfo{person}{Maria~Teresa
  Llano~Rodriguez}, \bibinfo{person}{Matthew Yee-King}, {and}
  \bibinfo{person}{Mark d'Inverno}.} \bibinfo{year}{2019}\natexlab{}.
\newblock \showarticletitle{In a silent way: Communication between ai and
  improvising musicians beyond sound}. In \bibinfo{booktitle}{\emph{Proceedings
  of the 2019 CHI Conference on Human Factors in Computing Systems}}.
  \bibinfo{pages}{1--11}.
\newblock


\bibitem[\protect\citeauthoryear{McCoy, Pavlick, and Linzen}{McCoy
  et~al\mbox{.}}{2019}]%
        {mccoy-etal-2019-right}
\bibfield{author}{\bibinfo{person}{Tom McCoy}, \bibinfo{person}{Ellie Pavlick},
  {and} \bibinfo{person}{Tal Linzen}.} \bibinfo{year}{2019}\natexlab{}.
\newblock \showarticletitle{Right for the Wrong Reasons: Diagnosing Syntactic
  Heuristics in Natural Language Inference}. In
  \bibinfo{booktitle}{\emph{Proceedings of the 57th Annual Meeting of the
  Association for Computational Linguistics}}. \bibinfo{publisher}{Association
  for Computational Linguistics}, \bibinfo{address}{Florence, Italy},
  \bibinfo{pages}{3428--3448}.
\newblock
\urldef\tempurl%
\url{https://doi.org/10.18653/v1/P19-1334}
\showDOI{\tempurl}


\bibitem[\protect\citeauthoryear{McGuirl and Sarter}{McGuirl and
  Sarter}{2006}]%
        {mcguirl2006supporting}
\bibfield{author}{\bibinfo{person}{John~M McGuirl} {and}
  \bibinfo{person}{Nadine~B Sarter}.} \bibinfo{year}{2006}\natexlab{}.
\newblock \showarticletitle{Supporting trust calibration and the effective use
  of decision aids by presenting dynamic system confidence information}.
\newblock \bibinfo{journal}{\emph{Human factors}} \bibinfo{volume}{48},
  \bibinfo{number}{4} (\bibinfo{year}{2006}), \bibinfo{pages}{656--665}.
\newblock


\bibitem[\protect\citeauthoryear{McKinney, Sieniek, Godbole, Godwin, Antropova,
  Ashrafian, Back, Chesus, Corrado, Darzi, et~al\mbox{.}}{McKinney
  et~al\mbox{.}}{2020}]%
        {mckinney2020international}
\bibfield{author}{\bibinfo{person}{Scott~Mayer McKinney},
  \bibinfo{person}{Marcin Sieniek}, \bibinfo{person}{Varun Godbole},
  \bibinfo{person}{Jonathan Godwin}, \bibinfo{person}{Natasha Antropova},
  \bibinfo{person}{Hutan Ashrafian}, \bibinfo{person}{Trevor Back},
  \bibinfo{person}{Mary Chesus}, \bibinfo{person}{Greg~C Corrado},
  \bibinfo{person}{Ara Darzi}, {et~al\mbox{.}}}
  \bibinfo{year}{2020}\natexlab{}.
\newblock \showarticletitle{International evaluation of an AI system for breast
  cancer screening}.
\newblock \bibinfo{journal}{\emph{Nature}} \bibinfo{volume}{577},
  \bibinfo{number}{7788} (\bibinfo{year}{2020}), \bibinfo{pages}{89--94}.
\newblock


\bibitem[\protect\citeauthoryear{Merritt, Lee, Unnerstall, and Huber}{Merritt
  et~al\mbox{.}}{2015}]%
        {merritt2015well}
\bibfield{author}{\bibinfo{person}{Stephanie~M Merritt},
  \bibinfo{person}{Deborah Lee}, \bibinfo{person}{Jennifer~L Unnerstall}, {and}
  \bibinfo{person}{Kelli Huber}.} \bibinfo{year}{2015}\natexlab{}.
\newblock \showarticletitle{Are well-calibrated users effective users?
  Associations between calibration of trust and performance on an
  automation-aided task}.
\newblock \bibinfo{journal}{\emph{Human Factors}} \bibinfo{volume}{57},
  \bibinfo{number}{1} (\bibinfo{year}{2015}), \bibinfo{pages}{34--47}.
\newblock


\bibitem[\protect\citeauthoryear{Miller}{Miller}{2018}]%
        {miller2018explanation}
\bibfield{author}{\bibinfo{person}{Tim Miller}.}
  \bibinfo{year}{2018}\natexlab{}.
\newblock \showarticletitle{Explanation in artificial intelligence: Insights
  from the social sciences}.
\newblock \bibinfo{journal}{\emph{Artificial Intelligence}}
  (\bibinfo{year}{2018}).
\newblock


\bibitem[\protect\citeauthoryear{Muir}{Muir}{1987}]%
        {muir1987trust}
\bibfield{author}{\bibinfo{person}{Bonnie~M Muir}.}
  \bibinfo{year}{1987}\natexlab{}.
\newblock \showarticletitle{Trust between humans and machines, and the design
  of decision aids}.
\newblock \bibinfo{journal}{\emph{International journal of man-machine
  studies}} \bibinfo{volume}{27}, \bibinfo{number}{5-6} (\bibinfo{year}{1987}),
  \bibinfo{pages}{527--539}.
\newblock


\bibitem[\protect\citeauthoryear{Nielsen, Clemmensen, and Yssing}{Nielsen
  et~al\mbox{.}}{2002}]%
        {nielsen2002getting}
\bibfield{author}{\bibinfo{person}{Janni Nielsen}, \bibinfo{person}{Torkil
  Clemmensen}, {and} \bibinfo{person}{Carsten Yssing}.}
  \bibinfo{year}{2002}\natexlab{}.
\newblock \showarticletitle{Getting access to what goes on in people's heads?:
  reflections on the think-aloud technique}. In
  \bibinfo{booktitle}{\emph{Proceedings of the second Nordic conference on
  Human-computer interaction}}. ACM, \bibinfo{pages}{101--110}.
\newblock


\bibitem[\protect\citeauthoryear{Parasuraman and Riley}{Parasuraman and
  Riley}{1997}]%
        {parasuraman1997humans}
\bibfield{author}{\bibinfo{person}{Raja Parasuraman} {and}
  \bibinfo{person}{Victor Riley}.} \bibinfo{year}{1997}\natexlab{}.
\newblock \showarticletitle{Humans and automation: Use, misuse, disuse, abuse}.
\newblock \bibinfo{journal}{\emph{Human factors}} \bibinfo{volume}{39},
  \bibinfo{number}{2} (\bibinfo{year}{1997}), \bibinfo{pages}{230--253}.
\newblock


\bibitem[\protect\citeauthoryear{Poursabzi-Sangdeh, Goldstein, Hofman,
  Wortman~Vaughan, and Wallach}{Poursabzi-Sangdeh et~al\mbox{.}}{2021}]%
        {poursabzi2021manipulating}
\bibfield{author}{\bibinfo{person}{Forough Poursabzi-Sangdeh},
  \bibinfo{person}{Daniel~G Goldstein}, \bibinfo{person}{Jake~M Hofman},
  \bibinfo{person}{Jennifer~Wortman Wortman~Vaughan}, {and}
  \bibinfo{person}{Hanna Wallach}.} \bibinfo{year}{2021}\natexlab{}.
\newblock \showarticletitle{Manipulating and measuring model interpretability}.
  In \bibinfo{booktitle}{\emph{Proceedings of the 2021 CHI Conference on Human
  Factors in Computing Systems}}. \bibinfo{pages}{1--52}.
\newblock


\bibitem[\protect\citeauthoryear{Quionero-Candela, Sugiyama, Schwaighofer, and
  Lawrence}{Quionero-Candela et~al\mbox{.}}{2009}]%
        {quionero2009dataset}
\bibfield{author}{\bibinfo{person}{Joaquin Quionero-Candela},
  \bibinfo{person}{Masashi Sugiyama}, \bibinfo{person}{Anton Schwaighofer},
  {and} \bibinfo{person}{Neil~D Lawrence}.} \bibinfo{year}{2009}\natexlab{}.
\newblock \bibinfo{booktitle}{\emph{Dataset shift in machine learning}}.
\newblock \bibinfo{publisher}{The MIT Press}.
\newblock


\bibitem[\protect\citeauthoryear{Ribeiro, Singh, and Guestrin}{Ribeiro
  et~al\mbox{.}}{2016}]%
        {ribeiro2016should}
\bibfield{author}{\bibinfo{person}{Marco~Tulio Ribeiro},
  \bibinfo{person}{Sameer Singh}, {and} \bibinfo{person}{Carlos Guestrin}.}
  \bibinfo{year}{2016}\natexlab{}.
\newblock \showarticletitle{Why should i trust you?: Explaining the predictions
  of any classifier}. In \bibinfo{booktitle}{\emph{Proceedings of KDD}}.
\newblock


\bibitem[\protect\citeauthoryear{Silver, Hubert, Schrittwieser, Antonoglou,
  Lai, Guez, Lanctot, Sifre, Kumaran, Graepel, et~al\mbox{.}}{Silver
  et~al\mbox{.}}{2018}]%
        {silver2018general}
\bibfield{author}{\bibinfo{person}{David Silver}, \bibinfo{person}{Thomas
  Hubert}, \bibinfo{person}{Julian Schrittwieser}, \bibinfo{person}{Ioannis
  Antonoglou}, \bibinfo{person}{Matthew Lai}, \bibinfo{person}{Arthur Guez},
  \bibinfo{person}{Marc Lanctot}, \bibinfo{person}{Laurent Sifre},
  \bibinfo{person}{Dharshan Kumaran}, \bibinfo{person}{Thore Graepel},
  {et~al\mbox{.}}} \bibinfo{year}{2018}\natexlab{}.
\newblock \showarticletitle{A general reinforcement learning algorithm that
  masters chess, shogi, and Go through self-play}.
\newblock \bibinfo{journal}{\emph{Science}} \bibinfo{volume}{362},
  \bibinfo{number}{6419} (\bibinfo{year}{2018}), \bibinfo{pages}{1140--1144}.
\newblock


\bibitem[\protect\citeauthoryear{Sugiyama and Kawanabe}{Sugiyama and
  Kawanabe}{2012}]%
        {sugiyama2012machine}
\bibfield{author}{\bibinfo{person}{Masashi Sugiyama} {and}
  \bibinfo{person}{Motoaki Kawanabe}.} \bibinfo{year}{2012}\natexlab{}.
\newblock \bibinfo{booktitle}{\emph{Machine learning in non-stationary
  environments: Introduction to covariate shift adaptation}}.
\newblock \bibinfo{publisher}{MIT press}.
\newblock


\bibitem[\protect\citeauthoryear{{Supreme Court of the United States}}{{Supreme
  Court of the United States}}{1993}]%
        {daubert}
\bibfield{author}{\bibinfo{person}{{Supreme Court of the United States}}.}
  \bibinfo{year}{1993}\natexlab{}.
\newblock \bibinfo{title}{Daubert v. Merrell Dow Pharmaceuticals, Inc.}
\newblock
\newblock
\newblock
\shownote{509 U.S. 579.}


\bibitem[\protect\citeauthoryear{{Supreme Court of Wisconsin}}{{Supreme Court
  of Wisconsin}}{2016}]%
        {wicourts}
\bibfield{author}{\bibinfo{person}{{Supreme Court of Wisconsin}}.}
  \bibinfo{year}{2016}\natexlab{}.
\newblock \bibinfo{title}{{State of Wisconsin, Plaintiff-Respondent, v. Eric L.
  Loomis, Defendant-Appellant}}.
\newblock
\newblock
\urldef\tempurl%
\url{https://www.wicourts.gov/sc/opinion/DisplayDocument.pdf?content=pdf&seqNo=171690}
\showURL{%
\tempurl}


\bibitem[\protect\citeauthoryear{Tenney, Wexler, Bastings, Bolukbasi, Coenen,
  Gehrmann, Jiang, Pushkarna, Radebaugh, Reif, et~al\mbox{.}}{Tenney
  et~al\mbox{.}}{2020}]%
        {tenney2020language}
\bibfield{author}{\bibinfo{person}{Ian Tenney}, \bibinfo{person}{James Wexler},
  \bibinfo{person}{Jasmijn Bastings}, \bibinfo{person}{Tolga Bolukbasi},
  \bibinfo{person}{Andy Coenen}, \bibinfo{person}{Sebastian Gehrmann},
  \bibinfo{person}{Ellen Jiang}, \bibinfo{person}{Mahima Pushkarna},
  \bibinfo{person}{Carey Radebaugh}, \bibinfo{person}{Emily Reif},
  {et~al\mbox{.}}} \bibinfo{year}{2020}\natexlab{}.
\newblock \showarticletitle{The Language Interpretability Tool: Extensible,
  Interactive Visualizations and Analysis for NLP Models}. In
  \bibinfo{booktitle}{\emph{Proceedings of the 2020 Conference on Empirical
  Methods in Natural Language Processing: System Demonstrations}}.
  \bibinfo{pages}{107--118}.
\newblock


\bibitem[\protect\citeauthoryear{Torrey and Shavlik}{Torrey and
  Shavlik}{2010}]%
        {torrey2010transfer}
\bibfield{author}{\bibinfo{person}{Lisa Torrey} {and} \bibinfo{person}{Jude
  Shavlik}.} \bibinfo{year}{2010}\natexlab{}.
\newblock \showarticletitle{Transfer learning}.
\newblock In \bibinfo{booktitle}{\emph{Handbook of research on machine learning
  applications and trends: algorithms, methods, and techniques}}.
  \bibinfo{publisher}{IGI global}, \bibinfo{pages}{242--264}.
\newblock


\bibitem[\protect\citeauthoryear{{United States Department of Justice. Office
  of Justice Programs. Bureau of Justice Statistics.}}{{United States
  Department of Justice. Office of Justice Programs. Bureau of Justice
  Statistics.}}{2014}]%
        {icpsr}
\bibfield{author}{\bibinfo{person}{{United States Department of Justice. Office
  of Justice Programs. Bureau of Justice Statistics.}}}
  \bibinfo{year}{2014}\natexlab{}.
\newblock \bibinfo{title}{State Court Processing Statistics, 1990-2009: Felony
  Defendants in Large Urban Counties.}
\newblock
\newblock


\bibitem[\protect\citeauthoryear{Wachter, Mittelstadt, and Russell}{Wachter
  et~al\mbox{.}}{2017}]%
        {wachter2017counterfactual}
\bibfield{author}{\bibinfo{person}{Sandra Wachter}, \bibinfo{person}{Brent
  Mittelstadt}, {and} \bibinfo{person}{Chris Russell}.}
  \bibinfo{year}{2017}\natexlab{}.
\newblock \showarticletitle{Counterfactual explanations without opening the
  black box: Automated decisions and the GDPR}.
\newblock


\bibitem[\protect\citeauthoryear{Wang and Yin}{Wang and Yin}{2021}]%
        {wang2021explanations}
\bibfield{author}{\bibinfo{person}{Xinru Wang} {and} \bibinfo{person}{Ming
  Yin}.} \bibinfo{year}{2021}\natexlab{}.
\newblock \showarticletitle{Are Explanations Helpful? A Comparative Study of
  the Effects of Explanations in AI-Assisted Decision-Making}. In
  \bibinfo{booktitle}{\emph{26th International Conference on Intelligent User
  Interfaces}}. \bibinfo{pages}{318--328}.
\newblock


\bibitem[\protect\citeauthoryear{Weerts, van Ipenburg, and Pechenizkiy}{Weerts
  et~al\mbox{.}}{2019}]%
        {weerts2019human}
\bibfield{author}{\bibinfo{person}{Hilde~JP Weerts}, \bibinfo{person}{Werner
  van Ipenburg}, {and} \bibinfo{person}{Mykola Pechenizkiy}.}
  \bibinfo{year}{2019}\natexlab{}.
\newblock \showarticletitle{A Human-Grounded Evaluation of SHAP for Alert
  Processing}.
\newblock \bibinfo{journal}{\emph{arXiv preprint arXiv:1907.03324}}
  (\bibinfo{year}{2019}).
\newblock


\bibitem[\protect\citeauthoryear{Wexler, Pushkarna, Bolukbasi, Wattenberg,
  Vi{\'e}gas, and Wilson}{Wexler et~al\mbox{.}}{2019}]%
        {wexler2019if}
\bibfield{author}{\bibinfo{person}{James Wexler}, \bibinfo{person}{Mahima
  Pushkarna}, \bibinfo{person}{Tolga Bolukbasi}, \bibinfo{person}{Martin
  Wattenberg}, \bibinfo{person}{Fernanda Vi{\'e}gas}, {and}
  \bibinfo{person}{Jimbo Wilson}.} \bibinfo{year}{2019}\natexlab{}.
\newblock \showarticletitle{The what-if tool: Interactive probing of machine
  learning models}.
\newblock \bibinfo{journal}{\emph{IEEE transactions on visualization and
  computer graphics}} \bibinfo{volume}{26}, \bibinfo{number}{1}
  (\bibinfo{year}{2019}), \bibinfo{pages}{56--65}.
\newblock


\bibitem[\protect\citeauthoryear{Wu, Wieland, Farivar, and Schiller}{Wu
  et~al\mbox{.}}{2017}]%
        {wu2017automatic}
\bibfield{author}{\bibinfo{person}{Shaomei Wu}, \bibinfo{person}{Jeffrey
  Wieland}, \bibinfo{person}{Omid Farivar}, {and} \bibinfo{person}{Julie
  Schiller}.} \bibinfo{year}{2017}\natexlab{}.
\newblock \showarticletitle{Automatic alt-text: Computer-generated image
  descriptions for blind users on a social network service}. In
  \bibinfo{booktitle}{\emph{Proceedings of the 2017 ACM Conference on Computer
  Supported Cooperative Work and Social Computing}}.
  \bibinfo{pages}{1180--1192}.
\newblock


\bibitem[\protect\citeauthoryear{Wu, Weld, and Heer}{Wu et~al\mbox{.}}{2019}]%
        {wu2019local}
\bibfield{author}{\bibinfo{person}{Tongshuang Wu}, \bibinfo{person}{Daniel~S
  Weld}, {and} \bibinfo{person}{Jeffrey Heer}.}
  \bibinfo{year}{2019}\natexlab{}.
\newblock \showarticletitle{Local Decision Pitfalls in Interactive Machine
  Learning: An Investigation into Feature Selection in Sentiment Analysis}.
\newblock \bibinfo{journal}{\emph{ACM Transactions on Computer-Human
  Interaction (TOCHI)}} \bibinfo{volume}{26}, \bibinfo{number}{4}
  (\bibinfo{year}{2019}), \bibinfo{pages}{1--27}.
\newblock


\bibitem[\protect\citeauthoryear{Xie, Chen, Kao, Gao, and Chen}{Xie
  et~al\mbox{.}}{2020}]%
        {xie2020chexplain}
\bibfield{author}{\bibinfo{person}{Yao Xie}, \bibinfo{person}{Melody Chen},
  \bibinfo{person}{David Kao}, \bibinfo{person}{Ge Gao}, {and}
  \bibinfo{person}{Xiang~`Anthony' Chen}.} \bibinfo{year}{2020}\natexlab{}.
\newblock \showarticletitle{CheXplain: Enabling Physicians to Explore and
  Understand Data-Driven, AI-Enabled Medical Imaging Analysis}. In
  \bibinfo{booktitle}{\emph{Proceedings of the 2020 CHI Conference on Human
  Factors in Computing Systems}}. \bibinfo{pages}{1--13}.
\newblock


\bibitem[\protect\citeauthoryear{Yang, Steinfeld, and Zimmerman}{Yang
  et~al\mbox{.}}{2019}]%
        {yang2019unremarkable}
\bibfield{author}{\bibinfo{person}{Qian Yang}, \bibinfo{person}{Aaron
  Steinfeld}, {and} \bibinfo{person}{John Zimmerman}.}
  \bibinfo{year}{2019}\natexlab{}.
\newblock \showarticletitle{Unremarkable ai: Fitting intelligent decision
  support into critical, clinical decision-making processes}. In
  \bibinfo{booktitle}{\emph{Proceedings of the 2019 CHI Conference on Human
  Factors in Computing Systems}}. \bibinfo{pages}{1--11}.
\newblock


\bibitem[\protect\citeauthoryear{Yin, Wortman~Vaughan, and Wallach}{Yin
  et~al\mbox{.}}{2019}]%
        {yin2019understanding}
\bibfield{author}{\bibinfo{person}{Ming Yin}, \bibinfo{person}{Jennifer
  Wortman~Vaughan}, {and} \bibinfo{person}{Hanna Wallach}.}
  \bibinfo{year}{2019}\natexlab{}.
\newblock \showarticletitle{Understanding the Effect of Accuracy on Trust in
  Machine Learning Models}. In \bibinfo{booktitle}{\emph{Proceedings of the
  2019 CHI Conference on Human Factors in Computing Systems}}. ACM,
  \bibinfo{pages}{279}.
\newblock


\bibitem[\protect\citeauthoryear{Zhang, Liao, and Bellamy}{Zhang
  et~al\mbox{.}}{2020}]%
        {zhang2020effect}
\bibfield{author}{\bibinfo{person}{Yunfeng Zhang}, \bibinfo{person}{Q~Vera
  Liao}, {and} \bibinfo{person}{Rachel~KE Bellamy}.}
  \bibinfo{year}{2020}\natexlab{}.
\newblock \showarticletitle{Effect of confidence and explanation on accuracy
  and trust calibration in AI-assisted decision making}. In
  \bibinfo{booktitle}{\emph{Proceedings of the 2020 Conference on Fairness,
  Accountability, and Transparency}}. \bibinfo{pages}{295--305}.
\newblock


\bibitem[\protect\citeauthoryear{Zhuang, Qi, Duan, Xi, Zhu, Zhu, Xiong, and
  He}{Zhuang et~al\mbox{.}}{2020}]%
        {zhuang2020comprehensive}
\bibfield{author}{\bibinfo{person}{Fuzhen Zhuang}, \bibinfo{person}{Zhiyuan
  Qi}, \bibinfo{person}{Keyu Duan}, \bibinfo{person}{Dongbo Xi},
  \bibinfo{person}{Yongchun Zhu}, \bibinfo{person}{Hengshu Zhu},
  \bibinfo{person}{Hui Xiong}, {and} \bibinfo{person}{Qing He}.}
  \bibinfo{year}{2020}\natexlab{}.
\newblock \showarticletitle{A comprehensive survey on transfer learning}.
\newblock \bibinfo{journal}{\emph{Proc. IEEE}} (\bibinfo{year}{2020}).
\newblock


\end{thebibliography}

\received{January 2021}
\received[revised]{April 2021}
\received[revised]{July 2021}
\received[accepted]{July 2021}

\end{document}